\documentclass{article} 
\usepackage{arxiv}
\usepackage{times}
\usepackage[round]{natbib}

\usepackage{amsmath,amsfonts,bm}

\def\eqref#1{equation~\ref{#1}}

\def\floor#1{\lfloor #1 \rfloor}
\def\1{\bm{1}}

\DeclareMathAlphabet{\mathsfit}{\encodingdefault}{\sfdefault}{m}{sl}
\SetMathAlphabet{\mathsfit}{bold}{\encodingdefault}{\sfdefault}{bx}{n}

\usepackage{hyperref}
\usepackage{url}

\usepackage{graphicx}
\usepackage{booktabs}
\usepackage{makecell}  
\usepackage{amssymb}  
\usepackage{adjustbox}
\usepackage{array}
\usepackage{rotating}
\usepackage{xcolor}
\usepackage{colortbl}
\usepackage{longtable}  
\usepackage{arydshln}  

\title{\normalfont From the Zoo to the BULLPEN: Sample-Efficient LLM Profiling via Multi-Channel Model Embeddings}

\author{
  \textbf{Russell Ro}$^{1,2}$, \textbf{Qiming Cui}$^{1}$, \textbf{Reza Abbasi-Asl}$^{1}$\thanks{Corresponding author: \texttt{reza.abbasiasl@ucsf.edu}} \\[6pt]
  $^{1}$University of California, San Francisco \quad
  $^{2}$University of California, Berkeley \\[4pt]
}

\date{October 28, 2026}

\newcommand{\bpArenaRatedN}{42}
\newcommand{\bpNModels}{260}
\newcommand{\bpNFullset}{104}
\newcommand{\bpNExtset}{156}
\newcommand{\bpNProbe}{31,286}
\newcommand{\bpNPool}{24,118}
\newcommand{\bpNCore}{1,181}
\newcommand{\bpNNew}{20}
\newcommand{\bpNEst}{26}
\newcommand{\bpNAll}{46}

\newcommand{\bpPrefScalar}{0.71}

\newcommand{\bpEstPcaBits}{0.44}
\newcommand{\bpEstJeIrt}{0.65}
\newcommand{\bpAllJeIrt}{0.52}

\newcommand{\bpPrefJeIrt}{0.69}

\newcommand{\bpEstIrtNet}{0.55}
\newcommand{\bpNewAnsFull}{0.48}

\newcommand{\bpAllAnsFull}{0.46}
\newcommand{\bpNewFusedFull}{0.43}
\newcommand{\bpEstFusedFull}{0.55}
\newcommand{\bpAllFusedFull}{0.50}
\newcommand{\bpNewBestPrior}{0.35}

\newcommand{\bpEstPcaBitsPm}{0.08}

\newcommand{\bpEstJeIrtPm}{0.08}
\newcommand{\bpEstIrtNetPm}{0.07}

\newcommand{\bpEstFusedFullPm}{0.07}

\newcommand{\bpNewAnsFullPm}{0.10}
\newcommand{\bpNewBestPriorPm}{0.10}
\newcommand{\bpNewFusedFullPm}{0.09}

\newcommand{\bpNewOursTop}{6}
\newcommand{\bpLeadNew}{13}

\newcommand{\bpLeadSep}{6}
\newcommand{\bpLiftBitsPos}{38}
\newcommand{\bpLiftBitsN}{46}

\newcommand{\bpLiftBitsQPos}{31}

\newcommand{\bpMeff}{36.9}
\newcommand{\bpPooledN}{31}
\newcommand{\bpTwinAnsDrop}{0.46}
\newcommand{\bpTwinAnsDropLo}{0.40}
\newcommand{\bpTwinAnsDropHi}{0.52}

\newcommand{\bpTwinFusedDrop}{0.07}
\newcommand{\bpTwinFusedDropLo}{0.03}
\newcommand{\bpTwinFusedDropHi}{0.11}
\newcommand{\bpTwinFusedVsPca}{+0.06}
\newcommand{\bpTwinFusedVsPcaLo}{+0.02}
\newcommand{\bpTwinFusedVsPcaHi}{+0.11}
\newcommand{\bpCompNeedBoth}{0}
\newcommand{\bpCompTested}{31}

\newcommand{\bpRetTasks}{43}

\newcommand{\bpRetKTwoHundred}{93.5}

\newcommand{\bpRetPctOne}{95.1}

\newcommand{\bpCostShareFull}{1.3}
\newcommand{\bpCostRatioFull}{75}
\newcommand{\bpMapOwnN}{247}

\newcommand{\bpMapOwnMinusStandNew}{$-$0.10}
\newcommand{\bpMapOwnMinusStandNewLo}{$-$0.16}
\newcommand{\bpMapOwnMinusStandNewHi}{$-$0.04}
\newcommand{\bpArenaMinusAdapted}{$-$0.17}
\newcommand{\bpArenaMinusAdaptedLo}{$-$0.31}
\newcommand{\bpArenaMinusAdaptedHi}{$-$0.01}
\newcommand{\bpArenaN}{19}
\newcommand{\bpSigLeadSep}{0}
\newcommand{\bpSigClasses}{10}
\newcommand{\bpSigTextBitsNew}{+0.21}
\newcommand{\bpSigTextBitsNewLo}{+0.12}
\newcommand{\bpSigTextBitsNewHi}{+0.30}

\newcommand{\bpSigLeadAll}{+0.02}
\newcommand{\bpSigLeadAllLo}{$-$0.04}
\newcommand{\bpSigLeadAllHi}{+0.06}

\newcommand{\bpFullTasks}{47}
\newcommand{\bpPoolN}{24,118}

\newcommand{\bpPoolAnsFourHundred}{102}
\newcommand{\bpPoolAnsNineHundred}{100}
\newcommand{\bpPoolAnsTwoThousand}{100}
\newcommand{\bpPoolAnsFourHundredLo}{99}
\newcommand{\bpPoolAnsFourHundredHi}{104}

\newcommand{\bpPoolBitsAnsFourHundred}{99}
\newcommand{\bpPoolBitsAnsNineHundred}{99}
\newcommand{\bpPoolBitsAnsTwoThousand}{101}
\newcommand{\bpPoolBitsAnsFourHundredLo}{98}
\newcommand{\bpPoolBitsAnsFourHundredHi}{99}

\newcommand{\bpPoolFusedFourHundred}{91}
\newcommand{\bpPoolFusedNineHundred}{101}
\newcommand{\bpPoolFusedTwoThousand}{100}
\newcommand{\bpPoolFusedFourHundredLo}{91}
\newcommand{\bpPoolFusedFourHundredHi}{92}

\newcommand{\bpPoolPcaBitsFourHundred}{83}
\newcommand{\bpPoolPcaBitsNineHundred}{90}
\newcommand{\bpPoolPcaBitsTwoThousand}{98}
\newcommand{\bpPoolPcaBitsFourHundredLo}{78}
\newcommand{\bpPoolPcaBitsFourHundredHi}{86}

\newcommand{\bpPoolPcaNeedK}{2,000}
\newcommand{\bpNTypeRegression}{34}
\newcommand{\bpNTypeClassification}{7}
\newcommand{\bpNTypeRanking}{2}

\newcommand{\bpDimCCALo}{51}
\newcommand{\bpDimCCAHi}{53}
\newcommand{\bpDimFusedLo}{28}
\newcommand{\bpDimFusedHi}{34}
\newcommand{\bpDimFusedTwinLo}{24}
\newcommand{\bpDimFusedTwinHi}{31}
\newcommand{\bpFusedPcShareLo}{5}
\newcommand{\bpFusedPcShareHi}{10}

\newcommand{\bpFusedHeatTop}{29}
\newcommand{\bpFusedHeatMinN}{30}

\newcommand{\bpFusedPcTwoAcc}{$-$0.95}
\newcommand{\bpFusedPcTwoReward}{$-$0.92}
\newcommand{\bpFusedPcTwoArena}{$-$0.77}
\newcommand{\bpFusedPcOneArch}{0.57}
\newcommand{\bpFusedPcFourFamily}{0.72}
\newcommand{\bpFusedDateAcc}{0.56}
\newcommand{\bpTsnePerplexity}{30}
\newcommand{\bpDecTemp}{0.6}
\newcommand{\bpDecTopP}{0.95}
\newcommand{\bpDecTopK}{20}
\newcommand{\bpDecMaxTok}{3,072}
\newcommand{\bpDecMaxTokThink}{12,288}
\newcommand{\bpCoverageGate}{90}
\newcommand{\bpServeBudget}{24}
\newcommand{\bpFastFloorDraws}{20}
\newcommand{\bpCcaRidge}{0.1}

\newcommand{\codeurl}{\url{https://github.com/wcquesera/bullpen}}

\hypersetup{colorlinks=true,linkcolor=blue!60!black,citecolor=blue!60!black,urlcolor=blue!60!black}
\renewcommand{\shorttitle}{BULLPEN: Multi-Channel Model Embeddings}
\begin{document}

\maketitle

\begin{abstract}
Choosing among the hundreds of thousands of available large language models (LLMs) requires a thorough understanding of each candidate model's properties. In practice, each property is measured by its own benchmark suite, so profiling a model on many properties means paying for many separate evaluations. To address this, model embeddings compress each model into a fixed-length vector from which many properties can be predicted. Existing methods, such as EmbedLLM, build these vectors from correctness patterns and question features, and often need large probe batteries to do so. However, many behaviors that are important when choosing a model, such as its verbosity, safety and
style, show up more in how it answers than in whether it is correct. We introduce BULLPEN (Behavioral Understanding of Large Language models via Profile ENcodings), a framework that extends EmbedLLM in three ways. First, BULLPEN organizes the downstream tasks that prior work uses and adds \bpNNew{} task benchmarks we built on
behavior, style and provenance. Second, it is the
first framework to read a model's own answer text jointly with its correctness and the question text. Third,
it encodes models from a few hundred questions instead of the full \bpNPool-question pool. Across \bpNModels{} open-weight models, our answer-text
encoder scores \bpNewAnsFull{}\,$\pm$\,\bpNewAnsFullPm{} on the new tasks against \bpNewBestPrior{}\,$\pm$\,\bpNewBestPriorPm{} for the best prior encoder.
Our encoders that read answer text keep at least \bpPoolFusedFourHundred\% of their full-pool skill from only 400 questions (\bpCostShareFull\% of the inference compute of the
full question pool), while a correctness-only baseline keeps \bpPoolPcaBitsFourHundred\% with the same 400 questions. Because these embeddings need only model outputs, they support behavioral auditing and low-cost
profiling, and could extend to API-only models. Code and data: \codeurl.
\end{abstract}

\section{Introduction}

Hugging Face now hosts over 300,000 text-to-text models \citep{patel2026locus}, and more are released every week. Most of these models are derivatives of a few shared base models (via fine-tunes, merges, and quantizations), yet they differ widely in what
they do well, how they answer and where they came from. To choose among these models, users can consult published benchmark results, read the metadata attached to each model, or evaluate candidate models themselves \citep{hendrycks2021mmlu,chiang2024chatbotarena,liang2023helm,modellens2026}. However, published results often do not exist for a given model, metadata is usually limited to a short description and coarse labels such as architecture, size and license, and running one's own benchmarks on many candidate models is costly. Public leaderboards also score only a fraction of models and are rarely rerun as new ones appear (e.g., only \bpArenaRatedN{} of the \bpNModels{} models in this paper have an LMArena rating). To address this, a profile of every model can be built from its own outputs, relating it to the other models.

\begin{figure}[t]
  \centering
  \includegraphics[width=\linewidth]{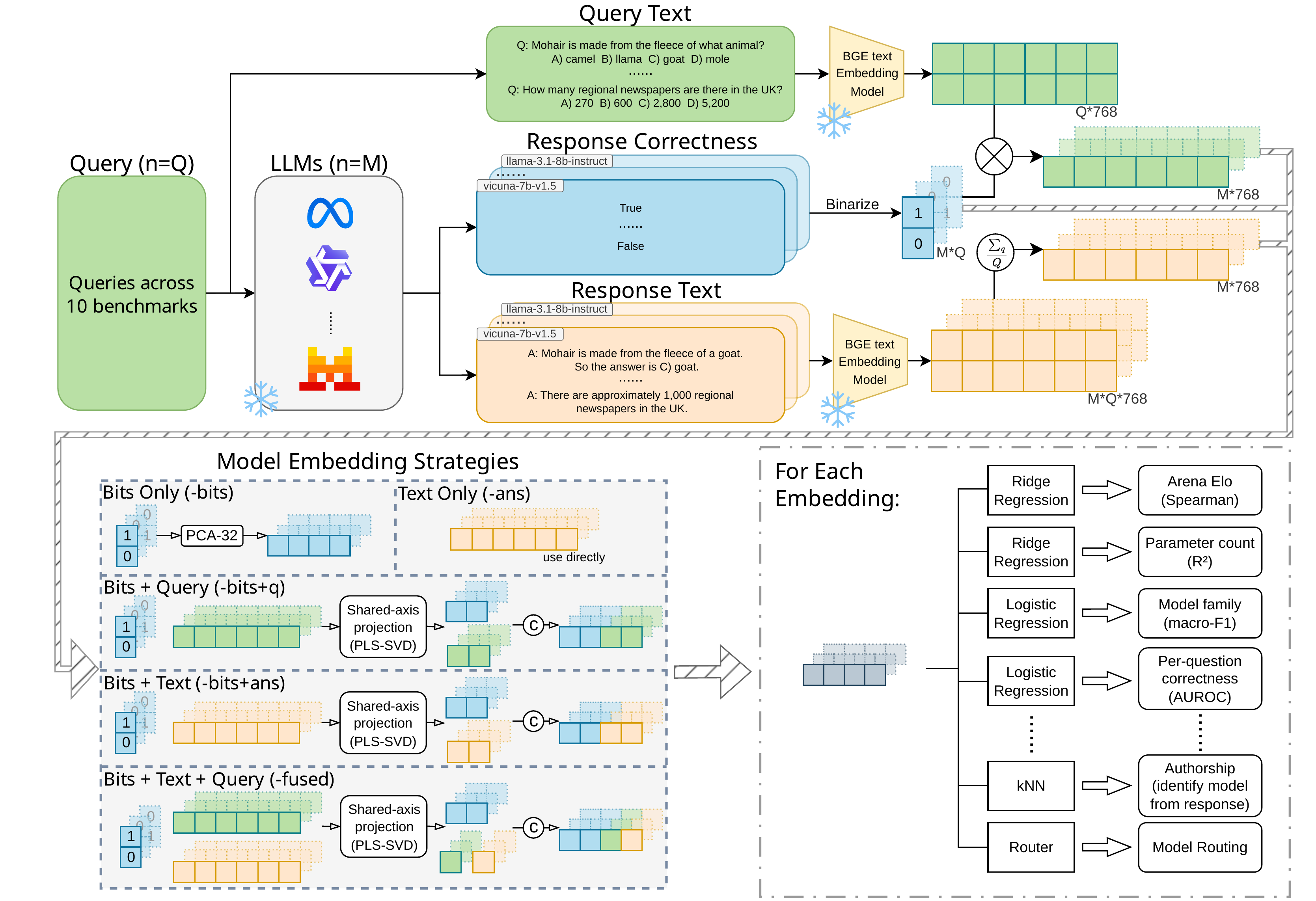}
  \caption{Overview of BULLPEN. Top: each LLM answers the probe questions, giving three channels (question text, correctness bits, answer text). Bottom left: the five embedding strategies. Bottom right: one prediction head per downstream task.}
  \label{fig:paper_overview}
\end{figure}
Model embeddings are one way to represent a model's properties, with one fixed-length vector per model. Embeddings have the advantage that model profiles can be compared with one another in a shared geometric space. A model \emph{encoder} produces the vector from the model's responses to a set of probe questions.  As established by
prior work \citep{zhuang2025embedllm}, we call each use of the vector a \emph{downstream task}, a
property of the model predicted from its embedding by a small prediction head. Existing encoders build this vector from how each model performs on a shared probe. For example,
EmbedLLM \citep{zhuang2025embedllm} learns the
vector by reconstructing whether each model answers each question of the probe correctly, with
an embedding of the question text as input, and evaluates it on correctness
forecasting, routing and benchmark prediction. LOCUS reads the same inputs with an attention encoder \citep{patel2026locus}. Embeddings based on item response theory (IRT) do the same with a model's ability and each question's difficulty \citep{chen2025irtnet,yao2026jeirt,song2025irtrouter}. These methods therefore describe a
model through two \emph{channels}, the \emph{correctness bits} (one record per probe question of
whether the model answered it correctly) and the \emph{question text}.

Current model encoders have two limitations. First, most read only the correctness bits and the
question text while leaving out what the model writes, which we call the \emph{answer text}. Holistic
evaluations argue that models should be compared on behaviors beyond accuracy \citep{liang2023helm},
and many of these behaviors are properties of the answers themselves. How long and stylized a
model's answers are can sway both human and LLM-judge ratings \citep{dubois2024lcalpacaeval,li2024arenastyle}.
How often a model refuses safe requests varies widely between models \citep{rottger2024xstest}. Even the
model family can be read from its answers alone \citep{sun2025idiosyncrasies}. An embedding built from
the answer text therefore has direct access to these properties, whereas one built from correctness
sees them only through their effect on accuracy. Fingerprinting methods
\citep{pasquini2025llmmap,wu2026llmdna} read the answer text but not correctness. No existing encoder combines the answer text with correctness and the question text, although the correctness bits are graded from these same answers, so every correctness-based encoder already pays for the answer text and then discards it. Second, these encoders are
evaluated almost entirely on competence and preference, through correctness forecasting, routing,
benchmark prediction and human ratings (Table~\ref{tab:paper_related_work}). Behavior and style are
not among these tasks, and provenance is shown mostly through clusters of similar models or lineage
trees \citep{zhuang2025embedllm,wu2026llmdna}, so no one has measured how much of these properties the channels capture.

\begin{table}[h!]
  \centering
  \small
  \setlength{\abovecaptionskip}{0pt}\setlength{\belowcaptionskip}{8pt}
  \caption{Model-embedding papers by input channels and task class.}
  \label{tab:paper_related_work}
  \begin{adjustbox}{max width=\linewidth}
\setlength{\tabcolsep}{2pt}
\begin{tabular}{>{\centering\arraybackslash}m{3.1cm}!{\color{black!20}\vrule}>{\centering\arraybackslash}m{2.2cm}>{\centering\arraybackslash}m{2.2cm}>{\centering\arraybackslash}m{2.2cm}!{\color{black!20}\vrule}>{\centering\arraybackslash}m{3.0cm}>{\centering\arraybackslash}m{2.2cm}>{\centering\arraybackslash}m{2.2cm}}
\toprule
& \multicolumn{3}{c}{\textit{behavior, style}} & \multicolumn{3}{c}{\textit{provenance}} \\
\cmidrule(lr){2-4} \cmidrule(lr){5-7}
& answer statistics & style effect & refusal, safety & identity, lineage & checkpoint specs & adoption \\
\textit{input channels} & {\scriptsize length, hedging, divergence} & {\scriptsize Arena style deltas, reward-model style} & {\scriptsize refusal rate, probe refusal (LLMmap)} & {\scriptsize authorship, family, architecture, base vs chat} & {\scriptsize size, license, release date, context} & {\scriptsize downloads, providers, quantizations} \\
\midrule
correctness + question text & BULLPEN & BULLPEN & BULLPEN & EmbedLLM, LOCUS, IrtNet, BULLPEN & BULLPEN & BULLPEN \\\arrayrulecolor{black!20}\hline\arrayrulecolor{black}
answer text & BULLPEN & BULLPEN & BULLPEN & LLMmap, LLM DNA, BULLPEN & BULLPEN & BULLPEN \\\arrayrulecolor{black!20}\hline\arrayrulecolor{black}
correctness + answer text & BULLPEN & BULLPEN & BULLPEN & BULLPEN & BULLPEN & BULLPEN \\\arrayrulecolor{black!20}\hline\arrayrulecolor{black}
correctness + question text + answer text & BULLPEN & BULLPEN & BULLPEN & BULLPEN & BULLPEN & BULLPEN \\
\bottomrule
\end{tabular}
  \end{adjustbox}
\end{table}

Here, we introduce BULLPEN (Behavioral Understanding of Large Language models via Profile ENcodings)
to address both limitations. BULLPEN treats one pass over the probe as a source of three channels,
namely the correctness bits, the question text and the model's own answer text. Our encoders read any subset of the three into one model embedding
(Figure~\ref{fig:paper_overview}). We also build \bpNNew{} downstream task benchmarks on behavior,
style and provenance from public leaderboards, model metadata and properties of the models' own answers
(Table~\ref{tab:paper_task_vocabulary}).

Our contributions are as follows.
\begin{enumerate}
\item[(1)] \textbf{A three-channel extension of EmbedLLM.} EmbedLLM embeds a model from its
correctness bits and the question text. BULLPEN adds the answer text from the same probe pass, and
its encoders can read any combination of the three channels.
\item[(2)] \textbf{Twenty downstream task benchmarks on behavior, style, and provenance.} We build
downstream tasks from public leaderboards, model metadata and properties of the models' own answers,
such as verbosity, refusal, Arena style ratings and authorship. 
\item[(3)] \textbf{Encoders that read answer text, and what that signal adds.} We represent the
answer text jointly with the correctness bits and the question text. Our answer-text encoders
outscore every prior encoder on the new tasks, most clearly on verbosity, refusal and authorship.
\item[(4)] \textbf{Profiling from a few hundred questions.} With 400 questions, our encoders that read
answer text keep at least \bpPoolFusedFourHundred\% of the skill they reach with the whole
\bpNPool-question pool, which would cost about \bpCostRatioFull{} times as much to collect.
\end{enumerate}

\section{Related work}
\label{sec:related}

Four lines of work are closest to BULLPEN.

\begin{table}[h!]
  \centering
  \small
  \setlength{\abovecaptionskip}{0pt}\setlength{\belowcaptionskip}{8pt}
  \caption{The \bpNNew{} new tasks, by class.}
  \label{tab:paper_task_vocabulary}
  \begin{adjustbox}{max width=\linewidth}
\begin{tabular}{>{\raggedright\arraybackslash}p{2.6cm}>{\raggedright\arraybackslash}p{4.6cm}>{\raggedright\arraybackslash}p{6.3cm}}
\toprule
class & what it measures & tasks \\
\midrule
\multicolumn{3}{l}{\textbf{behavior, style} (9 tasks)} \\
answer statistics\newline(3 tasks) & statistics of the model's own answers: length, hedging, divergence from other models & answer length; hedging; answer divergence$^{\ddagger}$ \\ \addlinespace[3pt]
style effect\newline(4 tasks) & how much a rating owes to answer style: LMArena's style-control deltas and the reward-model premium given correctness & style delta (LMArena); style delta, creative (LMArena); style delta, math (LMArena); reward style (Skywork-RM) \\ \addlinespace[3pt]
refusal, safety\newline(2 tasks) & whether a model refuses (our probe, LLMmap's probes) & refusal rate; probe refusal rate (LLMmap data) \\ \addlinespace[3pt]
\midrule
\multicolumn{3}{l}{\textbf{provenance} (11 tasks)} \\
identity, lineage\newline(4 tasks) & which model, family or lineage a checkpoint is; who wrote an answer & authorship$^{\dagger}$; architecture (HF)$^{\dagger}$; model family (HF)$^{\dagger}$; base vs chat (HF)$^{\ddagger}$ \\ \addlinespace[3pt]
checkpoint specs\newline(4 tasks) & facts of the checkpoint on Hugging Face: license, size, context, release date & license (HF); release date (HF); context length (HF); parameter count (HF)$^{\ddagger}$ \\ \addlinespace[3pt]
adoption\newline(3 tasks) & community uptake: downloads, serving providers, Ollama quantizations & downloads (HF); serving providers (HF); Ollama quantizations \\ \addlinespace[3pt]
\bottomrule
\end{tabular}
  \end{adjustbox}
  \par\smallskip
  \parbox{\linewidth}{\scriptsize $^{\dagger}$: notion from model fingerprinting \citep{pasquini2025llmmap,wu2026llmdna,sun2025idiosyncrasies}, posed here as prediction from an embedding, with our own labels; $^{\ddagger}$: partial precedent. Established tasks: Table~\ref{tab:paper_task_established}.}
\end{table}

\begin{table}[t]
  \centering
  \small
  \setlength{\abovecaptionskip}{0pt}\setlength{\belowcaptionskip}{8pt}
  \caption{Mean skill over the width-matched floor on the \bpNNew{} new tasks, by class.}
  \label{tab:paper_master_table}
  \begin{adjustbox}{max width=\linewidth}
\begin{tabular}{l!{\color{black!25}\vrule}rrr!{\color{black!25}\vrule}rrr!{\color{black!25}\vrule}rr}
\toprule
 & \multicolumn{3}{c}{\textit{behavior, style}} & \multicolumn{3}{c}{\textit{provenance}} &  &  \\
\cmidrule(lr){2-4} \cmidrule(lr){5-7}
 & \rotatebox{90}{answer statistics} & \rotatebox{90}{style effect} & \rotatebox{90}{refusal, safety} & \rotatebox{90}{identity, lineage} & \rotatebox{90}{checkpoint specs} & \rotatebox{90}{adoption} & \rotatebox{90}{\textbf{all tasks}} & \rotatebox{90}{$M_{\text{eff}}$-weighted} \\
\multicolumn{1}{r}{\textit{tasks per column}} & 3 & 4 & 2 & 4 & 4 & 3 & 20 & 15.2 (eff.) \\
\midrule
\textbf{embedding} \\
\multicolumn{9}{l}{\textit{reference: one number per model}} \\
\quad leaderboard scalar (d=1) & 0.43 & \cellcolor{green!5}\textbf{0.57} & -0.02 & 0.13 & 0.24 & 0.16 & 0.27$_{\pm0.12}$ & 0.25$_{\pm0.12}$ \\
\midrule
\textit{bits} & {\color{black!55}\textit{0.33$_{\pm0.22}$}} & {\color{black!55}\textit{0.40$_{\pm0.11}$}} & {\color{black!55}\textit{0.13$_{\pm0.09}$}} & {\color{black!55}\textit{0.20$_{\pm0.11}$}} & {\color{black!55}\textit{0.27$_{\pm0.18}$}} & {\color{black!55}\textit{0.19$_{\pm0.04}$}} & {\color{black!55}\textit{0.27$_{\pm0.07}$}} & {\color{black!55}\textit{0.25$_{\pm0.07}$}} \\
\quad PCA32-bits-budget & \cellcolor{red!6}0.28 & 0.28 & 0.36 & 0.22 & 0.27 & 0.21 & 0.26$_{\pm0.08}$ & 0.26$_{\pm0.09}$ \\
\quad PCA32-bits & 0.36 & 0.34 & 0.02 & 0.27 & \cellcolor{green!5}\textbf{0.31} & 0.22 & 0.27$_{\pm0.07}$ & 0.26$_{\pm0.07}$ \\
\quad IRT-2PL & 0.34 & \cellcolor{green!5}\textbf{0.60} & 0.03 & \cellcolor{red!6}0.12 & \cellcolor{red!6}0.24 & \cellcolor{red!6}0.16 & 0.27$_{\pm0.11}$ & \cellcolor{red!6}0.24$_{\pm0.11}$ \\
\midrule
\textit{bits + question text} & {\color{black!55}\textit{0.38$_{\pm0.21}$}} & {\color{black!55}\textit{0.43$_{\pm0.15}$}} & {\color{black!55}\textit{0.17$_{\pm0.12}$}} & {\color{black!55}\textit{0.22$_{\pm0.12}$}} & {\color{black!55}\textit{0.33$_{\pm0.17}$}} & {\color{black!55}\textit{0.22$_{\pm0.07}$}} & {\color{black!55}\textit{0.30$_{\pm0.08}$}} & {\color{black!55}\textit{0.29$_{\pm0.09}$}} \\
\quad EmbedLLM (retrained) & 0.40 & \cellcolor{green!5}\textbf{0.52} & 0.18 & 0.23 & 0.33 & \cellcolor{green!5}\textbf{0.25} & 0.33$_{\pm0.08}$ & 0.32$_{\pm0.09}$ \\
\quad LOCUS (retrained) & 0.41 & 0.39 & 0.02 & 0.19 & 0.31 & 0.21 & 0.27$_{\pm0.09}$ & 0.27$_{\pm0.09}$ \\
\quad RouteLLM-MF (retrained) & 0.41 & \cellcolor{green!5}\textbf{0.52} & 0.06 & 0.19 & 0.28 & 0.23 & 0.30$_{\pm0.09}$ & 0.28$_{\pm0.10}$ \\
\quad IRT-Router (retrained) & 0.36 & 0.39 & \cellcolor{red!6}-0.04 & 0.20 & \cellcolor{green!5}\textbf{0.34} & 0.20 & 0.27$_{\pm0.09}$ & 0.26$_{\pm0.09}$ \\
\quad JE-IRT (retrained) & 0.41 & \cellcolor{green!5}\textbf{0.60} & 0.08 & 0.23 & \cellcolor{green!5}\textbf{0.37} & 0.24 & 0.35$_{\pm0.09}$ & 0.32$_{\pm0.09}$ \\
\quad IrtNet (retrained) & 0.38 & 0.41 & 0.23 & 0.22 & 0.33 & 0.17 & 0.30$_{\pm0.08}$ & 0.29$_{\pm0.09}$ \\
\quad BULLPEN-bits+q-budget & 0.35 & 0.34 & 0.41 & 0.25 & \cellcolor{green!5}\textbf{0.33} & 0.20 & 0.31$_{\pm0.08}$ & 0.30$_{\pm0.09}$ \\
\quad BULLPEN-bits+q-full & 0.33 & 0.30 & \cellcolor{green!5}\textbf{0.42} & 0.28 & \cellcolor{green!5}\textbf{0.33} & 0.24 & 0.31$_{\pm0.08}$ & 0.31$_{\pm0.09}$ \\
\midrule
\textit{answer text} & {\color{black!55}\textit{0.68$_{\pm0.06}$}} & {\color{black!55}\textit{0.49$_{\pm0.08}$}} & {\color{black!55}\textit{0.29$_{\pm0.08}$}} & {\color{black!55}\textit{0.28$_{\pm0.12}$}} & {\color{black!55}\textit{0.31$_{\pm0.18}$}} & {\color{black!55}\textit{0.29$_{\pm0.08}$}} & {\color{black!55}\textit{0.39$_{\pm0.08}$}} & {\color{black!55}\textit{0.38$_{\pm0.10}$}} \\
\quad LLMmap (retrained) & 0.39 & \cellcolor{red!6}0.15 & 0.15 & 0.19 & 0.27 & 0.24 & \cellcolor{red!6}0.24$_{\pm0.07}$ & 0.25$_{\pm0.07}$ \\
\quad LLM DNA (recomputed) & 0.67 & \cellcolor{green!5}\textbf{0.55} & 0.03 & 0.22 & 0.29 & 0.27 & 0.35$_{\pm0.10}$ & 0.34$_{\pm0.11}$ \\
\quad BULLPEN-ans-budget & \cellcolor{green!5}\textbf{0.82} & \cellcolor{green!5}\textbf{0.62} & \cellcolor{green!5}\textbf{0.41} & \cellcolor{green!5}\textbf{0.35} & \cellcolor{green!5}\textbf{0.35} & \cellcolor{green!5}\textbf{0.32} & \cellcolor{green!5}\textbf{0.47}$_{\pm0.10}$ & \cellcolor{green!5}\textbf{0.46}$_{\pm0.11}$ \\
\quad BULLPEN-ans-full & \cellcolor{green!5}\textbf{0.82} & \cellcolor{green!5}\textbf{0.64} & \cellcolor{green!5}\textbf{0.46} & \cellcolor{green!5}\textbf{0.35} & \cellcolor{green!5}\textbf{0.32} & \cellcolor{green!5}\textbf{0.32} & \cellcolor{green!5}\textbf{0.48}$_{\pm0.10}$ & \cellcolor{green!5}\textbf{0.47}$_{\pm0.11}$ \\
\midrule
\textit{bits + answer text} & {\color{black!55}\textit{0.70$_{\pm0.02}$}} & {\color{black!55}\textit{0.45$_{\pm0.14}$}} & {\color{black!55}\textit{0.34$_{\pm0.14}$}} & {\color{black!55}\textit{0.32$_{\pm0.11}$}} & {\color{black!55}\textit{0.34$_{\pm0.20}$}} & {\color{black!55}\textit{0.30$_{\pm0.09}$}} & {\color{black!55}\textit{0.41$_{\pm0.09}$}} & {\color{black!55}\textit{0.40$_{\pm0.09}$}} \\
\quad BULLPEN-bits+ans-budget & 0.70 & 0.45 & 0.40 & 0.30 & \cellcolor{green!5}\textbf{0.33} & \cellcolor{green!5}\textbf{0.28} & 0.40$_{\pm0.09}$ & 0.40$_{\pm0.10}$ \\
\quad BULLPEN-bits+ans-full & 0.70 & 0.45 & 0.29 & \cellcolor{green!5}\textbf{0.34} & \cellcolor{green!5}\textbf{0.35} & \cellcolor{green!5}\textbf{0.32} & 0.41$_{\pm0.08}$ & 0.41$_{\pm0.09}$ \\
\midrule
\textit{bits + question text + answer text} & {\color{black!55}\textit{0.70$_{\pm0.00}$}} & {\color{black!55}\textit{0.48$_{\pm0.13}$}} & {\color{black!55}\textit{0.39$_{\pm0.21}$}} & {\color{black!55}\textit{0.31$_{\pm0.11}$}} & {\color{black!55}\textit{0.35$_{\pm0.19}$}} & {\color{black!55}\textit{0.29$_{\pm0.09}$}} & {\color{black!55}\textit{0.42$_{\pm0.09}$}} & {\color{black!55}\textit{0.41$_{\pm0.10}$}} \\
\quad BULLPEN-fused-budget & 0.69 & 0.48 & 0.33 & 0.30 & \cellcolor{green!5}\textbf{0.33} & \cellcolor{green!5}\textbf{0.28} & 0.40$_{\pm0.09}$ & 0.40$_{\pm0.10}$ \\
\quad BULLPEN-fused-full & 0.71 & 0.49 & \cellcolor{green!5}\textbf{0.45} & \cellcolor{green!5}\textbf{0.33} & \cellcolor{green!5}\textbf{0.36} & \cellcolor{green!5}\textbf{0.30} & \cellcolor{green!5}\textbf{0.43}$_{\pm0.09}$ & \cellcolor{green!5}\textbf{0.43}$_{\pm0.10}$ \\
\bottomrule
\end{tabular}
  \end{adjustbox}
  \par\smallskip
  \parbox{\linewidth}{\scriptsize Rows grouped by input channels (q: question text, ans: answer text); -budget: the $K$=400 questions. Bold and green: tied with the column's best (95\% interval); underlined: sole best; red: column minimum; gray rows: group means. $M_{\text{eff}}$-weighted: tasks weighted by their share of the effective number of independent tasks \citep{li2005meff}. LLMmap: \bpMapOwnN{} of \bpNModels{} models answered its 8 probes, the rest take the mean signature; its probe-refusal cell is omitted. Scoring rules: Appendix~\ref{app:master}.}
\end{table}

\paragraph{Model embeddings and routing.} Sending each input to the right model can cut cost and
raise quality. Choosing the model most likely to answer a query well, often under a cost limit
\citep{hu2024routerbench}, is a case of algorithm selection \citep{rice1976algorithmselection}. Some routers send each query to either a strong or a cheaper model, depending on how hard the query
looks \citep{ding2024hybridllm,ong2025routellm}. Others,
such as FrugalGPT \citep{chen2023frugalgpt}, try cheaper models first and move to a stronger one only
when the answer looks unreliable. A router needs a profile of each candidate model, and several build it explicitly from
past performance \citep{wang2026iclrouter,xu2026routeprofile,okamoto2026bella,li2026flyroute}.
EmbedLLM \citep{zhuang2025embedllm} turned such profiles into reusable model embeddings, learned from
correctness with the question text as side information. LOCUS \citep{patel2026locus} encodes the same
inputs with an attention encoder that embeds a new model from a few queries. Embeddings based on item response theory (IRT), such as IrtNet, JE-IRT and IRT-Router, fit each
model's ability and each question's difficulty to the correctness bits, and predict a question's
difficulty from its text \citep{birnbaum1968irt,chen2025irtnet,yao2026jeirt,song2025irtrouter}. Other work extracts more signal from correctness, such as the
wrong answer a model picks \citep{fei2026everywrong} or how much models' mistakes overlap
\citep{goel2025capa}. These methods evaluate their embeddings mostly on correctness forecasting, routing and benchmark
prediction. 

\paragraph{Efficient evaluation.} Running every model on every benchmark is costly, so several
methods evaluate a model from a small, well-chosen set of questions. Anchor Points and tinyBenchmarks pick representative questions whose answers estimate full benchmark
scores \citep{vivek2024anchorpoints,polo2024tinybenchmarks}, and core-set methods select prompt subsets for the
same purpose \citep{yao2026coresets}. Model embeddings face the same cost. LOCUS embeds a new model from
a few queries \citep{patel2026locus}, and LLMmap identifies a model from a handful of crafted prompts
\citep{pasquini2025llmmap}. We also ask each model only a few hundred questions, but use its answers
to build the whole embedding, answer text included.

\paragraph{Model fingerprinting.} Fingerprinting methods also build a representation of each model
from its outputs, and use it to recognize which model produced a piece of text. LLMmap \citep{pasquini2025llmmap}
identifies which model sits behind an interface from its answers to a few crafted queries, and
LLM DNA \citep{wu2026llmdna} derives a representation from a model's outputs to relate models. Behavioral fingerprinting \citep{behavioral2025fingerprinting} describes models
through diagnostic prompts judged by another model, and \citet{sun2025idiosyncrasies} show that a
classifier can tell model families apart from their outputs alone. These methods read only the answer text.

\paragraph{Style and behavior evaluation.} Accuracy misses much of what users value, so holistic
evaluations compare models on more than accuracy \citep{liang2023helm}. Chatbot Arena turns pairwise
human votes into ratings \citep{chiang2024chatbotarena}, AlpacaEval controls its win rates for answer
length \citep{dubois2024lcalpacaeval}, and Arena's style-controlled ratings remove the effect of
length and formatting by regression \citep{li2024arenastyle}. Reward models score answers
directly \citep{lambert2025rewardbench}, and XSTest measures how often models refuse safe
requests \citep{rottger2024xstest}.

\section{BULLPEN}
\label{sec:method}

\begin{figure}[t]
  \centering
  \includegraphics[width=\linewidth]{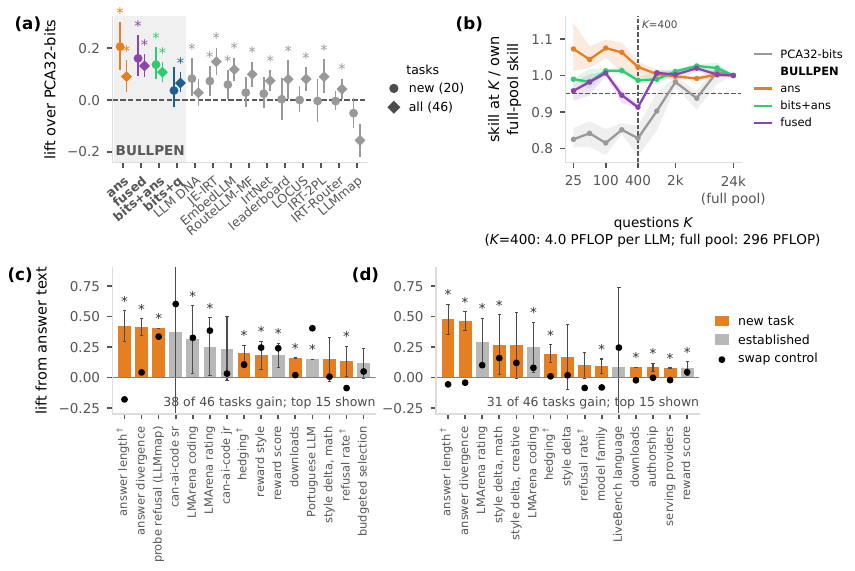}
  \caption{Answer-text lift and question budget. (a) Lift over PCA32-bits on the new and all tasks; $*$: 95\% interval above zero; gray: prior encoders. (b) Skill at $K$ questions over each encoder's own full-pool skill. (c, d) Per-task lift from adding answer text to the bits (BULLPEN-bits+ans $-$ PCA32-bits) and to bits+q (BULLPEN-fused $-$ BULLPEN-bits+q); dots: swap control.}
  \label{fig:paper_task_panels}
\end{figure}

\subsection{Problem} Our goal is to predict many properties of a language model from one pass over a fixed set of questions. We are given models $\mathcal{M}$ and a \emph{probe} $\mathcal{Q}$, a fixed set of questions that every model answers once. An \emph{encoder} $E$ maps each model $m$ to an embedding $z_m\in\mathbb{R}^d$ using any of three channels, namely which questions the model answers correctly, the questions themselves, and the text it writes. We use \emph{model} for the language models being embedded and \emph{encoder} for the methods that embed them. Each downstream task $i$ is a property $y_i(m)$, predicted from $z_m$ by a small head trained on some models and scored on held-out ones. A good encoder scores well across many such tasks, measured as skill over random embeddings of the same width (Eq.~\ref{eq:skill-over-floor}), while it reads only $K$ of the probe questions per model. \par \subsection{Data}
We use a \bpNProbe{}-question subset of the EmbedLLM probe \citep{zhuang2025embedllm}, drawn from ten
benchmarks: MMLU \citep{hendrycks2021mmlu}, GSM8K \citep{cobbe2021gsm8k}, GPQA
\citep{rein2024gpqa}, MathQA \citep{amini2019mathqa} and others. We collected answers to this probe
from \bpNModels{} open-weight models (Appendix~\ref{app:models}). Each model answers each question once,
with greedy decoding for most models and sampling for reasoning models, and a rule-based grader marks
each answer correct or incorrect (Appendix~\ref{app:encoders}).

We split the questions once, keeping each benchmark's share the same in every part. The encoders read
the \emph{pool} of \bpNPool{} questions, a smaller set is kept for tuning, and a held-out evaluation set
supplies the labels that we compute from the models' answers (Appendix~\ref{app:encoders} gives the
sizes). No encoder reads an evaluation question.

\subsection{Channels}
Each model has three \emph{channels}, all from one pass over the probe. We write $\mathcal{P}$ for the pool questions that model $m$ answered. Its \emph{correctness
bit} $b_{mq}$ is 1 if it answered question $q$ correctly and 0 otherwise. Its \emph{answer text} is
the average of its answers after an off-the-shelf text-embedding model, bge-base-en-v1.5, turns each answer into 768 numbers
\citep{xiao2023bge}.
The \emph{question text} is the same for every model, so it can describe a model only through the
questions the model gets right. We embed each question $q$ as $u_q$ with the same model and
weight it by $b_{mq}-\bar b_q$, where $\bar b_q$ is the share of models that answer $q$ correctly,
\begin{equation}\label{eq:qchannel}
q_m=\frac{1}{|\mathcal{P}|}\sum_{q\in\mathcal{P}}\big(b_{mq}-\bar b_q\big)\,u_q .
\end{equation}
The vector $q_m$ therefore points towards the questions on which the model beats the average model.

\subsection{Question selection}\label{sec:budget}
To keep the cost per model low, BULLPEN embeds each model from $K$ fixed questions. We fit a standard
two-parameter IRT model to the training models' correctness bits and keep the $K$ pool questions with
the highest Fisher information, which are the questions that best separate strong models from weak
ones (Appendix~\ref{app:encoders}). An encoder marked \emph{-full} reads all pool questions that every
model answered, and one marked \emph{-budget} reads only the $K$ selected ones. 

\subsection{Encoders}\label{sec:encoders}
An encoder maps a model's outputs to its model embedding $z_m$. The main text reports one encoder
per set of channels (Appendix~\ref{app:encoders} lists the other combinations). Each encoder outputs a
32-dimensional embedding, except BULLPEN-ans, which keeps all 768 dimensions of the sentence encoder. \emph{PCA32-bits}, the
correctness-only baseline, keeps 32 principal components of the matrix of correctness bits.
\emph{BULLPEN-ans} is the mean of the model's answer embeddings, scaled to length one. As a control, we compare each encoder that reads answer text against its \emph{swap control}, the same
encoder given another model's answers instead, which measures how much of its skill comes from the
model's own answers.

The joint encoders BULLPEN-bits+q, BULLPEN-bits+ans and BULLPEN-fused compute the covariance between
the channels they read, and keep for each channel the directions that vary most with the others, using
multi-block partial least squares SVD (PLS-SVD; cf.\ \citealt{tucker1958interbattery,westerhuis1998mbpls}). With one row per
model in a table $X_k$ for each channel $k$,
\begin{equation}\label{eq:pls}
W_k=\mathrm{svd}_{\mathrm{left}}\big(X_k^{\top}\,[\,\tilde X_j\,]_{j\neq k}\big),\qquad
z_m=\big[\,z(x_{m,1}W_1)\;\cdots\;z(x_{m,J}W_J)\,\big].
\end{equation}
Here each other channel $\tilde X_j$ is scaled to unit Frobenius norm, so the correctness bits do not
dominate the signal from the other channels (for the -full encoders the bits have far more columns than
the others), and $z(\cdot)$ standardizes each channel's scores.

\subsection{Heads and score}\label{sec:score}
For each downstream task (Section~\ref{sec:tasks}), a
prediction head predicts the model's property from its embedding. Most heads are linear (i.e., a ridge regression for numeric labels and a logistic regression for
categorical ones). Each prediction head is trained on a subset of the models and scored on the remaining held-out
models. Each task $i$ gets a normalized score $s_i$, the task's metric (such as a rank correlation)
rescaled so that 0 is chance and 1 is a perfect prediction. With many embedding dimensions and few models, a head can also fit patterns by luck, so we subtract the average score of random
Gaussian embeddings of the same width, the \emph{floor}. This also makes encoders of different sizes
comparable. We report \emph{skill over floor},
\begin{equation}\label{eq:skill-over-floor}
\Delta_i(E)=s_i(\text{embeddings of }E)-s_i(\text{random embeddings of the same width}),
\end{equation}
so 0 means no gain over random numbers.
Our summaries average $\Delta_i$ over tasks. Because some tasks rank models alike,
Table~\ref{tab:paper_master_table} also gives a weighted mean in which each group of such tasks counts
as about one task. The weights come from the effective number of independent tasks, which we
estimate from the eigenvalues of the task-by-task correlation matrix \citep{li2005meff}.

\section{Downstream tasks}
\label{sec:tasks}

The tasks test both halves of our claim, that answer text adds properties correctness misses and that it keeps what correctness carries. The \emph{new} tasks cover behavior, style and provenance, which practitioners choose models by but prior embeddings were not evaluated on (Table~\ref{tab:paper_task_vocabulary}). The \emph{established} tasks are of the kinds prior embeddings are evaluated on, and check that the correctness signal survives. No label is computed from anything an encoder reads.

\paragraph{Label sources.} Each task's labels describe the models from one source. Leaderboard
scores and preference ratings come from public boards, provenance labels come from each model's
Hugging Face page and configuration file, and behavior labels are statistics of the models' own
answers to the evaluation questions. We match each source to our models by name and drop boards and page fields that are not well represented in our set of models (Table~\ref{tab:paper_task_excluded} in
Appendix~\ref{app:tasks}, which also gives each task's source and metric).

\paragraph{Behavior and style.} These tasks ask how a model answers. Some labels are measured on
the model's own answers to the evaluation questions, such as answer length. Hedging counts
uncertainty words such as ``I think'', and refusal rate counts explicit declines such as ``I cannot''.
Answer divergence measures how far a model's answers are from other models' answers to the same
questions. Style effects measure how much of a rating comes from the look of the
answers, such as the gap between LMArena's ratings \citep{lmarena2025leaderboard} with and without a control for length and formatting
\citep{li2024arenastyle}. Another measures how much a reward model prefers a model's answers beyond
their correctness.

\paragraph{Provenance.} These tasks ask what a checkpoint is and how others use it. Most labels come
from each model's Hugging Face page and configuration file, such as its architecture, family, license
and parameter count. Others count downloads, serving providers and compressed versions of the model.
Following fingerprinting work, the model-family task asks a head to name a held-out model's family \citep{pasquini2025llmmap,sun2025idiosyncrasies}. The authorship task asks whether a
model can be recognized from its writing, by checking that the answer statistics its embedding
predicts lie closest to its own.

\paragraph{Competence and preference.} The established tasks (Table~\ref{tab:paper_task_established}
in Appendix~\ref{app:tasks}) predict accuracy on held-out probe questions, published benchmark scores,
and human or judge ratings such as LMArena's. They also include model selection, which routes each
question to the model predicted to answer it best \citep{zhuang2025embedllm,chen2025irtnet,patel2026locus}.

\section{Results}
\label{sec:protocol}

\begin{figure}[t]
  \centering
  \includegraphics[width=\linewidth]{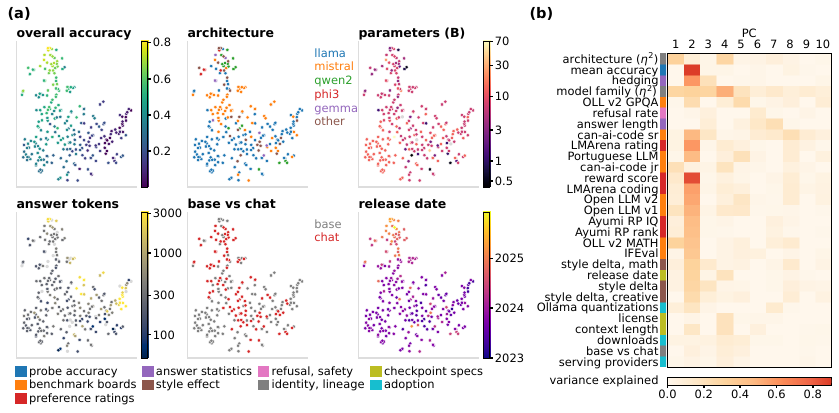}
  \caption{Structure of BULLPEN-fused-full over the \bpNModels{} models. (a) A t-SNE map colored by six model properties; gray points lack the property. (b) The share of each task's variance each principal component explains ($\rho^2$, Spearman; $\eta^2$ for architecture and model family) over the \bpFusedHeatTop{} tasks labeled on at least \bpFusedHeatMinN{} models, the first ten matched one to one to the components (Hungarian matching), the rest under the component that explains most of their variance (strip: task class); a component's sign is arbitrary.}
  \label{fig:paper_embedding_structure}
\end{figure}

\paragraph{Protocol.} Every number in this section is measured on held-out models. We split the models
into five folds, keeping every model released by the same organization (for example, all models from
one company's Hugging Face account) in the same fold, and refit the encoder and every prediction head
without the predicted fold. We compare against eight prior encoders that cover each set of channels prior work reads. EmbedLLM, LOCUS, RouteLLM-MF and three IRT encoders read correctness and the question text, and LLMmap and LLM DNA read answer text. A single leaderboard score and PCA32-bits show how far a trivial profile gets. We rebuild each on our data (Appendix~\ref{app:encoders}).

\subsection{Behavior, style and provenance tasks}
\label{sec:res-external}

Answer text carries most of the signal for behavior, style and provenance. Encoders that read it lead on the \bpNNew{} new tasks (Table~\ref{tab:paper_master_table}).
BULLPEN-ans-full has a score of \bpNewAnsFull{}\,$\pm$\,\bpNewAnsFullPm{} while the best prior encoder
only achieves \bpNewBestPrior{}\,$\pm$\,\bpNewBestPriorPm, and the \bpNewOursTop{} highest-scoring
encoders are all BULLPEN encoders that read answer text. Over the new tasks, BULLPEN-ans-full improves
on PCA32-bits by \bpSigTextBitsNew{} ([\bpSigTextBitsNewLo, \bpSigTextBitsNewHi]), the largest lift of
any encoder (Figure~\ref{fig:paper_task_panels}a). The largest gains are on answer length, answer
divergence, refusal rate and authorship (Table~\ref{tab:paper_master_wide} in
Appendix~\ref{app:master}). 

The swap control shows that these gains come from each model's own answers, and not from having any
coherent text embedding as input. When BULLPEN-ans-full is given another model's answers instead of
its own, its skill falls by \bpTwinAnsDrop{} ([\bpTwinAnsDropLo, \bpTwinAnsDropHi]), from
\bpAllAnsFull{} to about zero, the level of a random embedding
(Table~\ref{tab:paper_control_significance} in Appendix~\ref{app:controls}). BULLPEN-fused's
performance also falls, but notably much less, by \bpTwinFusedDrop{} ([\bpTwinFusedDropLo,
\bpTwinFusedDropHi]), due to the signal from its other input channels.

\subsection{Competence and preference tasks}
\label{sec:res-battery}

Reading answer text keeps most of the correctness signal that prior work measures. The \bpNEst{} established tasks check this (Table~\ref{tab:paper_master_wide} in
Appendix~\ref{app:master}). Encoders that read correctness lead here, with JE-IRT highest at
\bpEstJeIrt{}\,$\pm$\,\bpEstJeIrtPm. BULLPEN-fused-full reaches \bpEstFusedFull{}\,$\pm$\,\bpEstFusedFullPm,
level with IrtNet (\bpEstIrtNet{}\,$\pm$\,\bpEstIrtNetPm) and above PCA32-bits
(\bpEstPcaBits{}\,$\pm$\,\bpEstPcaBitsPm). On human preference, the leaderboard score alone matches the best encoders (\bpPrefScalar{} against \bpPrefJeIrt{} for JE-IRT and IRT-2PL).

\subsection{Channel combinations and embedding structure}
\label{sec:res-joint}

Each channel wins a different kind of task (Tables~\ref{tab:paper_master_table} and~\ref{tab:paper_master_wide}). The correctness bits and the question text carry competence, which the established tasks
reward, and the answer text carries behavior, style and provenance, which the new tasks reward.
BULLPEN-fused-full, which reads all three channels, keeps both. It is second on the new tasks with
\bpNewFusedFull{}\,$\pm$\,\bpNewFusedFullPm, behind only the BULLPEN-ans encoders. Over all \bpNAll{} tasks it ties the best-performing encoder, JE-IRT, with \bpAllFusedFull{} against
\bpAllJeIrt{} (margin \bpSigLeadAll, [\bpSigLeadAllLo, \bpSigLeadAllHi], an interval that includes zero). 

We examine the structure of the BULLPEN-fused space with t-SNE and principal component analysis (PCA). In the t-SNE projection, which preserves local neighborhood
relationships, accuracy follows a continuous gradient along a single direction. Architecture and
training stage (base versus chat), by contrast, form distinct regions that run across this gradient
(Figure~\ref{fig:paper_embedding_structure}a). The PCA panel gives the fraction of variance in each property explained by each of the first ten principal components (PCs) (Figure~\ref{fig:paper_embedding_structure}b).
Competence-related measures load predominantly on PC2, which explains most of the variance in mean
accuracy and reward score, and benchmark and preference measures such as LMArena rating and Open LLM
Leaderboard scores follow the same pattern. Lineage variables load instead on PC1 and PC4, which
together explain the largest share of the variance in architecture and model family
(Appendix~\ref{app:embedding}). Competence is thus one axis, and lineage and architecture occupy others.

\subsection{Sample efficiency}
\label{sec:res-budget}

At 400 of the \bpNPool{} pool questions, which costs \bpCostShareFull\% of the inference compute of the
whole pool, BULLPEN-fused keeps \bpPoolFusedFourHundred\% (\bpPoolFusedFourHundredLo--\bpPoolFusedFourHundredHi\%
across seeds) of the skill it reaches with the whole pool, while the correctness-only PCA32-bits keeps
\bpPoolPcaBitsFourHundred\% (\bpPoolPcaBitsFourHundredLo--\bpPoolPcaBitsFourHundredHi\%)
(Figure~\ref{fig:paper_task_panels}b). Each answer carries more information than one
correct-or-incorrect bit, so the encoders that read answer text need fewer questions to profile a
model than encoders that use only correctness bits.

\section{Limitations}
First, we have a limited number of models, and public leaderboards label only some of
them. Small differences, such as BULLPEN-fused-full against JE-IRT over all tasks, therefore cannot yet
be separated. More models with full profiles would sharpen these comparisons and let us try more complex encoders and heads. Second, every answer-text result rests on one text-embedding
model (BGE) and one set of generation settings, and stronger text encoders may capture more of the
answer-text signal. Lastly, every model we embed is open-weight and answers single-turn questions, but this work does not profile closed API models. Because BULLPEN needs only a model's answers, our future work includes
adding API models to our model set, and extending the questions to multi-turn and agentic behavior. Appendix~\ref{app:further-lim} lists further limitations.

\section{Conclusion}

BULLPEN extends model embeddings with each model's own answer text, read alongside its correctness
bits and the question text. On new tasks about behavior, style and provenance, embeddings that read answer text predict
properties that correctness-based embeddings miss. One embedding that reads all three channels is second only to the answer-text encoders on these tasks and ties the best-performing encoder over all tasks.
Because a few hundred questions recover most of this signal, BULLPEN makes it practical to profile a
large and growing pool of models by how they answer as well as by what they get right.

\subsection*{AI use statement}
We used LLM coding agents to assist in writing and testing the code for data collection,
processing and all experiments, in giving feedback on the experimental design and checking our
interpretation of the results, in finding and checking related literature, and in drafting and
editing the manuscript. Some prediction targets are
scores assigned by reward models or LLM judges (Appendix~\ref{app:tasks}). The authors reviewed all
AI-assisted work, verified the code with tests, and checked every claim, number and citation in the
text. We take full responsibility for the content of this paper.

\subsection*{Ethics statement}
The data consist of answers of public open-weight models to public benchmarks, together with
public leaderboard records. No human subjects were involved.

\subsection*{Reproducibility statement}
Code and configuration are available at \codeurl, and the repository links to the data. Section~\ref{sec:method}
and Appendix~\ref{app:encoders} define the question split, the encoders and the score.
Appendix~\ref{app:tasks} gives each task's source, join, coverage and metric, and
Appendix~\ref{app:encoders} also gives the decoding settings and quality gates for the collected
answers and the retraining settings of every competitor.

\subsubsection*{Acknowledgments}
R.R. is supported by the National Science Foundation Graduate Research Fellowship Program under Grant No. DGE 2146752. R.A.-A. would like to acknowledge funding from the UC Noyce Initiative and the Global Research Platforms. Any opinions, findings, and conclusions or recommendations expressed in this material are those of the authors and do not necessarily reflect the views of the National Science Foundation.

\bibliography{refs}
\bibliographystyle{plainnat}

\appendix
\setcounter{figure}{0}\renewcommand{\thefigure}{A\arabic{figure}}\renewcommand{\theHfigure}{A\arabic{figure}}
\setcounter{table}{0}\renewcommand{\thetable}{A\arabic{table}}\renewcommand{\theHtable}{A\arabic{table}}

\section{Task construction}
\label{app:tasks}

\paragraph{Keeping labels out of the encoder.} No task label is ever an encoder input, because an
embedding that contained its own label would make the evaluation circular. The encoders read only a
model's correctness bits, the question text and the model's own answer text. Labels that we compute from the answers, such as accuracy
or answer length, use only the held-out evaluation questions, which no encoder reads. An embedding
therefore cannot score well by having seen the questions its labels come from.

\paragraph{Label sources.} The \emph{Hugging Face metadata} labels are downloads, license class,
architecture, model family, context length, release date and parameter count, read from each model card
and configuration file. The \emph{published leaderboards} are the Open LLM
Leaderboard v1 and v2 averages, the v2 MATH Lvl 5, GPQA and IFEval columns, LiveBench, can-ai-code
(junior and senior pass rates) and a Portuguese community board. The \emph{preference boards} are
Chatbot Arena and LMArena ratings, their style-control deltas (the rating minus the style-controlled
rating, overall and per category), AlpacaEval 2 length-controlled win rates, and the Ayumi role-play
rank and IQ scores. The \emph{safety and values} labels are the UGI willingness score, the refusal rate
on the LLMmap probes, and the distance to human preferences in the Moral Machine experiment
\citep{awad2018moral}. The \emph{serving and packaging} labels count how many quantized versions
ollama.com offers for a checkpoint and how many inference providers serve it through the Hugging Face
router. The \emph{probe-derived} labels are
accuracy on held-out probe questions, answer behavior (length, hedging, refusal and answer divergence),
reward-model scores of each (question, answer) pair, authorship, and the routing and selection
decisions a deployer would make from the embedding. We score length, hedging and refusal from the
answers by pattern matching. Four of these boards label too few of our models to become tasks
(Table~\ref{tab:paper_task_excluded}).

\paragraph{Task admission criteria.} A board becomes a task only if we can match it to our models by
name, it labels enough of them, and we can set its chance level. We match names exactly first, then
through an alias map that we checked by hand, and we never use fuzzy matching. Before matching, we
ignore case and punctuation and remove the organization prefix, because sources write one checkpoint
as \texttt{meta-llama/Meta-Llama-3-8B-Instruct}, \texttt{Meta-Llama-3-8B-Instruct} or
\texttt{llama-3-8b-instruct}. A board runs only if it labels at least nine of our models, and it is scored only if every one of
the five folds leaves at least eight labeled models to train the head and four to score it. The four
boards of Table~\ref{tab:paper_task_excluded} run but fail the second test. Where a metric has a known chance level, such as one half
for a two-class AUROC or 0 for a rank correlation, we use it. Otherwise, we measure the chance level by
refitting the same head on embeddings shuffled across models. We do this because, with a few dozen
labeled models, a ridge head shrinks its predictions toward the mean of the training folds. That mean is
anti-correlated with the held-out labels, so a chance level fixed at 0 would reward every embedding for
this shrinkage.

\paragraph{Task types.} Each task's type, set by what the head predicts, fixes its metric. \emph{Regression} tasks (\bpNTypeRegression{} tasks) predict one number per model. We score
most of them by Spearman correlation, because published scores come on different scales, such as an
Elo in the thousands and a win rate in percent. The few with a meaningful scale (mean accuracy,
parameter count, reasoning tilt and specialization) report $R^2$. \emph{Classification} tasks (\bpNTypeClassification{} tasks) predict a category from the model
card or a verdict on each question. They report macro-F1 for unbalanced columns with many classes and
AUROC for two-class and per-question labels. Authorship reports top-1 retrieval accuracy, with chance
$1/n$. \emph{Ranking} tasks (\bpNTypeRanking{} tasks) order models on a benchmark the head never trained
on, scored by nDCG@10, or predict whether an unseen model lands in a board's top $K$, scored by AUROC.
The \emph{pairwise} task predicts which of two models scores higher on each
benchmark. The \emph{routing} task sends each question to the model that the
embedding predicts is best, and reports the fraction of the gap to an oracle router that it closes. The
\emph{selection} task picks a set of models within a \bpServeBudget\,GB
serving budget, and reports the fraction of the headroom above the best single affordable model
that it closes. We score these two decision tasks against a baseline because even an uninformative
embedding can route and select. With raw accuracy as the score, always deploying the most accurate
model would beat every encoder, and raw retention would put all encoders within a few percent.

\paragraph{Exclusions.} Some public signals look like tasks but would not test an embedding fairly,
so we leave them out. Targets computed from the pool questions or from the correctness bits only
restate what the encoder reads. Boards with too few labeled models in some fold are too small, since a head fitted on so few models
reflects its regularization more than the embedding. Router leaderboards
rank routing systems, not single models, and model-card fields with too few positive examples cannot
form a class. We also keep the older and current versions of a board apart, because they use
different prompts.

\begin{table}[t]
  \centering
  \small
  \setlength{\abovecaptionskip}{0pt}\setlength{\belowcaptionskip}{8pt}
  \caption{The 26 established tasks.}
  \label{tab:paper_task_established}
  \begin{adjustbox}{max width=\linewidth}
\begin{tabular}{>{\raggedright\arraybackslash}p{2.6cm}>{\raggedright\arraybackslash}p{3.6cm}>{\raggedright\arraybackslash}p{6.6cm}}
\toprule
class & task & prior source \\
\midrule
\multicolumn{3}{l}{\textbf{competence} (21 tasks)} \\
probe accuracy & mean accuracy & benchmark prediction \citep{zhuang2025embedllm,chen2025irtnet} \\
 & held-out benchmark & benchmark prediction \citep{zhuang2025embedllm,chen2025irtnet} \\
 & benchmark ranking & benchmark prediction \citep{zhuang2025embedllm,chen2025irtnet} \\
 & pairwise, per benchmark & benchmark prediction \citep{zhuang2025embedllm,chen2025irtnet} \\
 & per-question correctness & correctness forecasting \citep{zhuang2025embedllm,chen2025irtnet,patel2026locus,hundred2024instances} \\
 & specialization & specialization clusters \citep{chen2025irtnet} \\
 & reasoning tilt & new readout of correctness \\
 & error type & new readout of correctness \\
 & item information & IRT item parameters \citep{chen2025irtnet,yao2026jeirt} \\
\addlinespace[3pt]
benchmark boards & Open LLM v1 & benchmark prediction \citep{zhuang2025embedllm,modellens2026,zeng2026benchpress,beeching2023openllm} \\
 & Open LLM v2 & benchmark prediction \citep{zhuang2025embedllm,modellens2026,zeng2026benchpress,fourrier2024openllmv2} \\
 & OLL v2 MATH & benchmark prediction \citep{zhuang2025embedllm,modellens2026,zeng2026benchpress,fourrier2024openllmv2,hendrycks2021math} \\
 & OLL v2 GPQA & benchmark prediction \citep{zhuang2025embedllm,modellens2026,zeng2026benchpress,fourrier2024openllmv2,rein2024gpqa} \\
 & IFEval & benchmark prediction \citep{zhuang2025embedllm,modellens2026,zeng2026benchpress,fourrier2024openllmv2,zhou2023ifeval} \\
 & Portuguese LLM & benchmark prediction \citep{zhuang2025embedllm,modellens2026,zeng2026benchpress} \\
 & can-ai-code jr & benchmark prediction \citep{zhuang2025embedllm,modellens2026,zeng2026benchpress} \\
 & can-ai-code sr & benchmark prediction \citep{zhuang2025embedllm,modellens2026,zeng2026benchpress} \\
 & LiveBench language & benchmark prediction \citep{zhuang2025embedllm,modellens2026,zeng2026benchpress,white2025livebench} \\
\addlinespace[3pt]
model selection & top-k selection & portfolio selection \citep{patel2026locus} \\
 & query routing & routing \citep{zhuang2025embedllm,ong2025routellm,patel2026locus} \\
 & budgeted selection & portfolio selection \citep{patel2026locus} \\
\addlinespace[3pt]
\midrule
\multicolumn{3}{l}{\textbf{preference} (5 tasks)} \\
preference ratings & LMArena rating & Arena preferences \citep{chiang2024chatbotarena} \\
 & LMArena coding & Arena preferences \citep{chiang2024chatbotarena} \\
 & Ayumi RP rank & new readout of judge rating \\
 & Ayumi RP IQ & new readout of judge rating \\
 & reward score (Skywork-RM) & new readout of reward model \citep{liu2025skyworkrewardv2} \\
\addlinespace[3pt]
\bottomrule
\end{tabular}
  \end{adjustbox}
  \par\smallskip
  \parbox{\linewidth}{\scriptsize Competence (accuracy on held-out probe questions, published benchmark boards, model selection) and preference. Each is a kind of target prior model embeddings are evaluated on; the right column names that kind and where it comes from. Table~\ref{tab:paper_task_vocabulary} lists the 20 new tasks.}
\end{table}

\begin{table}[t]
  \centering
  \small
  \setlength{\abovecaptionskip}{0pt}\setlength{\belowcaptionskip}{8pt}
  \caption{Tasks the main tables do not score.}
  \label{tab:paper_task_excluded}
  \begin{adjustbox}{max width=\linewidth}
\begin{tabular}{llrl}
\toprule
task & class & $n$ & reason \\
\midrule
AlpacaEval & preference ratings & 20 & too few labeled models \\
Arena Elo (LMSYS) & preference ratings & 17 & too few labeled models \\
UGI willingness & refusal, safety & 15 & too few labeled models \\
Moral Machine (Takemoto) & refusal, safety & 23 & too few labeled models \\
\bottomrule
\end{tabular}
  \end{adjustbox}
  \par\smallskip
  \parbox{\linewidth}{\scriptsize Each labels too few of our models for five model-held-out folds. $n$: models with a label. The fast-protocol budget sweeps (Figures~\ref{fig:paper_task_panels}b and \ref{fig:paper_k_sweep_classes}) still score them.}
\end{table}

\section{Encoders, score and data in detail}
\label{app:encoders}

\paragraph{Notation.} Each model $m\in\mathcal{M}$ answers each question $q\in\mathcal{Q}$ once, with
answer $a_{mq}$, correctness bit $B_{mq}=\1[a_{mq}\text{ correct}]$ and answer embedding
$e_{mq}=e(a_{mq})$. Here $e$ is BGE with mean pooling over tokens (768 dimensions), and $u_q=e(q)$ embeds
the question. On a subset of questions $S\subseteq\mathcal{Q}$, $B_S$ holds the bit columns of $S$ and
$T_S$ holds each model's pooled answer text $t_m(S)=\bar e_m/\|\bar e_m\|$, with
$\bar e_m=|S|^{-1}\sum_{q\in S}e_{mq}$. To choose $S_K$, we fit a two-parameter IRT model
$\hat p_{mq}=\sigma\big(\hat\alpha_q(\hat\theta_m-\hat b_q)\big)$ on the training models, with model
ability $\hat\theta_m$, question discrimination $\hat\alpha_q$ and question difficulty $\hat b_q$. We
keep the $K$ pool questions with the largest Fisher information
$I_q=\sum_m\hat\alpha_q^2\hat p_{mq}(1-\hat p_{mq})$. \emph{Full} encoders read
$\mathcal{P}=\mathcal{Q}_{\mathrm{pool}}$ (the 900 pool questions every model answered, or all 24,118 for the models that answered every question), and
\emph{budget} encoders read $S_{400}$ in the tables and $S_K$ in the $K$ sweep.

\paragraph{Score.} With metric $\mu_i$, chance level $c_i$ and best attainable value $\mu_i^\ast$,
\begin{equation*}
s_i(E)=\mathrm{clip}\Big(\frac{\mu_i(\hat y_i,y_i)-c_i}{\mu_i^\ast-c_i},\,\pm1\Big),\qquad
\Delta_i(E)=s_i(E)-\frac1R\sum_{r=1}^{R}s_i\big(E_{\mathrm{null}}^{d,(r)}\big),
\end{equation*}
where each random embedding $E_{\mathrm{null}}^{d,(r)}$ draws an independent $z_m\sim\mathcal{N}(0,I_d)$
for every model. We use $R=\bpFastFloorDraws$ draws under the fast protocol and 10 per fold and seed
under the strict one (Protocol details below). A floor can fall below 0, so $\Delta_i$ can exceed 1
(Table~\ref{tab:paper_complementarity}). Over a set of tasks $\mathcal{T}$, the weighted mean is
$\bar\Delta_{\mathcal{T}}=\sum_{i}w_i\Delta_i/\sum_{i}w_i$. Because some tasks rank models alike, such
as Arena rating and Elo, $w_i$ is the task's share of the
effective number of independent tests \citep{li2005meff}. With $(\lambda_k,v_k)$ the eigenpairs of the
task-by-task association matrix of the labels, $w_i=\sum_k f(|\lambda_k|)\,v_{ik}^2$ with
$f(x)=\1[x\ge 1]+(x-\floor{x})$. The matrix holds the absolute Spearman correlation between numeric
labels, the bias-corrected Cram\'er's $V$ between categorical labels, and 0 for tasks without shared
labels. The weights sum to $M_{\mathrm{eff}}$ (\bpMeff{} over the \bpNAll{} tasks), an uncorrelated task
keeps $w_i=1$, and $k$ perfectly correlated tasks share one unit.

\begin{table}[h]
\caption{Every encoder in the appendix tables. $\mathrm{PCA}_d$: top-$d$ component scores of the
column-centered input, $\bar X$ that centering, $z(\cdot)$ a column z-score, $\tilde z(\cdot)$ a column z-score scaled to unit total block variance, $\mathrm{res}(X|Y)$ the
ridge residual of $X$ on $[\1\;Y]$, $\mathrm{PCA95}$ a PCA keeping 95\% of the variance, $\pi$ a
derangement of $\mathcal{M}$ (a plain permutation for BULLPEN-AE). A swap control reads what its encoder reads. Main-text encoders in bold.}
\label{tab:instantiations}
\centering\small
\begin{tabular}{@{}l@{\hspace{4pt}}l@{\hspace{4pt}}r@{\hspace{4pt}}>{\raggedright\arraybackslash}p{0.56\linewidth}@{}}
\toprule
Name & Reads & $d$ & $E(m)$ \\
\midrule
\textbf{Leaderboard scalar} & $B$ & 1 & $c_m=|\mathcal{P}|^{-1}\sum_{q\in\mathcal{P}}B_{mq}$ \\
\textbf{IRT-2PL} & $B$ & 1 & $\theta_m$ of the 2PL, logits $\hat\alpha_q(\theta_m-\hat b_q)$ \\
\textbf{PCA32-bits} & $B$ & 32 & $\mathrm{PCA}_{32}(z(B_{\mathcal{P}}))_m$ \\
PCA32-bits (centered) / -budget & $B$ & 32 & $\mathrm{PCA}_{32}(B_{\mathcal{P}})_m$ ($B_{S_{400}}$) \\
\makecell[l]{PCA32-$\langle$channels$\rangle$\\ / -budget} & $B,u,T$ & 32 & $\mathrm{PCA}_{32}([\,\tilde z(X_1)\,\cdots\,\tilde z(X_J)\,])_m$ over the channels in Reads; -budget reads $S_{400}$ \\
BULLPEN-LogPCA-full & $B$ & 32 & L2-penalized Bernoulli likelihood, logits $\delta_q+z_m^\top w_q$ \\
BULLPEN-AE-full / -budget & $B$ & 32 & autoencoder, cross-entropy on $B_{\mathcal{P}}$ ($B_{S_{400}}$) \\
BULLPEN-QMF-full & $B,u$ & 32 & EmbedLLM-style factorization, $\sigma(w^\top(p_m\odot Pu_q))$, $z_m=p_m$ \\
\textbf{BULLPEN-bits+q} & $B,u$ & 32 & $[\,z(\bar B_{\mathcal{P}}W_b)\;\,z(Q_{\mathcal{P}}W_q)\,]_m$, PLS-SVD of the bits and the question channel (Eq.~\ref{eq:pls}), 16 + 16 \\
\textbf{BULLPEN-ans} & $T$ & 768 & $t_m(\mathcal{P})$ \\
BULLPEN-AE-full / -budget & $T$ & 32 & autoencoder, squared error on $z(T)$ \\
BULLPEN-zConcat-budget & $B,T$ & 32 & $\mathrm{PCA}_{32}\big([\,z(B_{S_{400}})\;\,z(T_{\mathcal{P}})\,]\big)_m$; the $K$ sweep reads $T_{S_K}$ \\
BULLPEN-zSlice-budget & $B,T$ & 32 & $[\,c_m\;\,b^{\perp}_m\;\,t^{\perp}_m\,]$; $b^{\perp}=\mathrm{PCA95}(\mathrm{res}(\mathrm{PCA}_{32}(B_{S_K})|c))$, $t^{\perp}=\mathrm{PCA95}(\mathrm{res}(T_{S_K}|c,\mathrm{PCA}_{32}(B_{S_K})))$ \\
\textbf{BULLPEN-bits+ans} & $B,T$ & 32 & $[\,z(\bar B_{\mathcal{P}}W_b)\;\,z(\bar T_{\mathcal{P}}W_t)\,]_m$; $(W_b,W_t)$ top 16 singular pairs of $\bar B_{\mathcal{P}}^{\top}\bar T_{\mathcal{P}}$ \\
BULLPEN-CCA-full & $B,T$ & \bpDimCCALo--\bpDimCCAHi & $[\,h_m\;\,p^{b}_m\;\,p^{t}_m\,]$, CCA of $\mathrm{PCA}_{32}(B_{\mathcal{P}})$ and $\mathrm{PCA}_{32}(T_{\mathcal{P}})$; $p^{v}=\mathrm{PCA95}(\mathrm{res}(\text{view }v|h))$ \\
BULLPEN-AE-full & $B,T$ & 32 & autoencoder, cross-entropy on $B$ plus squared error on $z(T)$ \\
BULLPEN-CCAfused-full & $B,u,T$ & \bpDimFusedLo--\bpDimFusedHi & BULLPEN-CCA with bits view $\mathrm{PCA}_{32}(\mathrm{LOCUS}_{128}(B_{\mathcal{P}},u))$; -budget reads $S_{400}$ \\
\textbf{BULLPEN-fused} & $B,u,T$ & 32 & multi-block PLS-SVD of the bits, question and answer channels (Eq.~\ref{eq:pls}), 12 + 10 + 10 \\
Swap control & & & $E(B,T_{\pi(\cdot)})_m$, $E\in\{$zConcat, PCA32, PLS, Fused, AE$\}$ \\
Null & -- & $d$ & $z_m\sim\mathcal{N}(0,I_d)$ \\
\bottomrule
\end{tabular}
\end{table}

\paragraph{Joint objectives.} BULLPEN-zConcat minimizes the squared reconstruction error over both
z-scored channels, so the channel with more columns carries more weight. BULLPEN-zSlice keeps
competence once, as $c_m$. It then adds what the bits hold beyond competence and what the text holds
beyond both, each at unit total variance. PLS maximizes $w_b^\top\bar B^\top\bar Tw_t$, and CCA
\citep{hotelling1936cca}, with ridge $\lambda=\bpCcaRidge$, maximizes the correlation between the two
channels. The shared part of the CCA embedding, $h_m$, averages the z-scored canonical variates that
exceed the 95th percentile of a 200-permutation null, so BULLPEN-CCA and BULLPEN-CCAfused choose their
own width.

\paragraph{Models and question split.} We collected answers in two stages. The first \bpNFullset{}
models answered every probe question. We then added \bpNExtset{} models that public leaderboards list,
so that more models carry external labels, and these answered a \bpNCore-question subset of the probe.
We split the probe once, keeping each benchmark's share the same in every part, into a pool of
\bpNPool{} questions, 3,072 tuning questions and 4,096 evaluation questions. The subset splits the
same way into 900 pool, 122 tuning and 159 evaluation questions, so all \bpNModels{} models answered the
same 900 pool questions.

\paragraph{Choosing the subset.} The subset gives each of the 68 graded parts of the probe a quota
proportional to the square root of its size, with at least 12 questions each. Each quota is split
evenly over five difficulty levels, cut on the pass rates of the models that answered every question.

\paragraph{Embedding a new model.} A new model answers only the $K$ selected questions. We keep the
encoder's loadings fixed and place the model's bits by a ridge fit over the questions it answered. We
compute and project its question and answer channels in the same way. The held-out models in our
experiments are placed by exactly this procedure.

\paragraph{Budget against the full pool.} This check asks how much skill an encoder keeps when it
reads only $K$ questions instead of the whole \bpPoolN-question pool. For each seed, we divide an
encoder's skill over floor at $K$ by its skill over floor on the whole pool, and average this ratio
over three seeds. We use the \bpNFullset{} models that answered every question, the fast protocol, and
the \bpFullTasks{} tasks the encoder scores (the main tasks except those that predict held-out
benchmarks, plus the four boards of Table~\ref{tab:paper_task_excluded}).
Table~\ref{tab:budget_pool} gives the results. The encoders that read answer text keep nearly all of
their skill from about 400 questions, while PCA32-bits needs about \bpPoolPcaNeedK{} questions to reach the same
level. Values above 100\% are within seed noise. We did not run BULLPEN-bits+q on the whole pool.

\begin{table}[htbp]
\caption{Share of whole-pool skill kept with $K$ questions, in percent (range over seeds at $K=400$)}
\label{tab:budget_pool}
\centering\small
\begin{tabular}{@{}lccc@{}}
\toprule
Encoder & $K=400$ & $K=900$ & $K=2{,}000$ \\
\midrule
BULLPEN-ans & \bpPoolAnsFourHundred{} (\bpPoolAnsFourHundredLo--\bpPoolAnsFourHundredHi) & \bpPoolAnsNineHundred & \bpPoolAnsTwoThousand \\
BULLPEN-bits+ans & \bpPoolBitsAnsFourHundred{} (\bpPoolBitsAnsFourHundredLo--\bpPoolBitsAnsFourHundredHi) & \bpPoolBitsAnsNineHundred & \bpPoolBitsAnsTwoThousand \\
BULLPEN-fused & \bpPoolFusedFourHundred{} (\bpPoolFusedFourHundredLo--\bpPoolFusedFourHundredHi) & \bpPoolFusedNineHundred & \bpPoolFusedTwoThousand \\
PCA32-bits & \bpPoolPcaBitsFourHundred{} (\bpPoolPcaBitsFourHundredLo--\bpPoolPcaBitsFourHundredHi) & \bpPoolPcaBitsNineHundred & \bpPoolPcaBitsTwoThousand \\
\bottomrule
\end{tabular}
\end{table}
\begin{figure}[t]
  \centering
  \includegraphics[width=\linewidth]{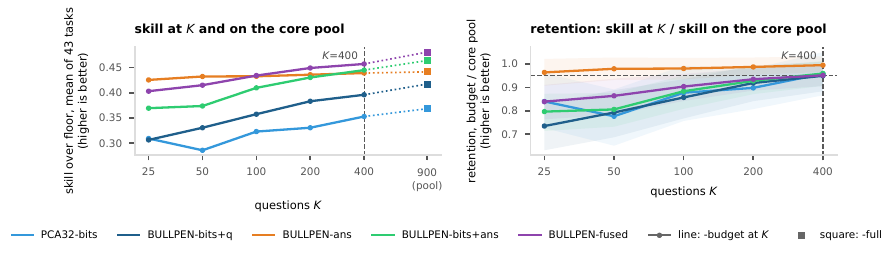}
  \caption{The held-out check of Figure~\ref{fig:paper_task_panels}b: budget encoders refitted on $K$ questions under the strict protocol (all \bpNModels{} models, 5 publisher-grouped folds $\times$ 3 seeds), against the 900-question core pool every model answered. Left: skill over the floor; squares: the same encoder on the whole core pool (900). Right: retention, skill at $K$ divided by skill on the core pool (dashed: 95\% and $K$=400, the strict rule's $K^\ast$); band: 95\% bootstrap over runs and tasks. Higher is better.}
  \label{fig:paper_k_sweep_core}
\end{figure}

\begin{figure}[t]
  \centering
  \includegraphics[width=\linewidth]{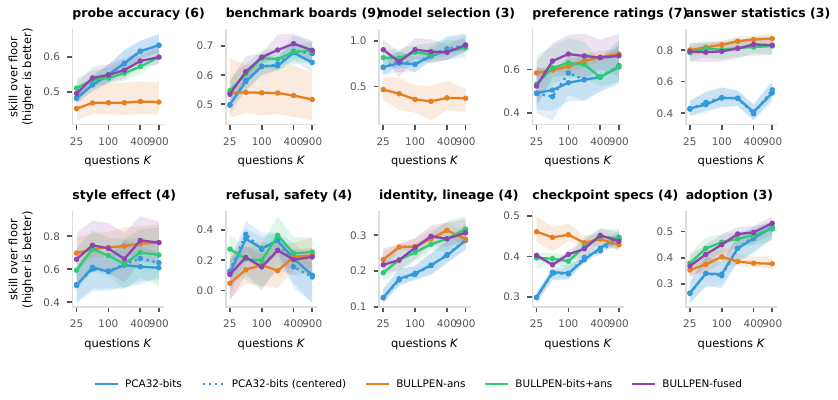}
  \caption{The $K$ sweep per task class (Tables~\ref{tab:paper_task_vocabulary} and \ref{tab:paper_task_established}), on the core view: skill over floor of each encoder refitted on $K$ questions (its -budget variant), averaged over the class's tasks with $M_{\text{eff}}$ weights (panel title: the class and its number of tasks). Fast protocol (every model embedded, task heads out of fold), so levels differ from the strict Figure~\ref{fig:paper_k_sweep_core}; at the pool an encoder reads all 900 core questions, and BULLPEN-bits+q is not in this sweep. Lines: mean over 3 seeds; band: lowest to highest seed. Higher is better. The budget of Table~\ref{tab:paper_master_table} is one $K$ (400) for all classes. The sweep scores the tasks of the budget sweep (Appendix~\ref{app:encoders}), so class counts differ from Tables~\ref{tab:paper_task_vocabulary} and \ref{tab:paper_task_established}.}
  \label{fig:paper_k_sweep_classes}
\end{figure}

\paragraph{Competitors.} The eight prior encoders are EmbedLLM, LOCUS, RouteLLM-MF, IrtNet, JE-IRT and
IRT-Router, which read the correctness bits and the question text
\citep{zhuang2025embedllm,patel2026locus,ong2025routellm,chen2025irtnet,yao2026jeirt,song2025irtrouter},
LLMmap, which reads the answers to its own eight published probes \citep{pasquini2025llmmap}, and LLM DNA,
which needs no training and which we recompute from our answers \citep{wu2026llmdna}. We refit each
competitor in every fold on the same training models and the same 900 pool questions as our encoders.
Each keeps its published width and follows the released code where there is one. EmbedLLM and
IrtNet were designed for the whole EmbedLLM probe, which has far more questions than our pool. We
changed the published recipes in five ways.
First, we removed inputs that are our prediction targets. IRT-Router's model profile (release date,
developer and type) repeats our provenance labels, so our IRT-Router reads only the bits and the
question text. RouteLLM-MF keeps its Bradley--Terry loss, but we train it on pairs derived from
correctness, in which a correct model beats an incorrect one, following its own golden-label
augmentation. We do not use Arena votes, which are our preference targets. The competitor audits
below restore both removed inputs.
Second, EmbedLLM, RouteLLM-MF, IrtNet, JE-IRT and IRT-Router learn a lookup table with one vector per
training model, and only JE-IRT describes how to embed an unseen model. For all five, we place every
model, training models included, by fitting its vector to its pool bits with the question parameters
frozen, as JE-IRT does. Scoring the trained vectors instead does not help these competitors
(competitor audits below). LOCUS,
LLMmap and LLM DNA embed a model in one forward pass.
Third, IrtNet and JE-IRT initialize the model table with a small variance instead of the published
unit variance. With unit variance and 900 questions, the selected checkpoint barely moves from its
random start, so these two methods would score like a random embedding.
Fourth, four competitors read a frozen text encoder from our pipeline in place of their own. LLM DNA,
LLMmap and RouteLLM-MF use bge-base in place of Qwen3-Embedding-8B, multilingual-e5-large and
text-embedding-3-small, in that order. IRT-Router uses all-mpnet-base-v2 in place of bert-base, and
EmbedLLM, LOCUS, IrtNet and JE-IRT use all-mpnet-base-v2 as published.
Fifth, LOCUS draws fewer anchor questions than published, because our pool is smaller than its anchor
set.

\paragraph{Competitor audits.}\label{app:comp-audits} We checked that our changes to the published
recipes do not handicap the competitors. Table~\ref{tab:paper_master_table} runs LLMmap as published,
on the answers to its own probes, which \bpMapOwnN{} of the \bpNModels{} models answered. A version
retrained on our pool answers scores a little higher (the own-probe version is lower by
\bpMapOwnMinusStandNew{} on the new tasks, [\bpMapOwnMinusStandNewLo, \bpMapOwnMinusStandNewHi]), and
both stay below BULLPEN-fused-full on the new tasks. The released LLMmap weights give the same result as
a retrained network. Restoring IRT-Router's model profile lowers its score, and training RouteLLM-MF on Chatbot Arena's own votes does not improve it on the \bpArenaN{} of our models that
appear in Arena's public battle data (a difference
of \bpArenaMinusAdapted, [\bpArenaMinusAdaptedLo, \bpArenaMinusAdaptedHi]). Scoring the lookup-table
competitors on their own trained vectors, instead of our placement rule, lowers or does not change
their scores.

\paragraph{Prediction heads.} The prediction heads are small and mostly linear
(Section~\ref{sec:score}). A head $D_i$ is a ridge regression for numeric labels or a logistic
regression for categorical and per-question labels on $z_m$. It is fitted out of fold over five model
folds, so the prediction $\hat y_i(m)=D_i^{(-f(m))}(z_m)$ comes from a head that never saw model $m$.
Two kinds of task add a simple step after the linear head. The routing and selection tasks send each
question, or spend a budget, on the models whose predicted correctness is highest. The authorship task
predicts each held-out model's profile of answer statistics with a ridge head and checks whether the
nearest true profile is the model's own.

\paragraph{Protocol details.} We use two protocols. The \emph{strict} protocol is the held-out
protocol of Section~\ref{sec:protocol}, and Table~\ref{tab:paper_master_table} and the held-out results of Section~\ref{sec:protocol} use it. It averages three seeds, and each seed redraws the folds and the
encoder's initialization. Where a task has one label per model (\bpPooledN{} of the \bpNAll{} tasks), we
pool the held-out predictions of all five folds and score them once against a floor pooled the same
way. Model selection and the per-question tasks cannot be pooled, so for them we average over folds.
The tasks that predict held-out benchmarks hold out 13 benchmark groups in turn, with MMLU split by
category. The \emph{fast} protocol fits the encoder on every model and only the heads out of fold. It
is cheaper, and we use it for the whole-pool budget sweep (Figure~\ref{fig:paper_task_panels}b) and for
the embedding-structure figure, which shows every model, as the embedding maps of LOCUS do
\citep{patel2026locus}. All intervals are 95\% paired bootstraps. For pooled scores, each draw
resamples the models. In Table~\ref{tab:paper_master_table}, it resamples folds and seeds in the class
columns and tasks in the two right-hand columns.

\paragraph{Further limitations.}\label{app:further-lim} The folds keep each organization's models
together, but they do not keep all models that share a base model together. Model selection and the
per-question tasks cannot be pooled, so they have no pooled held-out test. Because name matching ignores the organization, a re-upload can take another model's board row.
LLMmap's own probes cover \bpMapOwnN{} of the \bpNModels{} models, and
Table~\ref{tab:paper_master_table} imputes the rest, so LLMmap's row there is a lower bound.

\paragraph{Collection.} Non-reasoning models decode greedily, with up to \bpDecMaxTok{} new tokens.
Reasoning models loop under greedy decoding, so they sample with temperature \bpDecTemp, top-$p$
\bpDecTopP{} and top-$k$ \bpDecTopK, with up to \bpDecMaxTokThink{} tokens. We drop a model's
collection if it covers less than \bpCoverageGate\% of the subset, if it has no valid output, or if over half its answers on the subset are blank, template tokens or prompt echoes.

\section{Per-group results for every embedding}
\label{app:per-group}

\begin{figure}[t]
  \centering
  \includegraphics[width=\linewidth]{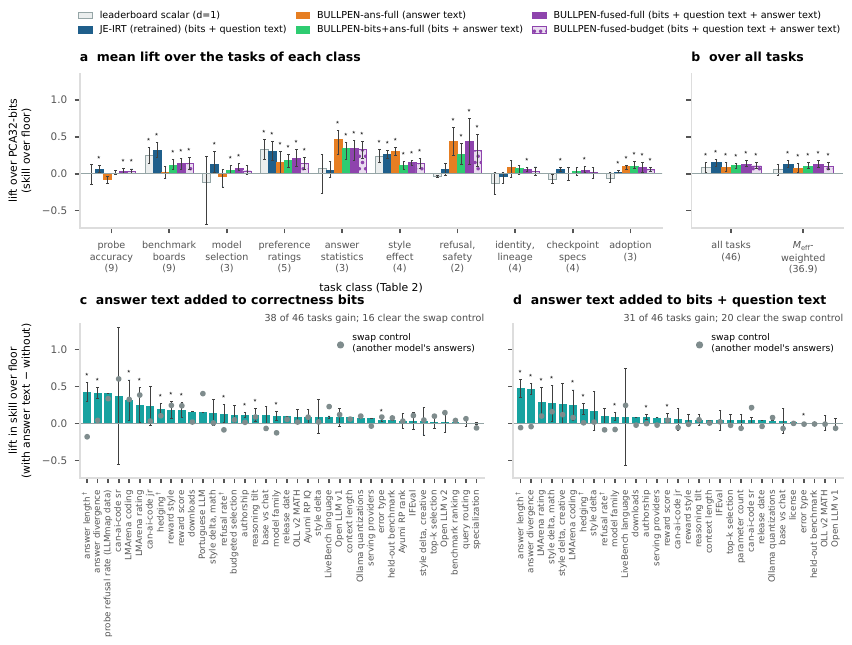}
  \caption{What each embedding adds over the correctness baseline, in the task classes of Table~\ref{tab:paper_task_vocabulary} (strict protocol; Table~\ref{tab:paper_master_table} prints the raw skill). (a) Mean lift in skill over floor over PCA32-bits (z-scored), per class; (b) the same over all tasks, unweighted and $M_{\text{eff}}$-weighted \citep{li2005meff}. Error bars: 95\% paired task bootstrap; $\star$: the interval is above zero. JE-IRT is the strongest retrained competitor; the retrained competitors have a fixed input, so only our encoders are tested with and without answer text. BULLPEN-fused-budget reads $K^\ast$ questions. (c,d) Tasks where adding answer text helps, each a pair of embeddings that differ only in that channel: (c) BULLPEN-bits+ans $-$ PCA32-bits, (d) BULLPEN-fused $-$ BULLPEN-bits+q. Error bars: 95\% interval over the 15 fold $\times$ seed runs; $\star$: that interval excludes zero. Gray dots: the same difference with each model given another model's answers. $\dagger$: pattern-matched behavior traits.}
  \label{fig:paper_competence_axes}
\end{figure}

Our encoders are named \texttt{BULLPEN-<channels>-<full|budget>}, where \emph{full} reads the whole
question pool (the 900 pool questions every model answered) and \emph{budget} reads the 400 questions chosen by Fisher
information. Table~\ref{tab:instantiations} (Appendix~\ref{app:encoders}) gives each encoder's width and
construction. Figure~\ref{fig:paper_competence_axes} splits the lift over PCA32-bits of
Figure~\ref{fig:paper_task_panels} by task class.


\section{Structure of the BULLPEN-fused embedding}
\label{app:embedding}

We ran PCA and t-SNE on the BULLPEN-fused-full embedding. Figure~\ref{fig:paper_embedding_structure}b
relates each of its first ten principal components to every task that has a per-model label. Each of these components holds between \bpFusedPcShareLo\% and \bpFusedPcShareHi\% of the variance. PC2 tracks competence, following mean probe
accuracy ($\rho=$\,\bpFusedPcTwoAcc), reward-model score (\bpFusedPcTwoReward) and LMArena rating
(\bpFusedPcTwoArena). Lineage loads on other components, with architecture on PC1
($\eta=$\,\bpFusedPcOneArch) and model family on PC4 ($\eta=$\,\bpFusedPcFourFamily). The t-SNE map in
Figure~\ref{fig:paper_embedding_structure}a shows the same split. Accuracy runs along one direction of
the map, while architecture and base versus chat models form regions that cut across it. Newer models
are also more accurate ($\rho=$\,\bpFusedDateAcc), so the release-date panel largely repeats the
accuracy panel. Both panels use the embedding fitted on all models (seed 0), and the t-SNE runs on its
z-scored values with perplexity \bpTsnePerplexity, and the heatmap keeps tasks labeled on at least
\bpFusedHeatMinN{} models.

\section{Every encoder on every task}
\label{app:master}
\paragraph{How Table~\ref{tab:paper_master_table} is scored and marked.} Each cell is the mean skill
over the floor of the encoder's width, averaged over the five organization-grouped folds, three seeds and the column's tasks.
The leaderboard scalar uses the exact floor for $d=1$. We fixed the class columns before seeing any
score. An encoder is bold when the 95\% paired-bootstrap interval of its difference from the column's
best includes 0, on the tasks both encoders score. The bootstrap resamples held-out folds and seeds in
the class columns and tasks in the two right-hand columns, which also give $\pm$ half the
task-bootstrap interval of the mean. An underline marks a sole winner. Green shading marks the bold
cells, and red marks the column's lowest embedding other than the leaderboard scalar. Gray italic rows
average a channel group.
After correction for multiple comparisons, \bpSigLeadSep{} of the \bpSigClasses{} class leaders separate
from their runner-up. Answer text alone leads every correctness encoder, with or without the question
text, on \bpLeadNew{} of the \bpNNew{} new tasks, and on \bpLeadSep{} of them the interval excludes zero.

Figure~\ref{fig:paper_master_heatmap} and Table~\ref{tab:paper_master_wide} expand
Table~\ref{tab:paper_master_table} to one column per task.
{\setkeys{Gin}{height=0.82\textheight,keepaspectratio}%
\begin{figure}[t]
  \centering
  \includegraphics[width=\textwidth]{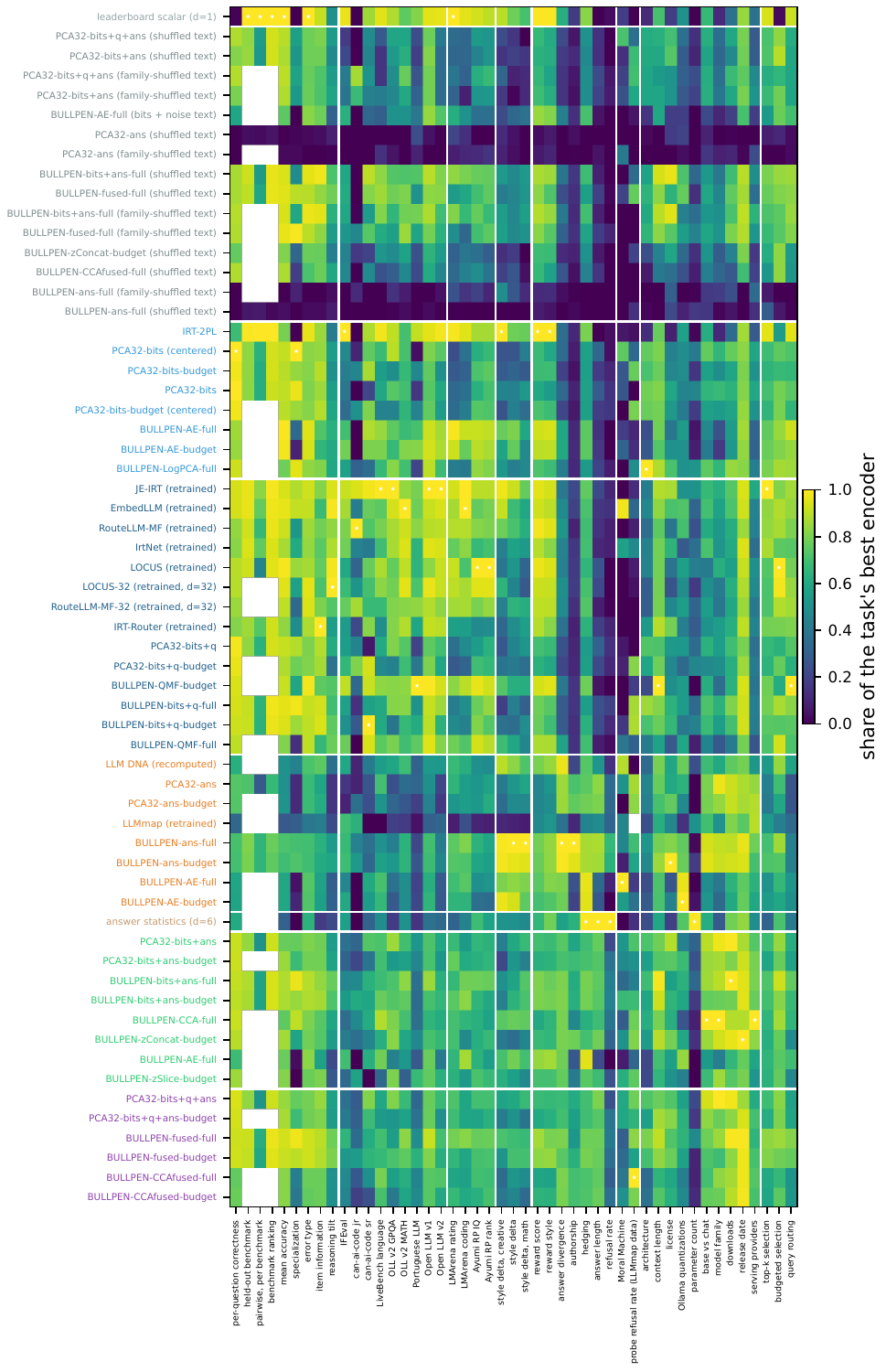}
  \caption{Every encoder (rows, grouped by input channel) on every task (columns, grouped by task group). Cell: skill over the encoder's width-matched floor as a share of the column's best encoder (star); values at or below the floor are 0. White: not scored (the three benchmark-transfer tasks are scored only for Table~\ref{tab:paper_master_table}'s encoders, and LLMmap and LLM DNA skip the per-benchmark refits). Strict protocol: 5 publisher-grouped folds $\times$ 3 seeds.}
  \label{fig:paper_master_heatmap}
\end{figure}
}
\begin{sidewaystable}
  \centering
  \small
  \setlength{\abovecaptionskip}{0pt}\setlength{\belowcaptionskip}{8pt}
  \caption{Skill over the width-matched floor of every encoder on every task.}
  \label{tab:paper_master_wide}
  \begin{adjustbox}{max totalsize={\textheight}{\dimexpr\textwidth-5\baselineskip\relax}}
\begin{tabular}{lrrrrrrrrrrrrrrrrrrrrrrrrrrrrrrrrrrrrrrrrrrrrrrr}
\toprule
 & \multicolumn{9}{c}{benchmark accuracy} & \multicolumn{9}{c}{public leaderboards} & \multicolumn{4}{c}{human preference} & \multicolumn{3}{c}{response style} & \multicolumn{7}{c}{answer behavior} & \multicolumn{2}{c}{safety} & \multicolumn{10}{c}{model metadata} & \multicolumn{3}{c}{model selection} \\
\cmidrule(lr){2-10} \cmidrule(lr){11-19} \cmidrule(lr){20-23} \cmidrule(lr){24-26} \cmidrule(lr){27-33} \cmidrule(lr){34-35} \cmidrule(lr){36-45} \cmidrule(lr){46-48}
 & \rotatebox{90}{per-question correctness} & \rotatebox{90}{held-out benchmark} & \rotatebox{90}{pairwise, per benchmark} & \rotatebox{90}{benchmark ranking} & \rotatebox{90}{mean accuracy} & \rotatebox{90}{specialization} & \rotatebox{90}{error type} & \rotatebox{90}{item information} & \rotatebox{90}{reasoning tilt} & \rotatebox{90}{IFEval} & \rotatebox{90}{can-ai-code jr} & \rotatebox{90}{can-ai-code sr} & \rotatebox{90}{LiveBench language} & \rotatebox{90}{OLL v2 GPQA} & \rotatebox{90}{OLL v2 MATH} & \rotatebox{90}{Portuguese LLM} & \rotatebox{90}{Open LLM v1} & \rotatebox{90}{Open LLM v2} & \rotatebox{90}{LMArena rating} & \rotatebox{90}{LMArena coding} & \rotatebox{90}{Ayumi RP IQ} & \rotatebox{90}{Ayumi RP rank} & \rotatebox{90}{style delta, creative} & \rotatebox{90}{style delta} & \rotatebox{90}{style delta, math} & \rotatebox{90}{reward score} & \rotatebox{90}{reward style} & \rotatebox{90}{answer divergence} & \rotatebox{90}{authorship} & \rotatebox{90}{hedging} & \rotatebox{90}{answer length} & \rotatebox{90}{refusal rate} & \rotatebox{90}{Moral Machine} & \rotatebox{90}{probe refusal rate (LLMmap data)} & \rotatebox{90}{architecture} & \rotatebox{90}{context length} & \rotatebox{90}{license} & \rotatebox{90}{Ollama quantizations} & \rotatebox{90}{parameter count} & \rotatebox{90}{base vs chat} & \rotatebox{90}{model family} & \rotatebox{90}{downloads} & \rotatebox{90}{release date} & \rotatebox{90}{serving providers} & \rotatebox{90}{top-k selection} & \rotatebox{90}{budgeted selection} & \rotatebox{90}{query routing} \\
\midrule
\multicolumn{48}{l}{\textit{floor}} \\
\quad leaderboard scalar (d=1) & 0.01 & \textbf{0.80} & \textbf{0.76} & \textbf{0.83} & 0.97 & -0.00 & 0.56 & 0.74 & 0.27 & 0.50 & -0.06 & 0.75 & 0.93 & 0.43 & 0.52 & 0.58 & 0.75 & 0.68 & 0.79 & 0.72 & 0.59 & 0.56 & 0.45 & 0.49 & 0.52 & \textbf{0.90} & \textbf{0.81} & 0.56 & 0.04 & 0.74 & -0.00 & 0.01 & 0.06 & -0.06 & 0.06 & 0.34 & 0.06 & 0.19 & 0.00 & 0.37 & 0.05 & 0.17 & 0.56 & 0.13 & 0.93 & 0.02 & 0.53 \\
\midrule
\multicolumn{48}{l}{\textit{shuffled text (control)}} \\
\quad PCA32-bits+q+ans (shuffled text) & 0.45 & 0.67 & 0.42 & 0.77 & 0.87 & 0.07 & 0.44 & 0.65 & 0.30 & 0.09 & -0.05 & 0.42 & 0.27 & 0.29 & 0.41 & 0.17 & 0.60 & 0.29 & 0.28 & 0.25 & 0.34 & 0.32 & 0.12 & 0.10 & -0.07 & 0.60 & 0.50 & 0.21 & 0.01 & 0.46 & 0.13 & -0.01 & 0.23 & 0.32 & 0.30 & 0.30 & 0.22 & 0.14 & 0.04 & 0.35 & 0.16 & 0.27 & 0.52 & 0.10 & 0.71 & 0.75 & 0.36 \\
\quad PCA32-bits+ans (shuffled text) & 0.41 & 0.68 & 0.41 & 0.77 & 0.85 & 0.07 & 0.44 & 0.64 & 0.26 & 0.12 & -0.03 & 0.40 & 0.15 & 0.27 & 0.43 & 0.13 & 0.63 & 0.25 & 0.26 & 0.25 & 0.28 & 0.32 & 0.08 & 0.13 & 0.12 & 0.58 & 0.51 & 0.16 & 0.03 & 0.43 & 0.13 & 0.00 & 0.04 & 0.59 & 0.29 & 0.29 & 0.23 & 0.11 & 0.03 & 0.32 & 0.15 & 0.21 & 0.52 & 0.10 & 0.72 & 0.80 & 0.35 \\
\quad PCA32-bits+q+ans (family-shuffled text) & 0.45 & -- & -- & -- & 0.88 & 0.07 & 0.43 & 0.59 & 0.30 & 0.21 & 0.24 & 0.45 & 0.15 & 0.24 & 0.53 & 0.19 & 0.62 & 0.25 & 0.31 & 0.24 & 0.39 & 0.42 & 0.17 & 0.06 & 0.09 & 0.60 & 0.47 & 0.14 & 0.01 & 0.44 & 0.21 & 0.03 & 0.06 & 0.60 & 0.28 & 0.26 & 0.16 & 0.11 & 0.08 & 0.34 & 0.17 & 0.31 & 0.50 & 0.16 & 0.62 & 0.59 & 0.34 \\
\quad PCA32-bits+ans (family-shuffled text) & 0.40 & -- & -- & -- & 0.85 & 0.07 & 0.43 & 0.59 & 0.32 & 0.20 & 0.19 & 0.38 & 0.41 & 0.23 & 0.42 & 0.22 & 0.62 & 0.25 & 0.25 & 0.06 & 0.33 & 0.33 & 0.08 & 0.01 & 0.09 & 0.62 & 0.51 & 0.13 & 0.01 & 0.44 & 0.19 & 0.04 & 0.08 & 0.52 & 0.28 & 0.28 & 0.16 & 0.10 & 0.06 & 0.35 & 0.18 & 0.27 & 0.52 & 0.18 & 0.68 & 0.70 & 0.34 \\
\quad BULLPEN-AE-full (bits + noise text) & 0.30 & -- & -- & -- & 0.73 & -0.01 & 0.46 & 0.64 & 0.17 & 0.33 & -0.16 & 0.29 & 0.29 & 0.25 & 0.35 & 0.28 & 0.65 & 0.51 & 0.49 & 0.38 & 0.23 & 0.23 & 0.12 & 0.21 & 0.28 & 0.76 & 0.66 & 0.39 & 0.03 & 0.63 & 0.05 & -0.05 & -0.08 & 0.05 & 0.07 & 0.28 & 0.06 & 0.07 & -0.02 & 0.12 & 0.05 & 0.09 & 0.47 & 0.06 & 0.54 & 0.70 & 0.41 \\
\quad PCA32-ans (shuffled text) & -0.02 & 0.03 & 0.02 & 0.04 & -0.05 & 0.00 & 0.01 & 0.03 & 0.05 & -0.09 & -0.12 & -0.01 & -0.06 & -0.08 & -0.03 & 0.17 & 0.04 & -0.08 & 0.17 & 0.10 & -0.07 & -0.06 & 0.06 & 0.03 & -0.01 & 0.03 & 0.06 & -0.01 & 0.01 & -0.01 & 0.00 & -0.01 & -0.04 & -0.20 & 0.01 & -0.01 & 0.05 & 0.08 & -0.09 & 0.05 & -0.01 & 0.00 & -0.01 & -0.02 & 0.09 & 0.10 & 0.02 \\
\quad PCA32-ans (family-shuffled text) & -0.03 & -- & -- & -- & 0.03 & 0.00 & -0.00 & -0.11 & 0.01 & -0.10 & -0.02 & -0.22 & 0.14 & 0.03 & -0.01 & 0.07 & -0.05 & -0.09 & 0.07 & 0.08 & 0.08 & 0.10 & 0.05 & 0.03 & -0.06 & 0.06 & 0.09 & -0.01 & 0.00 & -0.05 & 0.05 & -0.04 & 0.13 & -0.27 & -0.05 & -0.03 & -0.02 & 0.01 & -0.01 & 0.09 & 0.04 & 0.05 & 0.01 & 0.08 & -0.13 & -0.22 & 0.02 \\
\quad BULLPEN-bits+ans-full (shuffled text) & 0.44 & 0.74 & 0.42 & 0.80 & 0.92 & 0.06 & \underline{0.54} & \underline{0.80} & 0.39 & 0.31 & -0.07 & 0.79 & 0.81 & 0.36 & 0.50 & 0.52 & 0.70 & 0.57 & 0.65 & 0.60 & 0.48 & 0.47 & 0.27 & 0.30 & 0.40 & 0.79 & 0.69 & 0.35 & 0.04 & 0.62 & 0.08 & -0.02 & -0.01 & 0.30 & 0.28 & 0.45 & \underline{0.30} & 0.27 & 0.11 & 0.30 & 0.18 & 0.31 & 0.59 & 0.17 & 0.80 & 0.76 & 0.50 \\
\quad BULLPEN-fused-full (shuffled text) & 0.44 & 0.73 & 0.45 & 0.80 & 0.92 & 0.11 & 0.50 & 0.68 & 0.43 & 0.32 & -0.02 & 0.72 & 0.83 & 0.39 & 0.56 & 0.36 & 0.64 & 0.58 & 0.43 & 0.45 & 0.44 & 0.47 & 0.21 & 0.29 & 0.38 & 0.75 & 0.59 & 0.22 & 0.03 & 0.51 & 0.17 & 0.01 & 0.07 & 0.66 & 0.31 & 0.39 & 0.20 & 0.28 & 0.04 & 0.33 & 0.21 & 0.34 & 0.57 & 0.12 & 0.76 & 0.73 & 0.48 \\
\quad BULLPEN-bits+ans-full (family-shuffled text) & 0.45 & -- & -- & -- & 0.91 & 0.08 & \underline{0.54} & 0.78 & 0.35 & 0.31 & 0.00 & 0.48 & 0.79 & 0.32 & 0.54 & 0.56 & 0.68 & 0.49 & 0.64 & 0.50 & 0.48 & 0.42 & 0.28 & 0.29 & 0.42 & 0.80 & 0.69 & 0.36 & 0.03 & 0.62 & 0.08 & -0.02 & -0.02 & -0.11 & 0.29 & 0.43 & 0.29 & 0.16 & 0.12 & 0.29 & 0.23 & 0.32 & 0.60 & 0.22 & 0.72 & 0.67 & 0.49 \\
\quad BULLPEN-fused-full (family-shuffled text) & 0.45 & -- & -- & -- & 0.93 & \underline{0.12} & 0.50 & 0.68 & 0.30 & 0.27 & 0.13 & 0.65 & 0.60 & 0.28 & 0.59 & 0.41 & 0.61 & 0.45 & 0.39 & 0.25 & 0.44 & 0.47 & 0.29 & 0.30 & 0.37 & 0.74 & 0.60 & 0.28 & 0.03 & 0.52 & 0.17 & 0.05 & -0.02 & -0.37 & 0.31 & 0.39 & 0.23 & 0.21 & 0.14 & 0.33 & 0.23 & 0.36 & 0.58 & 0.18 & 0.68 & 0.66 & 0.47 \\
\quad BULLPEN-zConcat-budget (shuffled text) & 0.39 & -- & -- & -- & 0.79 & 0.04 & 0.43 & 0.63 & 0.23 & 0.23 & 0.05 & 0.18 & 0.50 & 0.24 & 0.40 & 0.18 & 0.51 & 0.39 & 0.25 & 0.29 & 0.23 & 0.19 & 0.07 & 0.11 & -0.01 & 0.62 & 0.55 & 0.11 & 0.02 & 0.46 & 0.07 & 0.00 & -0.03 & 0.17 & 0.11 & 0.21 & 0.18 & 0.13 & 0.04 & 0.29 & 0.13 & 0.20 & 0.53 & 0.06 & 0.62 & 0.75 & 0.31 \\
\quad BULLPEN-CCAfused-full (shuffled text) & 0.45 & -- & -- & -- & 0.85 & 0.02 & 0.40 & 0.63 & 0.32 & 0.24 & -0.04 & 0.42 & 0.24 & 0.15 & 0.32 & 0.16 & 0.48 & 0.32 & 0.27 & 0.33 & 0.31 & 0.24 & 0.10 & 0.17 & 0.03 & 0.60 & 0.47 & 0.24 & 0.01 & 0.40 & 0.14 & -0.02 & -0.07 & -0.02 & 0.03 & 0.19 & 0.11 & 0.09 & 0.03 & 0.24 & 0.15 & 0.13 & 0.46 & 0.07 & 0.52 & 0.75 & 0.31 \\
\quad BULLPEN-ans-full (family-shuffled text) & -0.02 & -- & -- & -- & -0.02 & 0.01 & -0.00 & -0.08 & 0.02 & -0.00 & 0.03 & -0.38 & -0.02 & 0.07 & 0.06 & -0.02 & -0.04 & 0.03 & 0.20 & 0.11 & 0.12 & 0.07 & 0.25 & 0.18 & 0.09 & 0.11 & 0.11 & 0.03 & 0.00 & -0.08 & -0.00 & -0.06 & -0.28 & -0.15 & -0.10 & -0.05 & 0.09 & -0.04 & 0.00 & 0.09 & 0.08 & 0.11 & -0.02 & 0.12 & 0.07 & -0.25 & -0.04 \\
\quad BULLPEN-ans-full (shuffled text) & -0.02 & 0.05 & 0.05 & 0.02 & 0.04 & -0.00 & 0.02 & 0.02 & 0.04 & -0.11 & -0.08 & -0.18 & -0.28 & 0.01 & 0.02 & -0.02 & 0.02 & -0.00 & -0.12 & -0.10 & -0.04 & -0.08 & -0.02 & -0.05 & -0.06 & 0.07 & 0.05 & -0.00 & 0.00 & -0.00 & -0.05 & -0.03 & -0.22 & 0.12 & 0.01 & -0.02 & 0.07 & 0.04 & 0.05 & 0.09 & 0.01 & 0.03 & 0.00 & 0.05 & 0.29 & 0.06 & 0.01 \\
\midrule
\multicolumn{48}{l}{\textit{bits}} \\
\quad IRT-2PL & 0.35 & \textbf{0.80} & \textbf{0.76} & \textbf{0.83} & 0.77 & -0.00 & \underline{0.54} & 0.69 & 0.23 & \textbf{0.52} & -0.04 & 0.77 & \underline{0.95} & 0.45 & 0.49 & \underline{0.64} & 0.75 & \underline{0.70} & \textbf{0.78} & \underline{0.71} & 0.55 & 0.54 & \textbf{0.51} & 0.50 & 0.57 & \textbf{0.90} & \textbf{0.81} & 0.33 & 0.03 & 0.71 & -0.03 & 0.02 & 0.02 & 0.04 & 0.06 & 0.37 & 0.03 & 0.21 & 0.01 & 0.35 & 0.05 & 0.16 & 0.57 & 0.10 & \underline{0.95} & 0.49 & \underline{0.56} \\
\quad PCA32-bits (centered) & \textbf{0.51} & 0.67 & 0.49 & 0.77 & 0.88 & \textbf{0.13} & 0.46 & 0.68 & 0.30 & 0.25 & 0.03 & 0.66 & 0.66 & 0.44 & 0.49 & 0.03 & 0.63 & 0.53 & 0.19 & 0.28 & 0.40 & 0.42 & 0.13 & 0.17 & 0.31 & 0.61 & 0.49 & 0.32 & 0.02 & 0.51 & 0.30 & 0.02 & 0.25 & 0.30 & 0.36 & 0.38 & 0.14 & 0.18 & 0.13 & 0.42 & 0.30 & 0.35 & 0.54 & 0.20 & 0.69 & 0.67 & 0.42 \\
\quad PCA32-bits-budget & \underline{0.50} & 0.67 & 0.47 & 0.78 & 0.87 & 0.11 & 0.46 & 0.73 & 0.27 & 0.22 & 0.13 & 0.76 & 0.54 & 0.25 & 0.46 & 0.33 & 0.59 & 0.35 & 0.22 & 0.24 & 0.45 & 0.46 & 0.15 & 0.21 & 0.21 & 0.62 & 0.56 & 0.18 & 0.01 & 0.50 & 0.16 & 0.05 & 0.09 & 0.66 & 0.34 & 0.37 & 0.14 & 0.18 & 0.05 & 0.32 & 0.23 & 0.27 & 0.53 & 0.17 & 0.65 & 0.60 & 0.39 \\
\quad PCA32-bits & \textbf{0.51} & 0.67 & 0.48 & 0.76 & 0.88 & \underline{0.12} & 0.45 & 0.70 & 0.30 & 0.21 & -0.11 & 0.19 & 0.58 & 0.37 & 0.48 & 0.11 & 0.58 & 0.43 & 0.27 & 0.27 & 0.40 & 0.44 & 0.23 & 0.28 & 0.39 & 0.55 & 0.45 & 0.31 & 0.02 & 0.51 & 0.26 & 0.07 & 0.09 & -0.03 & 0.38 & 0.39 & 0.18 & 0.17 & 0.14 & 0.36 & 0.31 & 0.29 & 0.53 & 0.20 & 0.70 & 0.71 & 0.43 \\
\quad PCA32-bits-budget (centered) & \underline{0.50} & -- & -- & -- & 0.87 & 0.11 & 0.46 & 0.72 & 0.32 & 0.19 & 0.14 & 0.71 & 0.43 & 0.22 & 0.45 & 0.32 & 0.59 & 0.39 & 0.28 & 0.32 & 0.46 & 0.45 & 0.13 & 0.27 & 0.26 & 0.64 & 0.56 & 0.19 & 0.01 & 0.51 & 0.17 & 0.08 & -0.05 & 0.69 & 0.37 & 0.38 & 0.16 & 0.15 & 0.06 & 0.32 & 0.25 & 0.26 & 0.55 & 0.17 & 0.65 & 0.72 & 0.38 \\
\quad BULLPEN-AE-full & 0.44 & -- & -- & -- & \textbf{0.96} & 0.04 & 0.53 & 0.59 & 0.34 & 0.36 & -0.01 & 0.77 & 0.82 & 0.37 & 0.47 & 0.50 & 0.72 & 0.61 & \textbf{0.78} & 0.69 & 0.58 & 0.56 & 0.39 & 0.51 & 0.44 & 0.84 & 0.76 & 0.54 & 0.03 & 0.67 & 0.13 & -0.04 & 0.06 & 0.10 & 0.12 & 0.38 & 0.12 & 0.20 & 0.02 & 0.26 & 0.11 & 0.25 & 0.57 & 0.11 & 0.75 & 0.76 & 0.54 \\
\quad BULLPEN-AE-budget & 0.43 & -- & -- & -- & \underline{0.95} & 0.01 & 0.53 & 0.54 & 0.33 & 0.41 & -0.02 & 0.70 & 0.78 & 0.33 & 0.56 & 0.58 & 0.72 & 0.62 & \underline{0.73} & 0.58 & 0.56 & 0.54 & 0.30 & 0.48 & 0.35 & 0.81 & 0.73 & 0.55 & 0.03 & 0.67 & 0.15 & -0.05 & 0.11 & 0.11 & 0.13 & 0.35 & 0.15 & 0.19 & 0.03 & 0.23 & 0.08 & 0.24 & 0.55 & 0.11 & 0.77 & 0.73 & 0.49 \\
\quad BULLPEN-LogPCA-full & 0.47 & -- & -- & -- & 0.76 & \underline{0.12} & 0.47 & 0.66 & 0.32 & 0.30 & 0.09 & 0.32 & 0.65 & 0.41 & 0.48 & 0.15 & 0.70 & 0.48 & 0.33 & 0.30 & 0.53 & 0.50 & 0.27 & 0.27 & 0.35 & 0.65 & 0.56 & 0.38 & 0.02 & 0.49 & 0.18 & 0.04 & -0.13 & 0.42 & \textbf{0.48} & 0.42 & 0.22 & 0.19 & \underline{0.17} & 0.42 & 0.30 & 0.38 & 0.54 & 0.15 & 0.80 & 0.78 & 0.50 \\
\midrule
\multicolumn{48}{l}{\textit{bits + question text}} \\
\quad JE-IRT (retrained) & 0.48 & \underline{0.78} & \underline{0.63} & \underline{0.82} & 0.91 & 0.11 & 0.52 & 0.66 & 0.51 & \underline{0.48} & \underline{0.26} & \underline{0.86} & \textbf{0.97} & \textbf{0.49} & 0.58 & 0.56 & \textbf{0.78} & \textbf{0.71} & \underline{0.73} & \textbf{0.75} & 0.56 & 0.56 & 0.49 & 0.59 & 0.56 & 0.87 & 0.76 & 0.36 & 0.05 & 0.67 & 0.20 & 0.05 & -0.04 & 0.11 & 0.20 & 0.42 & 0.27 & 0.19 & 0.16 & 0.37 & 0.28 & 0.33 & 0.62 & 0.21 & \textbf{1.00} & 0.73 & 0.50 \\
\quad EmbedLLM (retrained) & 0.47 & \underline{0.78} & 0.58 & 0.81 & 0.93 & 0.11 & 0.51 & 0.55 & \textbf{0.54} & 0.45 & 0.12 & 0.67 & 0.73 & \underline{0.46} & \textbf{0.69} & 0.55 & 0.71 & 0.66 & 0.68 & \textbf{0.75} & 0.48 & 0.49 & 0.46 & 0.43 & 0.44 & 0.84 & 0.73 & 0.43 & 0.04 & 0.64 & 0.15 & 0.07 & \underline{0.32} & 0.29 & 0.22 & 0.40 & 0.22 & 0.27 & 0.10 & 0.39 & 0.26 & 0.32 & 0.61 & 0.15 & 0.83 & 0.79 & 0.49 \\
\quad RouteLLM-MF (retrained) & 0.43 & 0.75 & 0.51 & 0.80 & 0.90 & 0.07 & 0.48 & 0.64 & 0.45 & 0.42 & \textbf{0.28} & 0.64 & 0.77 & 0.43 & 0.64 & 0.59 & 0.69 & 0.64 & 0.70 & 0.67 & 0.52 & 0.50 & 0.39 & 0.45 & 0.46 & \underline{0.88} & \underline{0.78} & 0.48 & 0.02 & 0.62 & 0.13 & 0.04 & -0.09 & 0.08 & 0.17 & 0.36 & 0.12 & 0.26 & 0.08 & 0.34 & 0.24 & 0.28 & 0.56 & 0.15 & 0.84 & 0.78 & 0.47 \\
\quad IrtNet (retrained) & 0.42 & 0.76 & 0.58 & 0.79 & 0.84 & 0.08 & 0.47 & 0.61 & 0.48 & 0.41 & 0.12 & 0.36 & 0.67 & 0.43 & 0.65 & 0.46 & 0.74 & 0.66 & 0.60 & 0.62 & 0.42 & 0.44 & 0.22 & 0.37 & 0.40 & 0.80 & 0.67 & 0.40 & 0.04 & 0.63 & 0.10 & 0.04 & 0.19 & 0.41 & 0.21 & 0.38 & 0.20 & 0.10 & 0.12 & 0.39 & 0.24 & 0.28 & 0.62 & 0.12 & 0.87 & 0.67 & 0.43 \\
\quad LOCUS (retrained) & 0.45 & 0.74 & 0.35 & 0.79 & \textbf{0.96} & 0.11 & 0.48 & 0.62 & \underline{0.53} & 0.32 & 0.09 & 0.67 & 0.85 & 0.38 & 0.62 & 0.46 & 0.65 & 0.65 & 0.72 & 0.54 & \textbf{0.63} & \textbf{0.62} & 0.20 & 0.34 & 0.24 & \underline{0.88} & 0.77 & 0.39 & 0.04 & 0.59 & 0.24 & -0.02 & -0.24 & 0.07 & 0.21 & 0.38 & 0.12 & 0.21 & 0.13 & 0.23 & 0.29 & 0.26 & 0.60 & 0.17 & 0.73 & \textbf{0.90} & 0.46 \\
\quad LOCUS-32 (retrained, d=32) & 0.45 & -- & -- & -- & 0.92 & 0.07 & \underline{0.54} & 0.58 & \textbf{0.54} & 0.36 & 0.14 & 0.62 & 0.80 & 0.39 & \underline{0.66} & 0.51 & 0.75 & 0.65 & 0.71 & 0.65 & 0.60 & \underline{0.59} & 0.33 & 0.29 & 0.20 & 0.83 & 0.74 & 0.45 & 0.04 & 0.63 & 0.20 & -0.05 & -0.06 & 0.04 & 0.14 & 0.38 & 0.08 & 0.18 & 0.05 & 0.28 & 0.18 & 0.29 & 0.58 & 0.16 & 0.71 & \underline{0.88} & 0.49 \\
\quad RouteLLM-MF-32 (retrained, d=32) & 0.43 & -- & -- & -- & 0.88 & 0.04 & 0.53 & 0.66 & 0.40 & 0.40 & 0.24 & 0.59 & 0.66 & 0.40 & 0.57 & 0.59 & 0.69 & 0.58 & 0.68 & 0.65 & 0.50 & 0.52 & 0.22 & 0.41 & 0.46 & 0.81 & 0.73 & 0.43 & 0.03 & 0.65 & 0.07 & 0.01 & -0.11 & -0.14 & 0.14 & 0.38 & 0.14 & 0.21 & 0.10 & 0.31 & 0.18 & 0.27 & 0.56 & 0.14 & 0.72 & 0.79 & 0.50 \\
\quad IRT-Router (retrained) & 0.41 & 0.69 & 0.56 & 0.77 & 0.81 & 0.06 & 0.49 & \textbf{0.82} & 0.30 & 0.29 & 0.06 & 0.72 & 0.48 & 0.40 & 0.61 & 0.42 & 0.71 & 0.63 & 0.45 & 0.42 & 0.30 & 0.30 & 0.17 & 0.32 & 0.42 & 0.80 & 0.67 & 0.34 & 0.02 & 0.58 & 0.17 & 0.05 & -0.33 & -0.12 & 0.26 & 0.43 & 0.24 & 0.16 & 0.09 & 0.28 & 0.25 & 0.28 & 0.60 & 0.15 & 0.79 & 0.71 & 0.47 \\
\quad PCA32-bits+q & \underline{0.50} & 0.67 & 0.49 & 0.78 & 0.88 & 0.11 & 0.45 & 0.72 & 0.33 & 0.29 & 0.12 & 0.06 & 0.59 & 0.40 & 0.51 & 0.14 & 0.58 & 0.49 & 0.29 & 0.36 & 0.35 & 0.40 & 0.14 & 0.29 & 0.39 & 0.63 & 0.55 & 0.23 & 0.02 & 0.49 & 0.29 & 0.07 & 0.02 & -0.10 & 0.35 & 0.34 & 0.17 & 0.20 & 0.11 & 0.36 & 0.25 & 0.31 & 0.52 & 0.13 & 0.67 & 0.58 & 0.43 \\
\quad PCA32-bits+q-budget & 0.49 & -- & -- & -- & 0.86 & 0.10 & 0.45 & 0.70 & 0.21 & 0.35 & 0.22 & 0.82 & 0.46 & 0.22 & 0.47 & 0.28 & 0.62 & 0.42 & 0.22 & 0.24 & 0.44 & 0.41 & 0.20 & 0.26 & 0.27 & 0.58 & 0.49 & 0.19 & 0.03 & 0.51 & 0.20 & 0.05 & 0.03 & 0.71 & 0.28 & 0.37 & 0.14 & 0.13 & 0.15 & 0.26 & 0.23 & 0.28 & 0.56 & 0.14 & 0.65 & 0.69 & 0.37 \\
\quad BULLPEN-QMF-budget & 0.48 & -- & -- & -- & 0.89 & 0.07 & \textbf{0.55} & 0.61 & 0.40 & 0.47 & 0.09 & 0.80 & 0.80 & 0.41 & 0.57 & \textbf{0.71} & \underline{0.77} & 0.66 & 0.65 & 0.67 & \underline{0.61} & 0.58 & 0.37 & 0.40 & 0.46 & 0.84 & 0.77 & 0.36 & 0.03 & 0.70 & 0.05 & -0.01 & -0.03 & 0.16 & 0.11 & \textbf{0.48} & 0.15 & 0.20 & -0.03 & 0.34 & 0.16 & 0.29 & 0.62 & 0.13 & 0.78 & 0.55 & \textbf{0.58} \\
\quad BULLPEN-bits+q-full & 0.47 & 0.73 & 0.50 & 0.81 & 0.93 & \underline{0.12} & 0.51 & 0.76 & 0.38 & 0.30 & 0.04 & 0.50 & 0.59 & 0.40 & 0.57 & 0.36 & 0.71 & 0.60 & 0.33 & 0.37 & 0.48 & 0.52 & 0.09 & 0.27 & 0.23 & 0.70 & 0.60 & 0.26 & 0.03 & 0.50 & 0.23 & 0.09 & 0.00 & 0.75 & \underline{0.41} & 0.39 & 0.20 & 0.20 & 0.11 & 0.40 & 0.29 & 0.36 & 0.61 & 0.15 & 0.79 & 0.78 & 0.48 \\
\quad BULLPEN-bits+q-budget & 0.47 & 0.74 & 0.49 & 0.80 & 0.92 & 0.11 & 0.51 & 0.77 & 0.41 & 0.26 & 0.08 & \textbf{0.88} & 0.59 & 0.19 & 0.46 & 0.43 & 0.70 & 0.45 & 0.37 & 0.33 & 0.51 & 0.46 & 0.19 & 0.23 & 0.36 & 0.70 & 0.56 & 0.26 & 0.03 & 0.57 & 0.21 & 0.06 & 0.05 & \underline{0.76} & 0.34 & 0.37 & 0.20 & 0.15 & 0.14 & 0.38 & 0.25 & 0.32 & 0.61 & 0.14 & 0.74 & 0.66 & 0.43 \\
\quad BULLPEN-QMF-full & 0.46 & -- & -- & -- & 0.79 & 0.02 & 0.51 & 0.54 & 0.42 & 0.40 & -0.09 & 0.83 & 0.73 & 0.39 & 0.48 & 0.53 & 0.74 & 0.58 & 0.61 & 0.55 & 0.59 & 0.55 & 0.35 & 0.38 & 0.34 & 0.79 & 0.71 & 0.53 & 0.03 & 0.62 & 0.07 & 0.02 & 0.13 & 0.27 & 0.19 & 0.27 & 0.09 & 0.22 & 0.03 & 0.25 & 0.12 & 0.19 & 0.52 & 0.09 & 0.72 & 0.83 & 0.42 \\
\midrule
\multicolumn{48}{l}{\textit{answer text}} \\
\quad LLM DNA (recomputed) & 0.32 & -- & -- & -- & 0.50 & 0.06 & 0.40 & 0.61 & 0.20 & 0.31 & 0.03 & 0.37 & 0.55 & 0.29 & 0.48 & 0.22 & 0.42 & 0.38 & 0.53 & 0.47 & 0.31 & 0.30 & 0.45 & 0.53 & 0.53 & 0.72 & 0.69 & 0.87 & 0.05 & 0.66 & 0.48 & 0.07 & 0.29 & -0.01 & 0.17 & 0.38 & 0.18 & 0.29 & 0.08 & 0.38 & 0.28 & 0.34 & 0.52 & 0.18 & 0.53 & 0.75 & 0.34 \\
\quad PCA32-ans & 0.37 & 0.58 & 0.21 & 0.56 & 0.49 & 0.01 & 0.38 & 0.48 & 0.08 & 0.08 & 0.03 & 0.23 & 0.68 & 0.12 & 0.24 & 0.23 & 0.39 & 0.15 & 0.51 & 0.42 & 0.33 & 0.36 & 0.21 & 0.34 & 0.43 & 0.41 & 0.42 & 0.77 & 0.16 & 0.67 & 0.75 & 0.16 & 0.17 & 0.75 & 0.22 & 0.37 & 0.19 & 0.24 & -0.05 & 0.44 & \underline{0.44} & 0.42 & 0.58 & 0.29 & 0.43 & 0.62 & 0.15 \\
\quad PCA32-ans-budget & 0.37 & -- & -- & -- & 0.50 & 0.02 & 0.38 & 0.51 & 0.08 & 0.05 & 0.10 & 0.21 & 0.50 & 0.16 & 0.19 & 0.14 & 0.34 & 0.12 & 0.41 & 0.42 & 0.34 & 0.38 & 0.18 & 0.30 & 0.35 & 0.41 & 0.40 & 0.76 & 0.17 & 0.70 & 0.73 & 0.16 & -0.05 & 0.72 & 0.21 & 0.34 & 0.23 & 0.18 & -0.05 & 0.44 & 0.38 & 0.41 & 0.57 & \underline{0.30} & 0.41 & 0.53 & 0.15 \\
\quad LLMmap (retrained) & 0.17 & -- & -- & -- & 0.27 & 0.04 & 0.22 & 0.34 & 0.13 & 0.36 & 0.18 & -0.18 & -0.21 & 0.09 & 0.12 & 0.04 & 0.24 & 0.23 & 0.07 & 0.18 & 0.07 & 0.05 & 0.06 & 0.06 & 0.06 & 0.43 & 0.42 & 0.34 & 0.03 & 0.40 & 0.44 & 0.15 & 0.15 & -- & 0.12 & 0.30 & 0.14 & 0.05 & 0.16 & 0.39 & 0.23 & 0.37 & 0.49 & \underline{0.30} & 0.54 & 0.38 & 0.20 \\
\quad BULLPEN-ans-full & 0.36 & 0.65 & 0.50 & 0.64 & 0.69 & 0.09 & 0.39 & 0.52 & 0.29 & 0.35 & 0.12 & 0.33 & 0.49 & 0.21 & 0.45 & 0.15 & 0.54 & 0.30 & 0.59 & 0.60 & 0.41 & 0.37 & \underline{0.50} & \textbf{0.64} & \textbf{0.74} & 0.71 & 0.68 & \textbf{0.90} & \textbf{0.22} & 0.77 & \underline{0.80} & \underline{0.31} & 0.15 & 0.60 & 0.25 & 0.38 & 0.25 & 0.28 & 0.01 & 0.52 & 0.41 & 0.41 & 0.64 & 0.27 & 0.70 & 0.77 & 0.25 \\
\quad BULLPEN-ans-budget & 0.36 & 0.64 & 0.50 & 0.62 & 0.67 & 0.09 & 0.39 & 0.50 & 0.32 & 0.36 & 0.15 & 0.30 & 0.55 & 0.26 & 0.45 & 0.19 & 0.53 & 0.33 & 0.58 & 0.54 & 0.37 & 0.35 & \underline{0.50} & \underline{0.61} & \underline{0.70} & 0.69 & 0.68 & \underline{0.89} & \underline{0.21} & 0.78 & 0.78 & 0.30 & 0.03 & 0.51 & 0.22 & 0.40 & \textbf{0.31} & 0.29 & 0.06 & \underline{0.53} & 0.42 & 0.42 & 0.62 & 0.24 & 0.72 & 0.66 & 0.25 \\
\quad BULLPEN-AE-full & 0.30 & -- & -- & -- & 0.58 & 0.01 & 0.41 & 0.47 & 0.16 & 0.24 & -0.11 & 0.53 & 0.69 & 0.18 & 0.34 & 0.14 & 0.49 & 0.21 & 0.54 & 0.55 & 0.32 & 0.32 & 0.41 & 0.53 & 0.59 & 0.67 & 0.66 & 0.72 & 0.06 & \underline{0.86} & 0.31 & 0.05 & \textbf{0.33} & 0.15 & 0.11 & 0.34 & 0.08 & \underline{0.36} & -0.04 & 0.27 & 0.18 & 0.29 & 0.49 & 0.22 & 0.51 & 0.60 & 0.30 \\
\quad BULLPEN-AE-budget & 0.30 & -- & -- & -- & 0.58 & 0.01 & 0.41 & 0.46 & 0.17 & 0.31 & -0.07 & 0.49 & 0.65 & 0.14 & 0.34 & 0.15 & 0.51 & 0.20 & 0.52 & 0.49 & 0.23 & 0.22 & 0.49 & 0.60 & 0.55 & 0.63 & 0.59 & 0.70 & 0.05 & 0.85 & 0.25 & -0.02 & 0.18 & 0.33 & 0.14 & 0.34 & 0.13 & \textbf{0.37} & 0.02 & 0.29 & 0.16 & 0.27 & 0.50 & 0.20 & 0.55 & 0.53 & 0.29 \\
\midrule
\multicolumn{48}{l}{\textit{answer statistics}} \\
\quad answer statistics (d=6) & 0.31 & -- & -- & -- & 0.30 & -0.03 & 0.36 & 0.14 & 0.13 & 0.29 & -0.13 & 0.28 & 0.48 & 0.09 & 0.35 & 0.11 & 0.29 & 0.24 & 0.25 & 0.34 & 0.43 & 0.36 & 0.26 & 0.32 & 0.35 & 0.59 & 0.64 & 0.52 & 0.14 & \textbf{0.89} & \textbf{0.91} & \textbf{0.47} & -0.08 & 0.12 & 0.11 & 0.28 & 0.06 & 0.27 & \textbf{0.32} & 0.34 & 0.13 & 0.36 & 0.55 & 0.17 & 0.32 & 0.64 & 0.24 \\
\midrule
\multicolumn{48}{l}{\textit{bits + answer text}} \\
\quad PCA32-bits+ans & 0.47 & 0.68 & 0.39 & 0.76 & 0.75 & 0.10 & 0.45 & 0.64 & 0.30 & 0.20 & 0.09 & 0.64 & 0.70 & 0.42 & 0.48 & 0.37 & 0.56 & 0.42 & 0.53 & 0.41 & 0.42 & 0.40 & 0.22 & 0.39 & 0.49 & 0.63 & 0.55 & 0.67 & 0.13 & 0.62 & 0.69 & 0.11 & 0.17 & 0.62 & 0.34 & 0.37 & 0.27 & 0.16 & 0.10 & 0.49 & \underline{0.44} & \textbf{0.45} & 0.57 & 0.25 & 0.69 & 0.75 & 0.37 \\
\quad PCA32-bits+ans-budget & 0.47 & -- & -- & -- & 0.82 & 0.09 & 0.44 & 0.64 & 0.31 & 0.17 & 0.06 & 0.34 & 0.68 & 0.27 & 0.40 & 0.33 & 0.55 & 0.38 & 0.60 & 0.46 & 0.40 & 0.41 & 0.11 & 0.33 & 0.32 & 0.57 & 0.53 & 0.64 & 0.15 & 0.63 & 0.67 & 0.18 & 0.16 & 0.44 & 0.27 & 0.35 & 0.23 & 0.25 & 0.08 & 0.47 & 0.39 & 0.36 & 0.61 & 0.27 & 0.58 & 0.69 & 0.34 \\
\quad BULLPEN-bits+ans-full & 0.46 & 0.71 & 0.44 & 0.78 & 0.86 & \underline{0.12} & 0.50 & 0.68 & 0.42 & 0.24 & 0.13 & 0.56 & 0.66 & 0.35 & 0.57 & 0.26 & 0.66 & 0.45 & 0.52 & 0.59 & 0.49 & 0.48 & 0.26 & 0.37 & 0.54 & 0.73 & 0.63 & 0.72 & 0.14 & 0.71 & 0.68 & 0.20 & 0.11 & 0.37 & 0.33 & \underline{0.47} & 0.17 & 0.24 & 0.11 & 0.47 & 0.41 & \textbf{0.45} & 0.63 & 0.27 & 0.72 & 0.83 & 0.44 \\
\quad BULLPEN-bits+ans-budget & 0.46 & 0.70 & 0.44 & 0.78 & 0.85 & 0.11 & 0.49 & 0.71 & 0.39 & 0.28 & 0.17 & 0.52 & 0.74 & 0.36 & 0.52 & 0.25 & 0.55 & 0.47 & 0.58 & 0.62 & 0.41 & 0.38 & 0.22 & 0.41 & 0.53 & 0.72 & 0.64 & 0.73 & 0.14 & 0.71 & 0.67 & 0.21 & 0.22 & 0.59 & 0.28 & 0.44 & 0.20 & 0.22 & 0.05 & 0.42 & 0.35 & 0.35 & 0.63 & 0.26 & 0.69 & 0.74 & 0.39 \\
\quad BULLPEN-CCA-full & 0.47 & -- & -- & -- & 0.73 & 0.11 & 0.45 & 0.60 & 0.32 & 0.19 & 0.13 & 0.54 & 0.86 & 0.40 & 0.49 & 0.26 & 0.51 & 0.44 & 0.55 & 0.49 & 0.40 & 0.36 & 0.38 & 0.49 & 0.56 & 0.54 & 0.53 & 0.72 & 0.10 & 0.57 & 0.63 & 0.18 & 0.10 & 0.68 & 0.30 & 0.38 & 0.19 & 0.10 & 0.03 & \textbf{0.55} & \textbf{0.45} & 0.40 & 0.58 & \textbf{0.34} & 0.65 & 0.77 & 0.38 \\
\quad BULLPEN-zConcat-budget & 0.45 & -- & -- & -- & 0.83 & 0.07 & 0.44 & 0.60 & 0.30 & 0.22 & 0.10 & 0.40 & 0.61 & 0.41 & 0.41 & 0.33 & 0.55 & 0.47 & 0.55 & 0.36 & 0.40 & 0.42 & 0.21 & 0.23 & 0.34 & 0.61 & 0.54 & 0.72 & 0.17 & 0.62 & 0.71 & 0.15 & 0.17 & 0.70 & 0.26 & 0.43 & 0.24 & 0.22 & 0.04 & 0.47 & 0.42 & 0.43 & \textbf{0.66} & 0.28 & 0.62 & 0.69 & 0.28 \\
\quad BULLPEN-AE-full & 0.35 & -- & -- & -- & 0.75 & 0.00 & 0.46 & 0.47 & 0.12 & 0.36 & -0.02 & 0.55 & 0.76 & 0.22 & 0.37 & 0.20 & 0.60 & 0.29 & 0.52 & 0.65 & 0.41 & 0.35 & 0.36 & 0.50 & 0.51 & 0.76 & 0.70 & 0.70 & 0.06 & 0.85 & 0.22 & -0.02 & 0.03 & 0.29 & 0.10 & 0.34 & 0.17 & 0.31 & 0.00 & 0.30 & 0.19 & 0.30 & 0.50 & 0.20 & 0.65 & 0.67 & 0.40 \\
\quad BULLPEN-zSlice-budget & 0.43 & -- & -- & -- & 0.79 & -0.00 & 0.43 & 0.65 & 0.15 & 0.18 & 0.16 & 0.01 & 0.33 & 0.40 & 0.21 & 0.48 & 0.59 & 0.45 & 0.50 & 0.40 & 0.45 & 0.42 & 0.32 & 0.36 & 0.35 & 0.64 & 0.57 & 0.69 & 0.09 & 0.66 & 0.62 & 0.12 & 0.21 & 0.65 & 0.11 & 0.32 & 0.16 & 0.16 & -0.06 & 0.38 & 0.39 & 0.33 & 0.52 & 0.18 & 0.64 & 0.68 & 0.27 \\
\midrule
\multicolumn{48}{l}{\textit{bits + question text + answer text}} \\
\quad PCA32-bits+q+ans & 0.48 & 0.67 & 0.39 & 0.77 & 0.82 & 0.10 & 0.44 & 0.69 & 0.31 & 0.20 & 0.08 & 0.71 & 0.58 & 0.40 & 0.47 & 0.24 & 0.60 & 0.48 & 0.57 & 0.44 & 0.44 & 0.43 & 0.30 & 0.39 & 0.50 & 0.61 & 0.55 & 0.61 & 0.14 & 0.62 & 0.73 & 0.12 & 0.20 & 0.67 & 0.35 & 0.37 & 0.22 & 0.21 & 0.07 & 0.52 & \textbf{0.45} & \underline{0.44} & 0.56 & 0.21 & 0.68 & 0.74 & 0.34 \\
\quad PCA32-bits+q+ans-budget & 0.47 & -- & -- & -- & 0.84 & 0.09 & 0.44 & 0.66 & 0.30 & 0.21 & 0.08 & 0.60 & 0.53 & 0.26 & 0.40 & 0.27 & 0.60 & 0.36 & 0.54 & 0.44 & 0.37 & 0.40 & 0.21 & 0.32 & 0.28 & 0.55 & 0.52 & 0.63 & 0.13 & 0.62 & 0.67 & 0.18 & 0.20 & 0.56 & 0.27 & 0.41 & 0.25 & 0.24 & 0.05 & 0.46 & 0.35 & 0.37 & 0.62 & 0.24 & 0.62 & 0.72 & 0.34 \\
\quad BULLPEN-fused-full & 0.47 & 0.74 & 0.48 & 0.81 & 0.90 & \underline{0.12} & 0.52 & 0.73 & 0.42 & 0.34 & 0.10 & 0.54 & 0.67 & 0.28 & 0.57 & 0.35 & 0.71 & 0.50 & 0.62 & 0.62 & 0.48 & 0.48 & 0.36 & 0.44 & 0.50 & 0.78 & 0.65 & 0.72 & 0.12 & 0.70 & 0.71 & 0.19 & 0.10 & 0.71 & 0.38 & 0.43 & 0.22 & 0.24 & 0.15 & 0.43 & 0.38 & \textbf{0.45} & \underline{0.65} & 0.22 & 0.83 & 0.76 & 0.48 \\
\quad BULLPEN-fused-budget & 0.47 & 0.73 & 0.45 & 0.79 & 0.89 & 0.11 & 0.52 & 0.72 & 0.44 & 0.28 & 0.15 & 0.57 & 0.65 & 0.35 & 0.52 & 0.45 & 0.63 & 0.45 & 0.52 & 0.44 & 0.44 & 0.44 & 0.26 & 0.45 & 0.52 & 0.75 & 0.69 & 0.67 & 0.14 & 0.71 & 0.70 & 0.17 & 0.12 & 0.50 & 0.34 & 0.40 & 0.24 & 0.21 & 0.03 & 0.40 & 0.31 & 0.38 & 0.64 & 0.24 & 0.81 & 0.70 & 0.45 \\
\quad BULLPEN-CCAfused-full & 0.38 & -- & -- & -- & 0.80 & 0.05 & 0.43 & 0.63 & 0.27 & 0.24 & 0.08 & 0.44 & 0.77 & 0.28 & 0.43 & 0.32 & 0.57 & 0.52 & 0.46 & 0.43 & 0.44 & 0.43 & 0.18 & 0.34 & 0.36 & 0.53 & 0.47 & 0.72 & 0.16 & 0.64 & 0.69 & 0.14 & 0.13 & \textbf{0.87} & 0.17 & 0.31 & 0.18 & 0.16 & 0.03 & 0.39 & 0.38 & 0.39 & 0.64 & 0.25 & 0.60 & 0.60 & 0.30 \\
\quad BULLPEN-CCAfused-budget & 0.39 & -- & -- & -- & 0.80 & 0.05 & 0.43 & 0.63 & 0.27 & 0.24 & 0.05 & 0.22 & 0.69 & 0.20 & 0.41 & 0.34 & 0.58 & 0.43 & 0.53 & 0.46 & 0.40 & 0.44 & 0.20 & 0.29 & 0.26 & 0.58 & 0.52 & 0.72 & 0.16 & 0.65 & 0.70 & 0.17 & 0.17 & 0.69 & 0.18 & 0.35 & 0.19 & 0.18 & 0.04 & 0.42 & 0.35 & 0.40 & \underline{0.65} & 0.24 & 0.59 & 0.67 & 0.31 \\
\bottomrule
\end{tabular}
  \end{adjustbox}
  \par\smallskip
  \parbox{\textheight}{\scriptsize The floor is the mean of 10 Gaussian draws per fold and seed. Columns grouped by task group (Table~\ref{tab:paper_task_vocabulary}); best per column in bold. --: not scored: the three benchmark-transfer tasks are scored only for Table~\ref{tab:paper_master_table}'s encoders, and LLMmap and LLM DNA skip them too (no per-benchmark refits). Strict protocol: 5 publisher-grouped folds $\times$ 3 seeds.}
\end{sidewaystable}

\begin{table}[t]
  \centering
  \small
  \setlength{\abovecaptionskip}{0pt}\setlength{\belowcaptionskip}{8pt}
  \caption{Table~\ref{tab:paper_master_table}'s embeddings on all 46 tasks, by label source.}
  \label{tab:paper_master_table_by_source}
  \begin{adjustbox}{max width=\linewidth}
\begin{tabular}{l!{\color{black!25}\vrule}rrrrrr!{\color{black!25}\vrule}rr}
\toprule
 & \rotatebox{90}{probe correctness} & \rotatebox{90}{published benchmark score} & \rotatebox{90}{human judgment} & \rotatebox{90}{model judgment} & \rotatebox{90}{answer-text statistic} & \rotatebox{90}{registry record} & \rotatebox{90}{\textbf{all tasks}} & \rotatebox{90}{$M_{\text{eff}}$-weighted} \\
\multicolumn{1}{r}{\textit{tasks per column}} & 11 & 10 & 5 & 4 & 5 & 11 & 46 & 36.9 (eff.) \\
\midrule
\textbf{embedding} \\
\multicolumn{9}{l}{\textit{reference: one number per model}} \\
\quad leaderboard scalar (d=1) & 0.50 & 0.60 & \cellcolor{green!5}\textbf{0.60} & \cellcolor{green!5}\textbf{0.71} & 0.25 & 0.18 & 0.45$_{\pm0.09}$ & 0.43$_{\pm0.10}$ \\
\midrule
\multicolumn{9}{l}{\textit{bits}} \\
\quad PCA32-bits-budget & 0.53 & 0.43 & 0.20 & 0.52 & 0.31 & 0.24 & 0.38$_{\pm0.06}$ & 0.39$_{\pm0.07}$ \\
\quad PCA32-bits & 0.55 & 0.35 & 0.29 & 0.46 & 0.22 & 0.27 & 0.37$_{\pm0.06}$ & 0.37$_{\pm0.07}$ \\
\quad IRT-2PL & 0.55 & 0.62 & \cellcolor{green!5}\textbf{0.61} & \cellcolor{green!5}\textbf{0.70} & 0.21 & \cellcolor{red!6}0.17 & 0.46$_{\pm0.08}$ & 0.44$_{\pm0.10}$ \\
\midrule
\multicolumn{9}{l}{\textit{bits + question text}} \\
\quad EmbedLLM (retrained) & \cellcolor{green!5}\textbf{0.60} & 0.59 & \cellcolor{green!5}\textbf{0.55} & 0.64 & 0.31 & 0.27 & 0.48$_{\pm0.07}$ & 0.47$_{\pm0.07}$ \\
\quad LOCUS (retrained) & 0.58 & 0.54 & 0.41 & \cellcolor{green!5}\textbf{0.72} & 0.25 & 0.24 & 0.45$_{\pm0.07}$ & 0.45$_{\pm0.08}$ \\
\quad RouteLLM-MF (retrained) & 0.57 & 0.59 & \cellcolor{green!5}\textbf{0.54} & \cellcolor{green!5}\textbf{0.67} & 0.27 & 0.23 & 0.47$_{\pm0.07}$ & 0.46$_{\pm0.08}$ \\
\quad IRT-Router (retrained) & 0.55 & 0.51 & 0.35 & 0.52 & \cellcolor{red!6}0.20 & 0.25 & 0.41$_{\pm0.07}$ & 0.41$_{\pm0.08}$ \\
\quad JE-IRT (retrained) & \cellcolor{green!5}\textbf{0.60} & \cellcolor{green!5}\underline{\textbf{0.67}} & \cellcolor{green!5}\textbf{0.62} & \cellcolor{green!5}\textbf{0.69} & 0.28 & 0.28 & \cellcolor{green!5}\textbf{0.52}$_{\pm0.07}$ & \cellcolor{green!5}\textbf{0.51}$_{\pm0.08}$ \\
\quad IrtNet (retrained) & 0.56 & 0.54 & 0.44 & 0.58 & 0.32 & 0.25 & 0.44$_{\pm0.07}$ & 0.43$_{\pm0.07}$ \\
\quad BULLPEN-bits+q-budget & 0.57 & 0.48 & 0.30 & 0.56 & 0.37 & 0.27 & 0.43$_{\pm0.07}$ & 0.44$_{\pm0.07}$ \\
\quad BULLPEN-bits+q-full & \cellcolor{green!5}\textbf{0.59} & 0.49 & 0.26 & 0.58 & 0.37 & 0.29 & 0.43$_{\pm0.07}$ & 0.44$_{\pm0.07}$ \\
\midrule
\multicolumn{9}{l}{\textit{answer text}} \\
\quad LLMmap (retrained) & \cellcolor{red!6}0.22 & \cellcolor{red!6}0.14 & \cellcolor{red!6}0.09 & \cellcolor{red!6}0.24 & 0.33 & 0.23 & \cellcolor{red!6}0.20$_{\pm0.05}$ & \cellcolor{red!6}0.20$_{\pm0.06}$ \\
\quad LLM DNA (recomputed) & 0.40 & 0.36 & \cellcolor{green!5}\textbf{0.50} & 0.51 & 0.41 & 0.26 & 0.38$_{\pm0.06}$ & 0.37$_{\pm0.07}$ \\
\quad BULLPEN-ans-budget & 0.46 & 0.38 & \cellcolor{green!5}\textbf{0.59} & 0.52 & 0.65 & \cellcolor{green!5}\textbf{0.34} & \cellcolor{green!5}\textbf{0.45}$_{\pm0.06}$ & \cellcolor{green!5}\textbf{0.45}$_{\pm0.06}$ \\
\quad BULLPEN-ans-full & 0.47 & 0.36 & \cellcolor{green!5}\textbf{0.61} & 0.54 & \cellcolor{green!5}\underline{\textbf{0.68}} & \cellcolor{green!5}\textbf{0.33} & \cellcolor{green!5}\textbf{0.46}$_{\pm0.06}$ & \cellcolor{green!5}\textbf{0.45}$_{\pm0.07}$ \\
\midrule
\multicolumn{9}{l}{\textit{bits + answer text}} \\
\quad BULLPEN-bits+ans-budget & 0.55 & 0.46 & 0.47 & 0.54 & 0.58 & 0.30 & \cellcolor{green!5}\textbf{0.47}$_{\pm0.06}$ & \cellcolor{green!5}\textbf{0.47}$_{\pm0.06}$ \\
\quad BULLPEN-bits+ans-full & 0.57 & 0.46 & 0.45 & 0.58 & 0.54 & \cellcolor{green!5}\textbf{0.34} & \cellcolor{green!5}\textbf{0.47}$_{\pm0.06}$ & \cellcolor{green!5}\textbf{0.48}$_{\pm0.06}$ \\
\midrule
\multicolumn{9}{l}{\textit{bits + question text + answer text}} \\
\quad BULLPEN-fused-budget & 0.57 & 0.49 & 0.44 & 0.58 & 0.55 & \cellcolor{green!5}\textbf{0.30} & \cellcolor{green!5}\textbf{0.47}$_{\pm0.06}$ & \cellcolor{green!5}\textbf{0.48}$_{\pm0.06}$ \\
\quad BULLPEN-fused-full & \cellcolor{green!5}\textbf{0.58} & 0.49 & 0.51 & 0.60 & 0.61 & \cellcolor{green!5}\textbf{0.33} & \cellcolor{green!5}\textbf{0.50}$_{\pm0.06}$ & \cellcolor{green!5}\textbf{0.50}$_{\pm0.06}$ \\
\bottomrule
\end{tabular}
  \end{adjustbox}
  \par\smallskip
  \parbox{\linewidth}{\scriptsize New and established tasks, grouped by who produced the label: our probe's correctness, a published benchmark score, human judgment, a model judge (LLM or reward model), a statistic of the answer text, or a registry record. LLMmap and LLM DNA skip the per-benchmark refits, so their rows leave out the three benchmark-transfer tasks (43 tasks). Strict protocol: models held out in 5 publisher-grouped folds $\times$ 3 seeds; a cell is the mean skill over the encoder's width-matched Gaussian floor (the leaderboard scalar reads the exact $d$=1 floor), averaged over folds, seeds and the column's tasks. -full: the 900-question pool; -budget: the $K$=400 questions chosen by Fisher information, the smallest $K$ at which BULLPEN-fused keeps 95\% of its full-pool mean over the \bpRetTasks{} tasks other than benchmark transfer under this protocol (\bpRetPctOne\% at $K$=400, \bpRetKTwoHundred\% at 200). Bold: the 95\% paired-bootstrap CI of the difference from the column's best covers 0, on the tasks both encoders score; the bootstrap resamples held-out folds and seeds in the class columns (their tasks are fixed) and tasks in the two right-hand columns. Underlined: sole winner. The two right-hand columns carry $\pm$ half the task-bootstrap interval of the mean. $M_{\text{eff}}$-weighted: each task weighted by its share of the effective number of independent tasks \citep{li2005meff}, so redundant tasks split one unit. Green shading marks the bold cells; red marks the column's lowest embedding (leaderboard scalar excluded). Gray italic rows: the mean over the embeddings in that channel group. LLMmap: \bpMapOwnN{} of \bpNModels{} models answered its 8 probes, the rest take the mean signature; its probe-refusal cell is omitted (the label comes from those probes).}
\end{table}

\section{Swap controls}
\label{app:controls}

The swap control of Section~\ref{sec:encoders} gives each encoder another model's answers, so any skill
it keeps cannot come from the model's own writing. The tables call these controls shuffled-text twins,
and a \emph{family-shuffled} twin takes answers only from models of another organization.
Table~\ref{tab:paper_control_significance} compares each swap control with the floor, with the
bits-only embedding and with its own encoder (Section~\ref{sec:res-external}). Two comparisons with the
bits also differ slightly in budget or width, since the BULLPEN-zConcat control reads 400 bit questions
and the BULLPEN-CCAfused control keeps \bpDimFusedTwinLo--\bpDimFusedTwinHi{} dimensions. The
BULLPEN-fused swap control still scores above PCA32-bits by \bpTwinFusedVsPca{}
([\bpTwinFusedVsPcaLo, \bpTwinFusedVsPcaHi]), which is the part of BULLPEN-fused's lift that does not
depend on each model's own answers. Adding answer text to the bits helps on \bpLiftBitsPos{} of
\bpLiftBitsN{} tasks, and adding it to the bits and question text helps on \bpLiftBitsQPos{}
(Figure~\ref{fig:paper_task_panels}c,d).
\begin{table}[t]
  \centering
  \small
  \setlength{\abovecaptionskip}{0pt}\setlength{\belowcaptionskip}{8pt}
  \caption{Paired tests of each encoder against its shuffled-text twins.}
  \label{tab:paper_control_significance}
  \begin{adjustbox}{max width=\linewidth}
\begin{tabular}{llrcrr}
\toprule
$A$ & $B$ & $\Delta$ & 95\% CI & wins & $q$ \\
\midrule
BULLPEN-fused-full & BULLPEN-fused-full, shuffled & $+0.068$ & $[+0.03, +0.11]$ & 37/46 & $<$0.001 \\
BULLPEN-fused-full & BULLPEN-fused-full, family-shuffled & $+0.108$ & $[+0.06, +0.17]$ & 35/43 & $<$0.001 \\
BULLPEN-fused-full, shuffled & floor & $+0.431$ & $[+0.36, +0.50]$ & 45/46 & $<$0.001 \\
BULLPEN-fused-full, family-shuffled & floor & $+0.379$ & $[+0.31, +0.45]$ & 42/43 & $<$0.001 \\
BULLPEN-fused-full, shuffled & PCA32-bits & $+0.063$ & $[+0.02, +0.11]$ & 31/46 & 0.005 \\
BULLPEN-fused-full, family-shuffled & PCA32-bits & $+0.030$ & $[-0.00, +0.07]$ & 26/43 & 0.101 \\
BULLPEN-fused-full, shuffled & PCA32-bits (centered) & $+0.043$ & $[+0.01, +0.07]$ & 30/46 & 0.017 \\
BULLPEN-fused-full, family-shuffled & PCA32-bits (centered) & $+0.008$ & $[-0.04, +0.05]$ & 24/43 & 0.243 \\
\midrule
BULLPEN-ans-full & BULLPEN-ans-full, shuffled & $+0.458$ & $[+0.40, +0.52]$ & 45/46 & $<$0.001 \\
BULLPEN-ans-full & BULLPEN-ans-full, family-shuffled & $+0.432$ & $[+0.36, +0.50]$ & 43/43 & $<$0.001 \\
BULLPEN-ans-full, shuffled & floor & $-0.001$ & $[-0.03, +0.02]$ & 26/46 & 0.648 \\
BULLPEN-ans-full, family-shuffled & floor & $+0.016$ & $[-0.02, +0.05]$ & 24/43 & 0.164 \\
BULLPEN-ans-full, shuffled & PCA32-bits & $-0.380$ & $[-0.44, -0.32]$ & 1/46 & $<$0.001 \\
BULLPEN-ans-full, family-shuffled & PCA32-bits & $-0.345$ & $[-0.42, -0.27]$ & 1/43 & $<$0.001 \\
BULLPEN-ans-full, shuffled & PCA32-bits (centered) & $-0.401$ & $[-0.46, -0.34]$ & 1/46 & $<$0.001 \\
BULLPEN-ans-full, family-shuffled & PCA32-bits (centered) & $-0.366$ & $[-0.45, -0.29]$ & 3/43 & $<$0.001 \\
\midrule
BULLPEN-bits+ans-full & BULLPEN-bits+ans-full, shuffled & $+0.027$ & $[-0.01, +0.07]$ & 25/46 & 0.472 \\
BULLPEN-bits+ans-full & BULLPEN-bits+ans-full, family-shuffled & $+0.053$ & $[+0.01, +0.10]$ & 28/43 & 0.033 \\
BULLPEN-bits+ans-full, shuffled & floor & $+0.448$ & $[+0.38, +0.52]$ & 44/46 & $<$0.001 \\
BULLPEN-bits+ans-full, family-shuffled & floor & $+0.409$ & $[+0.34, +0.48]$ & 41/43 & $<$0.001 \\
BULLPEN-bits+ans-full, shuffled & PCA32-bits & $+0.080$ & $[+0.04, +0.13]$ & 35/46 & $<$0.001 \\
BULLPEN-bits+ans-full, family-shuffled & PCA32-bits & $+0.061$ & $[+0.02, +0.10]$ & 30/43 & 0.005 \\
BULLPEN-bits+ans-full, shuffled & PCA32-bits (centered) & $+0.060$ & $[+0.02, +0.10]$ & 33/46 & 0.004 \\
BULLPEN-bits+ans-full, family-shuffled & PCA32-bits (centered) & $+0.039$ & $[-0.01, +0.09]$ & 27/43 & 0.058 \\
\midrule
PCA32-bits+q+ans & PCA32-bits+q+ans, shuffled & $+0.125$ & $[+0.09, +0.17]$ & 39/46 & $<$0.001 \\
PCA32-bits+q+ans & PCA32-bits+q+ans, family-shuffled & $+0.120$ & $[+0.08, +0.17]$ & 36/43 & $<$0.001 \\
PCA32-bits+q+ans, shuffled & floor & $+0.325$ & $[+0.26, +0.39]$ & 43/46 & $<$0.001 \\
PCA32-bits+q+ans, family-shuffled & floor & $+0.319$ & $[+0.26, +0.38]$ & 43/43 & $<$0.001 \\
PCA32-bits+q+ans, shuffled & PCA32-bits & $-0.043$ & $[-0.08, -0.01]$ & 13/46 & 0.001 \\
PCA32-bits+q+ans, family-shuffled & PCA32-bits & $-0.030$ & $[-0.07, +0.02]$ & 11/43 & 0.006 \\
PCA32-bits+q+ans, shuffled & PCA32-bits (centered) & $-0.064$ & $[-0.09, -0.04]$ & 8/46 & $<$0.001 \\
PCA32-bits+q+ans, family-shuffled & PCA32-bits (centered) & $-0.052$ & $[-0.09, -0.01]$ & 9/43 & $<$0.001 \\
\midrule
PCA32-ans & PCA32-ans, shuffled & $+0.351$ & $[+0.29, +0.41]$ & 46/46 & $<$0.001 \\
PCA32-ans & PCA32-ans, family-shuffled & $+0.357$ & $[+0.29, +0.43]$ & 42/43 & $<$0.001 \\
PCA32-ans, shuffled & floor & $+0.005$ & $[-0.01, +0.02]$ & 25/46 & 0.614 \\
PCA32-ans, family-shuffled & floor & $-0.008$ & $[-0.03, +0.02]$ & 23/43 & 0.910 \\
PCA32-ans, shuffled & PCA32-bits & $-0.363$ & $[-0.43, -0.30]$ & 1/46 & $<$0.001 \\
PCA32-ans, family-shuffled & PCA32-bits & $-0.357$ & $[-0.42, -0.29]$ & 1/43 & $<$0.001 \\
PCA32-ans, shuffled & PCA32-bits (centered) & $-0.384$ & $[-0.45, -0.32]$ & 1/46 & $<$0.001 \\
PCA32-ans, family-shuffled & PCA32-bits (centered) & $-0.379$ & $[-0.45, -0.31]$ & 1/43 & $<$0.001 \\
\midrule
PCA32-bits+ans & PCA32-bits+ans, shuffled & $+0.124$ & $[+0.08, +0.17]$ & 38/46 & $<$0.001 \\
PCA32-bits+ans & PCA32-bits+ans, family-shuffled & $+0.126$ & $[+0.08, +0.17]$ & 38/43 & $<$0.001 \\
PCA32-bits+ans, shuffled & floor & $+0.323$ & $[+0.26, +0.39]$ & 45/46 & $<$0.001 \\
PCA32-bits+ans, family-shuffled & floor & $+0.310$ & $[+0.25, +0.37]$ & 43/43 & $<$0.001 \\
PCA32-bits+ans, shuffled & PCA32-bits & $-0.044$ & $[-0.08, -0.00]$ & 13/46 & $<$0.001 \\
PCA32-bits+ans, family-shuffled & PCA32-bits & $-0.039$ & $[-0.08, +0.01]$ & 8/43 & 0.002 \\
PCA32-bits+ans, shuffled & PCA32-bits (centered) & $-0.065$ & $[-0.10, -0.03]$ & 10/46 & $<$0.001 \\
PCA32-bits+ans, family-shuffled & PCA32-bits (centered) & $-0.061$ & $[-0.09, -0.03]$ & 10/43 & $<$0.001 \\
\midrule
BULLPEN-zConcat-budget & BULLPEN-zConcat-budget, shuffled & $+0.142$ & $[+0.10, +0.19]$ & 37/43 & $<$0.001 \\
BULLPEN-zConcat-budget, shuffled & floor & $+0.276$ & $[+0.22, +0.34]$ & 42/43 & $<$0.001 \\
BULLPEN-zConcat-budget, shuffled & PCA32-bits & $-0.073$ & $[-0.11, -0.04]$ & 9/43 & $<$0.001 \\
BULLPEN-zConcat-budget, shuffled & PCA32-bits (centered) & $-0.094$ & $[-0.13, -0.06]$ & 9/43 & $<$0.001 \\
\midrule
BULLPEN-CCAfused-full & BULLPEN-CCAfused-full, shuffled & $+0.150$ & $[+0.10, +0.21]$ & 37/43 & $<$0.001 \\
BULLPEN-CCAfused-full, shuffled & floor & $+0.263$ & $[+0.20, +0.33]$ & 40/43 & $<$0.001 \\
BULLPEN-CCAfused-full, shuffled & PCA32-bits & $-0.091$ & $[-0.13, -0.06]$ & 9/43 & $<$0.001 \\
BULLPEN-CCAfused-full, shuffled & PCA32-bits (centered) & $-0.113$ & $[-0.15, -0.08]$ & 5/43 & $<$0.001 \\
\midrule
BULLPEN-AE-full (bits+text) & BULLPEN-AE-full (bits + noise text) & $+0.094$ & $[+0.05, +0.13]$ & 34/43 & $<$0.001 \\
BULLPEN-AE-full (text) & BULLPEN-AE-full (bits + noise text) & $+0.062$ & $[+0.02, +0.11]$ & 27/43 & 0.032 \\
BULLPEN-AE-full (bits + noise text) & floor & $+0.289$ & $[+0.22, +0.37]$ & 39/43 & $<$0.001 \\
BULLPEN-AE-full (bits + noise text) & PCA32-bits & $-0.059$ & $[-0.10, -0.02]$ & 14/43 & 0.013 \\
BULLPEN-AE-full (bits + noise text) & PCA32-bits (centered) & $-0.081$ & $[-0.13, -0.03]$ & 12/43 & 0.002 \\
\midrule
PCA32-bits & leaderboard scalar & $-0.062$ & $[-0.13, +0.02]$ & 16/46 & 0.053 \\
PCA32-bits (centered) & leaderboard scalar & $-0.041$ & $[-0.11, +0.03]$ & 19/46 & 0.159 \\
\bottomrule
\end{tabular}
  \end{adjustbox}
  \par\smallskip
  \parbox{\linewidth}{\scriptsize Per task under the strict protocol (every strict run: the 46 tasks of Table~\ref{tab:paper_master_table_by_source} where both encoders have a run that scores the three benchmark-transfer tasks, else the other 43; the wins column gives $n$. Moral Machine left out, UGI scored only when pooled). Each twin is the same encoder given another model's answers: shuffled, a derangement over all models; family-shuffled, only answers from another publisher family (scored on the tasks its run covers, see wins). Each block tests one encoder against its twins, then each twin against the floor and both PCA32-bits baselines. $\Delta$: mean paired difference in skill ($s$; skill over floor against the floor row), with a 95\% task bootstrap interval; wins: tasks where $A$ scores higher; $q$: Benjamini--Hochberg within each bits reference's tests. Tasks are correlated ($M_{\mathrm{eff}}\approx37$ over the 46, $\approx34$ over the 43), so $p$ and $q$ are optimistic.}
\end{table}

\section{Tasks that need both channels}
\label{app:complementarity}
Table~\ref{tab:paper_complementarity} tests, for each task, whether the joint embedding
BULLPEN-bits+ans-full beats both the better of the bits and answer-text embeddings and its own swap
control, on the held-out scores of Table~\ref{tab:paper_master_table}. Both tests pool the held-out
predictions where a task has one label per model. After correction, \bpCompNeedBoth{} of the
\bpCompTested{} tasks pass both tests.
\begin{table}[p]
  \centering
  \small
  \setlength{\abovecaptionskip}{0pt}\setlength{\belowcaptionskip}{8pt}
  \caption{Per-task skill of the bits, answer-text and joint embeddings.}
  \label{tab:paper_complementarity}
  \begin{adjustbox}{max width=\linewidth}
\begin{tabular}{llrrrrrc}
\toprule
task & group & bits & text & joint & shuffled & $\Delta$ joint $-$ best single (strict) & both \\
\midrule
\multicolumn{8}{l}{\textit{bits-led}} \\
\quad per-question correctness & benchmark accuracy & 0.51 & 0.36 & 0.46 & 0.44 & -- &  \\
\quad held-out benchmark & benchmark accuracy & 0.67 & 0.65 & 0.71 & 0.74 & -- &  \\
\quad error type & benchmark accuracy & 0.46 & 0.39 & 0.50 & 0.54 & -- &  \\
\quad item information & benchmark accuracy & 0.68 & 0.52 & 0.68 & 0.80 & -- &  \\
\quad benchmark ranking & benchmark accuracy & 0.77 & 0.64 & 0.78 & 0.80 & -- &  \\
\quad mean accuracy & benchmark accuracy & 0.88 & 0.69 & 0.86 & 0.92 & -0.00 [-0.00, -0.00] &  \\
\quad specialization & benchmark accuracy & 0.13 & 0.09 & 0.12 & 0.06 & -- &  \\
\quad can-ai-code sr & public leaderboards & 0.66 & 0.33 & 0.56 & 0.79 & -0.16 [-0.32, -0.02] &  \\
\quad LiveBench language & public leaderboards & 0.66 & 0.49 & 0.66 & 0.81 & -0.26 [-0.51, -0.06] &  \\
\quad OLL v2 GPQA & public leaderboards & 0.44 & 0.21 & 0.35 & 0.36 & -0.14 [-0.29, +0.01] &  \\
\quad OLL v2 MATH & public leaderboards & 0.49 & 0.45 & 0.57 & 0.50 & -0.02 [-0.07, +0.03] &  \\
\quad Open LLM v1 & public leaderboards & 0.63 & 0.54 & 0.66 & 0.70 & -0.09 [-0.15, -0.04] &  \\
\quad Open LLM v2 & public leaderboards & 0.53 & 0.30 & 0.45 & 0.57 & -0.02 [-0.09, +0.06] &  \\
\quad Ayumi RP rank & human preference & 0.42 & 0.37 & 0.48 & 0.47 & -0.18 [-0.29, -0.08] &  \\
\quad architecture & model metadata & 0.36 & 0.25 & 0.33 & 0.28 & -0.08 [-0.17, -0.01] &  \\
\quad parameter count & model metadata & 0.13 & 0.01 & 0.11 & 0.11 & -- &  \\
\quad query routing & model selection & 0.42 & 0.25 & 0.44 & 0.50 & -- &  \\
\midrule
\multicolumn{8}{l}{\textit{shared}} \\
\quad pairwise, per benchmark & benchmark accuracy & 0.49 & 0.50 & 0.44 & 0.42 & -- &  \\
\quad reasoning tilt & benchmark accuracy & 0.30 & 0.29 & 0.42 & 0.39 & -- &  \\
\quad Ayumi RP IQ & human preference & 0.40 & 0.41 & 0.49 & 0.48 & -0.20 [-0.32, -0.08] &  \\
\quad context length & model metadata & 0.38 & 0.38 & 0.47 & 0.45 & -0.05 [-0.10, -0.00] &  \\
\quad top-k selection & model selection & 0.69 & 0.70 & 0.72 & 0.80 & -- &  \\
\midrule
\multicolumn{8}{l}{\textit{text-led}} \\
\quad can-ai-code jr & public leaderboards & 0.03 & 0.12 & 0.13 & -0.07 & +0.04 [-0.16, +0.20] &  \\
\quad IFEval & public leaderboards & 0.25 & 0.35 & 0.24 & 0.31 & +0.03 [-0.10, +0.09] &  \\
\quad Portuguese LLM & public leaderboards & 0.03 & 0.15 & 0.26 & 0.52 & -0.01 [-0.16, +0.15] &  \\
\quad LMArena rating & human preference & 0.19 & 0.59 & 0.52 & 0.65 & -0.11 [-0.29, +0.01] &  \\
\quad LMArena coding & human preference & 0.28 & 0.60 & 0.59 & 0.60 & -0.06 [-0.29, +0.05] &  \\
\quad style delta & response style & 0.17 & 0.64 & 0.37 & 0.30 & -0.17 [-0.34, -0.03] &  \\
\quad style delta, creative & response style & 0.13 & 0.50 & 0.26 & 0.27 & -0.11 [-0.35, +0.03] &  \\
\quad style delta, math & response style & 0.31 & 0.74 & 0.54 & 0.40 & -0.15 [-0.30, -0.02] &  \\
\quad answer divergence & answer behavior & 0.32 & 0.90 & 0.72 & 0.35 & -- &  \\
\quad hedging & answer behavior & 0.51 & 0.77 & 0.71 & 0.62 & +0.09 [+0.01, +0.14] &  \\
\quad answer length & answer behavior & 0.30 & 0.80 & 0.68 & 0.08 & -0.02 [-0.09, +0.06] &  \\
\quad authorship & answer behavior & 0.02 & 0.22 & 0.14 & 0.04 & -- &  \\
\quad refusal rate & answer behavior & 0.02 & 0.31 & 0.20 & -0.02 & -0.18 [-0.29, -0.06] &  \\
\quad reward score & answer behavior & 0.61 & 0.71 & 0.73 & 0.79 & +0.01 [-0.03, +0.03] &  \\
\quad reward style & answer behavior & 0.49 & 0.68 & 0.63 & 0.69 & -- &  \\
\quad probe refusal rate (LLMmap data) & safety & 0.30 & 0.60 & 0.37 & 0.30 & -0.01 [-0.28, +0.26] &  \\
\quad base vs chat & model metadata & 0.42 & 0.52 & 0.47 & 0.30 & -0.03 [-0.08, +0.02] &  \\
\quad model family & model metadata & 0.30 & 0.41 & 0.41 & 0.18 & +0.00 [+0.00, +0.00] &  \\
\quad downloads & model metadata & 0.35 & 0.41 & 0.45 & 0.31 & -0.07 [-0.12, -0.02] &  \\
\quad license & model metadata & 0.14 & 0.25 & 0.17 & 0.30 & -0.06 [-0.12, -0.02] &  \\
\quad release date & model metadata & 0.54 & 0.64 & 0.63 & 0.59 & -0.00 [-0.05, +0.05] &  \\
\quad Ollama quantizations & model metadata & 0.18 & 0.28 & 0.24 & 0.27 & +0.02 [-0.07, +0.10] &  \\
\quad serving providers & model metadata & 0.20 & 0.27 & 0.27 & 0.17 & -0.02 [-0.08, +0.04] &  \\
\quad budgeted selection & model selection & 0.67 & 0.77 & 0.83 & 0.76 & -- &  \\
\midrule
\multicolumn{2}{l}{bits-led (17 tasks), retention: joint / bits} & \multicolumn{6}{l}{0.99} \\
\multicolumn{2}{l}{text-led (24 tasks), recovery: joint / text} & \multicolumn{6}{l}{0.86} \\
\multicolumn{2}{l}{tasks that need both (strict)} & \multicolumn{6}{l}{0 of 31 tested} \\
\bottomrule
\end{tabular}
  \end{adjustbox}
  \par\smallskip
  \parbox{\linewidth}{\scriptsize Does a task need both channels? Per task, skill over floor (strict protocol, 5 publisher-grouped folds $\times$ 3 seeds, the 46 tasks of Table~\ref{tab:paper_master_table_by_source}) of the bits (PCA32-bits (centered)), answer-text (BULLPEN-ans-full), joint (BULLPEN-bits+ans-full) and shuffled-joint (BULLPEN-bits+ans-full (shuffled text)) embeddings, grouped by which single channel leads by more than 0.02. $\Delta$: joint minus the better single channel under model holdout, with its 95\% model-bootstrap interval; -- where no per-model label can be pooled. both: $\Delta>0$ and joint $>$ shuffled (Benjamini--Hochberg $q<0.05$). Summary rows: the joint's share of bits on bits-led tasks (retention) and of text on text-led tasks (recovery).}
\end{table}

\clearpage
\section{Models}
\label{app:models}
{\tiny
\setlength{\tabcolsep}{1.5pt}
\renewcommand{\arraystretch}{1.2}
\setlength{\LTcapwidth}{\textwidth}
\begin{longtable}{@{}>{\raggedright\arraybackslash}p{3.4cm}>{\raggedright\arraybackslash}p{1.2cm}rll@{\hspace{8pt}}c@{\hspace{8pt}}>{\raggedright\arraybackslash}p{3.4cm}>{\raggedright\arraybackslash}p{1.2cm}rll@{}}
\caption{Models in the embedded pool.}\label{tab:paper_model_list}\\
\multicolumn{11}{@{}p{\textwidth}@{}}{\scriptsize All 260 models, by Hugging Face id. $^\dagger$: one of the 104 fullset models, the pool of the full-view $K$ sweep (Figure~\ref{fig:paper_task_panels}b). Family: the publisher or model family that keeps a model's relatives in one fold. Params: billions of parameters. Date: month the repository was created. Fold: its fold of five under the strict protocol's first seed. --: not recorded.}\\
\addlinespace[4pt]
\toprule
model & family & params & date & fold & & model & family & params & date & fold \\
\midrule
\endfirsthead
\multicolumn{11}{@{}l}{\small Table~\ref{tab:paper_model_list} (continued)}\\
\addlinespace[2pt]
\toprule
model & family & params & date & fold & & model & family & params & date & fold \\
\midrule
\endhead
\bottomrule
\endfoot
01-ai/Yi-1.5-9B-Chat$^\dagger$ & yi & 9.0 & 2024-05 & 1 & & deepseek-ai/deepseek-math-7b-instruct$^\dagger$ & deepseek-ai & 7.0 & 2024-02 & 2 \\
01-ai/Yi-6B-Chat & 01-ai & 6.1 & 2023-11 & 2 & & deepseek-ai/DeepSeek-R1-Distill-Llama-8B$^\dagger$ & deepseek-r1 & 8.0 & 2025-01 & 5 \\
Aashraf995/Creative-7B-nerd$^\dagger$ & creative & 7.6 & 2024-12 & 5 & & deepseek-ai/DeepSeek-R1-Distill-Qwen-1.5B$^\dagger$ & qwen & 0.8 & 2025-01 & 2 \\
AdaptLLM/medicine-chat$^\dagger$ & adaptllm & 7.0 & 2023-12 & 2 & & dphn/dolphin-2.1-mistral-7b & dphn & 7.2 & 2023-10 & 5 \\
AdaptLLM/medicine-LLM$^\dagger$ & adaptllm & 7.0 & 2023-09 & 2 & & dphn/dolphin-2.2.1-mistral-7b & dphn & 7.2 & 2023-10 & 5 \\
AdaptLLM/medicine-LLM-13B$^\dagger$ & adaptllm & 13.0 & 2023-12 & 2 & & DreadPoor/Aspire-8B-model\_stock$^\dagger$ & DreadPoor & 8.0 & 2024-09 & 4 \\
ajibawa-2023/Uncensored-Frank-13B & ajibawa-2023 & 13.0 & 2023-09 & 3 & & EleutherAI/llemma\_7b$^\dagger$ & eleutherai & 6.7 & 2023-09 & 3 \\
ajibawa-2023/Uncensored-Frank-7B & ajibawa-2023 & 6.7 & 2023-09 & 3 & & EleutherAI/pythia-12b$^\dagger$ & eleutherai & 12.0 & 2023-02 & 3 \\
ajibawa-2023/Uncensored-Jordan-13B & ajibawa-2023 & 13.0 & 2023-10 & 3 & & eren23/ogno-monarch-jaskier-merge-7b-OH-PREF-DPO$^\dagger$ & eren23 & 7.0 & 2024-02 & 4 \\
ajibawa-2023/Uncensored-Jordan-7B & ajibawa-2023 & 6.7 & 2023-10 & 3 & & EssentialAI/rnj-1-instruct & EssentialAI & 8.3 & 2025-12 & 2 \\
akjindal53244/Arithmo-Mistral-7B & akjindal53244 & 7.2 & 2023-10 & 3 & & FelixChao/llama2-13b-math1.2$^\dagger$ & felixchao & 13.0 & 2023-08 & 2 \\
Alchan/mpt-7b-chat & Alchan & 6.6 & 2024-03 & 3 & & FelixChao/Scorpio-7B$^\dagger$ & felixchao & 7.0 & 2024-02 & 2 \\
allenai/OLMo-2-1124-7B-Instruct$^\dagger$ & olmo & 6.4 & 2024-12 & 3 & & FelixChao/vicuna-7B-chemical$^\dagger$ & felixchao & 7.0 & 2023-08 & 2 \\
allenai/tulu-2-dpo-7b & allenai & 6.7 & 2023-11 & 4 & & FelixChao/vicuna-7B-physics$^\dagger$ & felixchao & 7.0 & 2023-08 & 2 \\
aloobun/TinyAiroboros-2.2.1 & aloobun & 1.1 & 2023-10 & 1 & & FPHam/Free\_Sydney\_V2\_13b\_HF & FPHam & 13.0 & 2023-10 & 5 \\
amazon/MistralLite & amazon & 7.2 & 2023-10 & 5 & & Fredithefish/Guanaco-7B-Uncensored & Fredithefish & 6.7 & 2023-09 & 1 \\
ATH-MaaS/Marco-o1 & marco & 7.6 & 2024-11 & 3 & & golaxy/gowizardlm$^\dagger$ & golaxy & 7.0 & 2023-07 & 4 \\
Austism/chronos-hermes-13b-v2 & Austism & 13.0 & 2023-08 & 3 & & google/gemma-1.1-2b-it & google & 2.5 & 2024-03 & 5 \\
baichuan-inc/Baichuan2-7B-Chat & baichuan-inc & 7.5 & 2023-08 & 3 & & google/gemma-2-9b & google & 9.2 & 2024-06 & 5 \\
bardsai/jaskier-7b-dpo-v5.6$^\dagger$ & bardsai & 7.0 & 2024-02 & 1 & & google/gemma-2-9b-it$^\dagger$ & gemma & 9.2 & 2024-06 & 4 \\
berkeley-nest/Starling-LM-7B-alpha$^\dagger$ & berkeley-nest & 7.0 & 2023-11 & 1 & & google/gemma-2b-it$^\dagger$ & google & 2.0 & 2024-02 & 5 \\
Biomimicry-AI/ANIMA-Nectar-v2$^\dagger$ & biomimicry-ai & 7.0 & 2023-12 & 5 & & google/gemma-3-12b-it$^\dagger$ & gemma & 6.4 & 2025-03 & 4 \\
BioMistral/BioMistral-7B$^\dagger$ & biomistral & 7.2 & 2024-02 & 1 & & google/gemma-3-1b-it$^\dagger$ & gemma & 0.4 & 2025-03 & 4 \\
BioMistral/BioMistral-7B-DARE$^\dagger$ & biomistral & 7.0 & 2024-02 & 1 & & google/gemma-3-4b-it$^\dagger$ & gemma & 6.4 & 2025-02 & 4 \\
BlueNipples/Trion-M-7b & BlueNipples & 7.2 & 2023-11 & 2 & & google/gemma-7b-it$^\dagger$ & gemma & 8.5 & 2024-02 & 4 \\
bofenghuang/vigostral-7b-chat & bofenghuang & 7.2 & 2023-09 & 1 & & Gryphe/MythoLogic-13b & Gryphe & 13.0 & 2023-07 & 1 \\
Brouz/Slerpeno & Brouz & 13.0 & 2023-09 & 1 & & Gryphe/MythoMax-L2-13b & Gryphe & 13.0 & 2023-08 & 1 \\
cais/zephyr\_7b\_r2d2 & cais & 7.2 & 2024-02 & 3 & & haoranxu/ALMA-7B-Pretrain & haoranxu & 6.7 & 2023-09 & 4 \\
CalderaAI/Naberius-7B & CalderaAI & 7.2 & 2023-11 & 5 & & Harshvir/Llama-2-7B-physics$^\dagger$ & harshvir & 2.0 & 2023-08 & 2 \\
cgato/Thespis-13b-v0.3 & cgato & 13.0 & 2023-10 & 4 & & Heralax/Augmental-13b & Heralax & 13.0 & 2023-10 & 5 \\
cgato/Thespis-13b-v0.4 & cgato & 13.0 & 2023-10 & 4 & & Heralax/Augmental-13b-v1.50\_A & Heralax & 13.0 & 2023-10 & 5 \\
cgato/Thespis-13b-v0.5 & cgato & 13.0 & 2023-10 & 4 & & Heralax/Augmental-13b-v1.50\_B & Heralax & 13.0 & 2023-10 & 5 \\
cgato/Thespis-13b-v0.6 & cgato & 13.0 & 2023-11 & 4 & & Heralax/MistralMakise-Merged-13b & Heralax & 13.0 & 2023-10 & 5 \\
cloudyu/Mixtral\_11Bx2\_MoE\_19B$^\dagger$ & cloudyu & 19.0 & 2023-12 & 2 & & Heralax/MythoMakiseMerged-13b & Heralax & 13.0 & 2023-09 & 5 \\
codefuse-ai/CodeFuse-DeepSeek-33B$^\dagger$ & codefuse-ai & 33.0 & 2024-01 & 2 & & hfl/chinese-alpaca-2-7b & hfl & 6.9 & 2023-07 & 1 \\
codellama/CodeLlama-13b-Instruct-hf$^\dagger$ & codellama & 13.0 & 2023-08 & 3 & & HuggingFaceH4/zephyr-7b-alpha & HuggingFaceH4 & 7.2 & 2023-10 & 2 \\
codellama/CodeLlama-7b-hf$^\dagger$ & codellama & 6.7 & 2023-08 & 3 & & HuggingFaceH4/zephyr-7b-beta$^\dagger$ & huggingfaceh4 & 7.2 & 2023-10 & 5 \\
codellama/CodeLlama-7b-Instruct-hf & codellama & 6.7 & 2023-08 & 3 & & HuggingFaceTB/SmolLM2-1.7B-Instruct$^\dagger$ & smollm2 & 1.2 & 2024-10 & 1 \\
CorticalStack/pastiche-crown-clown-7b-dare-dpo$^\dagger$ & corticalstack & 7.0 & 2024-03 & 5 & & huggyllama/llama-13b & huggyllama & 13.0 & 2023-04 & 4 \\
cris177/llama-2-7b-Arguments & cris177 & 6.7 & 2023-10 & 4 & & ibm-granite/granite-3.0-2b-instruct & ibm-granite & 2.6 & 2024-10 & 4 \\
CultriX/NeuralTrix-bf16$^\dagger$ & CultriX & 7.2 & 2024-02 & 2 & & ibm-granite/granite-3.1-2b-instruct & ibm-granite & 2.5 & 2024-12 & 4 \\
Danielbrdz/Barcenas-Mistral-7b & Danielbrdz & 7.2 & 2023-10 & 5 & & ibm-granite/granite-3.1-8b-instruct$^\dagger$ & granite & 8.1 & 2024-12 & 5 \\
Dans-Archive/Dans-TotSirocco-7b & Dans-Archive & 7.2 & 2023-10 & 3 & & IkariDev/Athena-v2 & IkariDev & 13.0 & 2023-09 & 3 \\
deepseek-ai/deepseek-coder-6.7b-instruct$^\dagger$ & deepseek-ai & 6.7 & 2023-10 & 2 & & IkariDev/Athena-v3 & IkariDev & 13.0 & 2023-09 & 3 \\
\newpage
IkariDev/Athena-v4 & IkariDev & 13.0 & 2023-10 & 3 & & microsoft/Orca-2-13b$^\dagger$ & microsoft & 13.0 & 2023-11 & 4 \\
IkariDev/Athnete-13B & IkariDev & 13.0 & 2023-10 & 3 & & microsoft/Orca-2-7b & microsoft & 6.7 & 2023-11 & 4 \\
Intel/neural-chat-7b-v3-3$^\dagger$ & intel & 7.0 & 2023-12 & 3 & & microsoft/phi-1\_5$^\dagger$ & microsoft & 1.0 & 2023-09 & 4 \\
internlm/internlm-chat-7b & -- & -- & 2023-07 & 2 & & microsoft/phi-2$^\dagger$ & phi & 2.8 & 2023-12 & 5 \\
internlm/internlm3-8b-instruct$^\dagger$ & internlm & 9.7 & 2025-01 & 3 & & microsoft/Phi-3-mini-4k-instruct & phi & 3.6 & 2024-04 & 5 \\
itspat/RemmMistral13b & itspat & 13.0 & 2023-11 & 2 & & microsoft/Phi-3-small-8k-instruct & phi & 6.4 & 2024-05 & 5 \\
jaspionjader/fr-18-8b$^\dagger$ & fr & 6.4 & 2025-01 & 3 & & microsoft/Phi-4-mini-instruct$^\dagger$ & phi & 3.6 & 2025-02 & 5 \\
jebcarter/Psyfighter-13B & jebcarter & 13.0 & 2023-11 & 4 & & microsoft/Phi-4-mini-reasoning$^\dagger$ & phi & 3.8 & 2025-04 & 5 \\
jondurbin/airoboros-l2-13b-2.2.1 & jondurbin & 13.0 & 2023-09 & 3 & & migtissera/SynthIA-7B-v1.3 & migtissera & 7.2 & 2023-09 & 4 \\
jondurbin/airoboros-l2-13b-3.0 & jondurbin & 13.0 & 2023-10 & 3 & & mistralai/Ministral-8B-Instruct-2410 & mistralai & 8.0 & 2024-10 & 3 \\
jondurbin/airoboros-l2-13b-3.1.1 & jondurbin & 13.0 & 2023-10 & 3 & & mistralai/Mistral-7B-Instruct-v0.1$^\dagger$ & mistral & 7.2 & 2023-09 & 2 \\
jondurbin/airoboros-l2-13b-gpt4-m2.0 & jondurbin & 13.0 & 2023-07 & 3 & & mistralai/Mistral-7B-Instruct-v0.2 & mistralai & 7.2 & 2023-12 & 3 \\
jondurbin/airoboros-l2-7b-2.2.1 & jondurbin & 6.7 & 2023-09 & 3 & & mistralai/Mistral-7B-Instruct-v0.3$^\dagger$ & mistral & 6.4 & 2024-05 & 2 \\
jondurbin/airoboros-l2-7b-3.0 & jondurbin & 6.7 & 2023-10 & 3 & & mistralai/Mistral-Small-24B-Instruct-2501$^\dagger$ & mistral & 23.6 & 2025-01 & 2 \\
jondurbin/airoboros-m-7b-3.0 & jondurbin & 7.2 & 2023-10 & 3 & & mlabonne/AlphaMonarch-7B$^\dagger$ & mlabonne & 7.0 & 2024-02 & 3 \\
jondurbin/airoboros-m-7b-3.1.2 & jondurbin & 7.2 & 2023-10 & 3 & & Neko-Institute-of-Science/metharme-7b$^\dagger$ & neko-institute-of-science & 7.0 & 2023-04 & 2 \\
jondurbin/spicyboros-13b-2.2 & jondurbin & 13.0 & 2023-09 & 3 & & Neko-Institute-of-Science/pygmalion-7b$^\dagger$ & neko-institute-of-science & 7.0 & 2023-04 & 2 \\
jpacifico/Chocolatine-3B-Instruct-DPO-Revised$^\dagger$ & chocolatine & 3.8 & 2024-07 & 4 & & NeverSleep/Echidna-13b-v0.1 & NeverSleep & 13.0 & 2023-10 & 4 \\
jphme/em\_german\_13b\_v01 & jphme & 13.0 & 2023-09 & 1 & & NeverSleep/Echidna-13b-v0.2 & NeverSleep & 13.0 & 2023-10 & 4 \\
kevin009/llamaRAGdrama$^\dagger$ & kevin009 & 7.0 & 2024-02 & 4 & & NeverSleep/Echidna-13b-v0.3 & NeverSleep & 13.0 & 2023-10 & 4 \\
kfkas/Llama-2-ko-7b-Chat & kfkas & 6.9 & 2023-07 & 5 & & NeverSleep/HornyEchidna-13b-v0.1 & NeverSleep & 13.0 & 2023-10 & 4 \\
kingbri/airolima-chronos-grad-l2-13B & kingbri & 13.0 & 2023-08 & 5 & & NeverSleep/Mistral-11B-SynthIAirOmniMix & NeverSleep & 10.7 & 2023-10 & 4 \\
kingbri/chronolima-airo-grad-l2-13B & kingbri & 13.0 & 2023-08 & 5 & & NeverSleep/Nethena-13B & NeverSleep & 13.0 & 2023-10 & 4 \\
kyujinpy/Sakura-SOLRCA-Math-Instruct-DPO-v1$^\dagger$ & kyujinpy & 7.0 & 2023-12 & 1 & & Nexusflow/Starling-LM-7B-beta$^\dagger$ & nexusflow & 7.0 & 2024-03 & 4 \\
LeoLM/leo-hessianai-7b-chat & LeoLM & 6.7 & 2023-09 & 4 & & nomic-ai/gpt4all-13b-snoozy$^\dagger$ & nomic-ai & 13.0 & 2023-04 & 5 \\
LeoLM/leo-mistral-hessianai-7b-chat & LeoLM & 7.2 & 2023-10 & 4 & & Norquinal/Mistral-7B-claude-chat & Norquinal & 7.2 & 2023-09 & 4 \\
lesubra/ECE-PRYMMAL-3B-SLERP-V1$^\dagger$ & ece & 3.8 & 2024-10 & 3 & & Norquinal/PetrolLM & Norquinal & 7.2 & 2023-09 & 4 \\
LGAI-EXAONE/EXAONE-3.5-7.8B-Instruct$^\dagger$ & exaone & 6.4 & 2024-12 & 3 & & NousResearch/Nous-Capybara-7B-V1.9 & NousResearch & 7.2 & 2023-10 & 5 \\
Lil-R/2\_PRYMMAL-ECE-7B-SLERP$^\dagger$ & 2 & 7.6 & 2024-11 & 2 & & NousResearch/Nous-Hermes-13b$^\dagger$ & nousresearch & 13.0 & 2023-06 & 5 \\
LilRg/10PRYMMAL-3B-slerp$^\dagger$ & 10prymmal & 3.8 & 2024-09 & 3 & & NousResearch/Nous-Hermes-Llama2-13b & NousResearch & 13.0 & 2023-07 & 5 \\
lmsys/vicuna-13b-v1.3 & lmsys & 13.0 & 2023-06 & 4 & & NousResearch/Yarn-Mistral-7b-64k & NousResearch & 7.2 & 2023-10 & 5 \\
lmsys/vicuna-7b-v1.3 & lmsys & 6.7 & 2023-06 & 4 & & oh-yeontaek/llama-2-7B-LoRA-assemble & oh-yeontaek & 6.7 & 2023-09 & 1 \\
lmsys/vicuna-7b-v1.5$^\dagger$ & lmsys & 6.7 & 2023-07 & 4 & & Open-Orca/Mistral-7B-OpenOrca & Open-Orca & 7.2 & 2023-09 & 5 \\
lmsys/vicuna-7b-v1.5-16k$^\dagger$ & lmsys & 7.0 & 2023-07 & 4 & & openbmb/UltraLM-13b-v2.0 & openbmb & 13.0 & 2023-09 & 2 \\
ludis/tsukasa-limarp-7b & ludis & 6.7 & 2023-09 & 3 & & OpenBuddy/openbuddy-codellama2-34b-v11.1-bf16$^\dagger$ & OpenBuddy & 33.8 & 2023-09 & 3 \\
maywell/Synatra-7B-v0.3-RP & maywell & 7.2 & 2023-10 & 1 & & OpenBuddy/openbuddy-mistral-7b-v13 & OpenBuddy & 7.3 & 2023-10 & 3 \\
meta-llama/Llama-2-13b-chat-hf$^\dagger$ & llama2 & 13.0 & 2023-07 & 1 & & OpenBuddy/openbuddy-zephyr-7b-v14.1 & OpenBuddy & 7.3 & 2023-11 & 3 \\
meta-llama/Llama-2-7b-chat-hf$^\dagger$ & llama2 & 6.7 & 2023-07 & 1 & & openchat/openchat-3.5-0106$^\dagger$ & openchat & 3.5 & 2024-01 & 2 \\
meta-llama/Llama-3.1-8B-Instruct$^\dagger$ & llama3 & 8.0 & 2024-07 & 3 & & openchat/openchat-3.5-1210 & openchat & 7.2 & 2023-12 & 2 \\
meta-llama/Llama-3.2-1B-Instruct$^\dagger$ & llama3 & 1.2 & 2024-09 & 3 & & openchat/openchat\_3.5$^\dagger$ & openchat & 3.5 & 2023-10 & 2 \\
meta-llama/Llama-3.2-3B-Instruct$^\dagger$ & llama3 & 3.2 & 2024-09 & 3 & & PharMolix/BioMedGPT-LM-7B$^\dagger$ & pharmolix & 7.0 & 2023-08 & 5 \\
meta-llama/Meta-Llama-3-8B-Instruct$^\dagger$ & meta-llama & 3.0 & 2024-04 & 2 & & pipizhao/Pandalyst-7B-V1.1 & pipizhao & 6.7 & 2023-09 & 2 \\
meta-math/MetaMath-Llemma-7B$^\dagger$ & meta-math & 7.0 & 2023-11 & 3 & & pipizhao/Pandalyst-7B-V1.2 & pipizhao & 6.7 & 2023-10 & 2 \\
meta-math/MetaMath-Mistral-7B$^\dagger$ & meta-math & 7.0 & 2023-10 & 3 & & posicube/Llama2-chat-AYB-13B & posicube & 13.0 & 2023-10 & 4 \\
\newpage
posicube/Llama2-chat-AYT-13B & posicube & 13.0 & 2023-09 & 4 & & TIGER-Lab/MAmmoTH-7B & TIGER-Lab & 6.7 & 2023-09 & 4 \\
princeton-nlp/gemma-2-9b-it-SimPO & princeton-nlp & 9.2 & 2024-07 & 1 & & tiiuae/falcon-40b-instruct$^\dagger$ & tiiuae & 40.0 & 2023-05 & 3 \\
project-baize/baize-v2-13b$^\dagger$ & project-baize & 13.0 & 2023-05 & 4 & & tiiuae/Falcon3-10B-Instruct$^\dagger$ & falcon & 4.5 & 2024-12 & 3 \\
PygmalionAI/mythalion-13b & PygmalionAI & 13.0 & 2023-09 & 5 & & tiiuae/Falcon3-7B-Instruct$^\dagger$ & falcon & 3.2 & 2024-11 & 3 \\
PygmalionAI/pygmalion-2-13b & PygmalionAI & 13.0 & 2023-09 & 5 & & TinyLlama/TinyLlama-1.1B-Chat-v1.0$^\dagger$ & tinyllama & 1.1 & 2023-12 & 1 \\
Q-bert/Optimus-7B$^\dagger$ & q-bert & 7.0 & 2023-12 & 1 & & Tostino/Inkbot-13b-4k & Tostino & 13.0 & 2023-09 & 5 \\
QuixiAI/samantha-mistral-instruct-7b & QuixiAI & 7.2 & 2023-09 & 5 & & Undi95/Amethyst-13B & Undi95 & 13.0 & 2023-09 & 1 \\
QuixiAI/Wizard-Vicuna-13B-Uncensored & QuixiAI & 13.0 & 2023-05 & 5 & & Undi95/Emerhyst-13B & Undi95 & 13.0 & 2023-09 & 1 \\
QuixiAI/Wizard-Vicuna-7B-Uncensored & QuixiAI & 6.7 & 2023-05 & 5 & & Undi95/Mistral-11B-Airoboros-RP-v1 & Undi95 & 10.7 & 2023-10 & 1 \\
QuixiAI/WizardLM-7B-Uncensored & QuixiAI & 6.7 & 2023-05 & 5 & & Undi95/Mistral-ClaudeLimaRP-v3-7B & Undi95 & 7.2 & 2023-09 & 1 \\
Qwen/Qwen-7B-Chat & Qwen & 7.7 & 2023-08 & 5 & & Undi95/Mistral-PetroLimaRP-v3-12B & Undi95 & 11.2 & 2023-09 & 1 \\
Qwen/Qwen1.5-0.5B-Chat$^\dagger$ & qwen & 0.5 & 2024-01 & 2 & & Undi95/MistRP-Airoboros-7B & Undi95 & 7.2 & 2023-10 & 1 \\
Qwen/Qwen1.5-4B-Chat$^\dagger$ & qwen & 4.0 & 2024-01 & 2 & & Undi95/MLewdBoros-LRPSGPT-2Char-13B & Undi95 & 13.0 & 2023-09 & 1 \\
Qwen/Qwen1.5-7B-Chat$^\dagger$ & qwen & 7.0 & 2024-01 & 2 & & Undi95/Nete-13B & Undi95 & 13.0 & 2023-10 & 1 \\
Qwen/Qwen2-0.5B-Instruct & Qwen & 0.5 & 2024-06 & 5 & & Undi95/OpenRP-13B & Undi95 & 13.0 & 2023-09 & 1 \\
Qwen/Qwen2-1.5B-Instruct & Qwen & 1.5 & 2024-06 & 5 & & Undi95/OpenRP-13B-SuperCOT & Undi95 & 13.0 & 2023-09 & 1 \\
Qwen/Qwen2-7B-Instruct & Qwen & 7.6 & 2024-06 & 5 & & Undi95/PsyMedRP-v1-13B & Undi95 & 13.0 & 2023-10 & 1 \\
Qwen/Qwen2.5-0.5B-Instruct$^\dagger$ & qwen2.5 & 0.5 & 2024-09 & 2 & & Undi95/ReMM-Lion-13B & Undi95 & 13.0 & 2023-09 & 1 \\
Qwen/Qwen2.5-1.5B-Instruct$^\dagger$ & qwen & 0.8 & 2024-09 & 2 & & Undi95/ReMM-S-Kimiko-v2-13B & Undi95 & 13.0 & 2023-09 & 1 \\
Qwen/Qwen2.5-3B-Instruct$^\dagger$ & qwen & 1.8 & 2024-09 & 2 & & Undi95/ReMM-v1-LRPSGPT-2Char-13B & Undi95 & 13.0 & 2023-09 & 1 \\
Qwen/Qwen2.5-7B-Instruct & qwen & 4.3 & 2024-09 & 2 & & Undi95/ReMM-v2-Kimiko-v2-13B & Undi95 & 13.0 & 2023-09 & 1 \\
Qwen/Qwen2.5-Coder-7B-Instruct$^\dagger$ & qwen & 4.3 & 2024-09 & 2 & & Undi95/UndiMix-v1-13b & Undi95 & 13.0 & 2023-08 & 1 \\
Qwen/Qwen2.5-Math-7B-Instruct$^\dagger$ & qwen & 4.3 & 2024-09 & 2 & & Undi95/UndiMix-v2-13b & Undi95 & 13.0 & 2023-09 & 1 \\
Qwen/Qwen3-4B$^\dagger$ & qwen & 2.8 & 2025-04 & 2 & & Undi95/UndiMix-v3-13B & Undi95 & 13.0 & 2023-09 & 1 \\
Riiid/sheep-duck-llama-2-13b & Riiid & 13.0 & 2023-10 & 5 & & Undi95/UndiMix-v4-13B & Undi95 & 13.0 & 2023-09 & 1 \\
rishiraj/CatPPT-base$^\dagger$ & rishiraj & 7.0 & 2023-12 & 3 & & Undi95/Unholy-v1-10L-13B & Undi95 & 13.0 & 2023-09 & 1 \\
royallab/Pygmalion-2-13b-SuperCOT2 & royallab & 13.0 & 2023-09 & 4 & & Undi95/Unholy-v1-12L-13B & Undi95 & 13.0 & 2023-09 & 1 \\
sail/Sailor2-8B-Chat$^\dagger$ & sailor & 4.9 & 2024-12 & 4 & & Undi95/Unholy-v1.1-13B & Undi95 & 13.0 & 2023-09 & 1 \\
SciPhi/SciPhi-Mistral-7B-32k$^\dagger$ & sciphi & 7.0 & 2023-10 & 1 & & Undi95/Utopia-13B & Undi95 & 13.0 & 2023-11 & 1 \\
SciPhi/SciPhi-Self-RAG-Mistral-7B-32k$^\dagger$ & sciphi & 7.0 & 2023-10 & 1 & & Undi95/Xwin-MLewd-13B-V0.2 & Undi95 & 13.0 & 2023-10 & 1 \\
shleeeee/mistral-ko-tech-science-v1$^\dagger$ & shleeeee & 7.0 & 2023-12 & 1 & & Undi95/ZettaPi-13B & Undi95 & 13.0 & 2023-09 & 1 \\
SkunkworksAI/Mistralic-7B-1 & -- & -- & 2023-09 & 2 & & upstage/SOLAR-10.7B-Instruct-v1.0$^\dagger$ & upstage & 10.7 & 2023-12 & 1 \\
stabilityai/StableBeluga-13B & stabilityai & 13.0 & 2023-07 & 4 & & vanillaOVO/WizardCoder-Python-7B-V1.0 & vanillaOVO & 6.7 & 2024-06 & 4 \\
stabilityai/StableBeluga-7B & stabilityai & 6.7 & 2023-07 & 4 & & Weyaxi/SlimOpenOrca-Mistral-7B & Weyaxi & 7.2 & 2023-10 & 5 \\
stabilityai/stablelm-tuned-alpha-7b$^\dagger$ & stabilityai & 7.0 & 2023-04 & 4 & & WizardLM/WizardLM-13B-V1.2 & WizardLM & 13.0 & 2023-07 & 2 \\
teknium/CollectiveCognition-v1.1-Mistral-7B & teknium & 7.2 & 2023-10 & 2 & & XiaomiMiMo/MiMo-7B-RL$^\dagger$ & mimo & 7.6 & 2025-04 & 4 \\
teknium/Mistral-Trismegistus-7B & teknium & 7.2 & 2023-10 & 2 & & Xwin-LM/Xwin-LM-13B-V0.2 & Xwin-LM & 13.0 & 2023-10 & 4 \\
teknium/OpenHermes-2-Mistral-7B & teknium & 7.2 & 2023-10 & 2 & & yam-peleg/Experiment26-7B$^\dagger$ & yam-peleg & 7.0 & 2024-02 & 2 \\
teknium/OpenHermes-2.5-Mistral-7B & teknium & 7.2 & 2023-10 & 2 & & zai-org/agentlm-13b & zai-org & 13.0 & 2023-10 & 5 \\
TheBloke/guanaco-7B-HF & TheBloke & 6.7 & 2023-05 & 2 & & zai-org/agentlm-7b & zai-org & 6.7 & 2023-10 & 5 \\
TheBloke/Llama-2-70B-Chat-AWQ$^\dagger$ & thebloke & 70.0 & 2023-09 & 5 & & zai-org/chatglm3-6b & zai-org & 6.2 & 2023-10 & 5 \\
THUDM/chatglm2-6b & -- & -- & -- & 2 & & zai-org/glm-4-9b-chat$^\dagger$ & glm & 8.1 & 2024-06 & 1 \\
\end{longtable}
}

\end{document}